%% file: main.tex
\documentclass[twoside]{article}

\usepackage{natbib}

\PassOptionsToPackage{dvipsnames}{xcolor}
\usepackage[accepted]{aistats2023}

\usepackage[utf8]{inputenc} 
\usepackage[T1]{fontenc}    
\usepackage{hyperref}       
\usepackage{url}            
\usepackage{booktabs}       
\usepackage{amsfonts}       
\usepackage{nicefrac}       
\usepackage{microtype}      
\usepackage{xcolor}         
\usepackage{amsmath}
\usepackage{bbm}
\usepackage{graphicx}
\usepackage{subcaption}
\usepackage{soul}

\usepackage{algorithm}
\usepackage{algorithmic}

\usepackage{amsfonts}
\usepackage{amsmath}
\usepackage{amssymb}
\usepackage{dsfont}
\usepackage{tcolorbox}
\usepackage{amsthm}
\theoremstyle{definition}
\newtheorem{proposition}{Proposition}[section]
\newtheorem{definition}{Definition}[section]

\usepackage{xspace}  
\usepackage[normalem]{ulem}  
\usepackage{wrapfig}
\usepackage{appendix}
\usepackage[algo2e,ruled,vlined]{algorithm2e}
\usepackage{mathrsfs}

\input{macros}

\begin{document}

\runningtitle{Bayesian Optimization with Conformal Prediction Sets}
\runningauthor{}

\twocolumn[
    \aistatstitle{Bayesian Optimization with Conformal Prediction Sets}
    \aistatsauthor{ Samuel Stanton\textsuperscript{1,2} \And Wesley Maddox\textsuperscript{2} \And  Andrew Gordon Wilson\textsuperscript{2} }
    \aistatsaddress{ Prescient Design, Genentech\textsuperscript{1} \And New York University\textsuperscript{2} }
]

\definecolor{lightblue}{HTML}{afe9af}
\definecolor{lighterblue}{HTML}{afe9af}
\newtcolorbox{mybox}{colback=lighterblue,colframe=lightblue}

\begin{abstract}
Bayesian optimization is a coherent, ubiquitous approach to decision-making under uncertainty, with applications including multi-arm bandits, active learning, and black-box optimization.
Bayesian optimization selects decisions (i.e. objective function queries) with maximal expected utility with respect to the posterior distribution of a Bayesian model, which quantifies reducible, epistemic uncertainty about query outcomes.
In practice, subjectively implausible outcomes can occur regularly for two reasons: 1) model misspecification and 2) covariate shift.
Conformal prediction is an uncertainty quantification method with coverage guarantees even for misspecified models and a simple mechanism to correct for covariate shift.
We propose conformal Bayesian optimization, which directs queries towards regions of search space where the model predictions have guaranteed validity, and investigate its behavior on a suite of black-box optimization tasks and tabular ranking tasks.
In many cases we find that query coverage can be significantly improved without harming sample-efficiency.
\end{abstract}

\section{INTRODUCTION}
\label{sec:intro}

Bayesian optimization (BayesOpt) is a popular strategy to focus data collection towards improving a specific objective, such as discovering useful new materials or drugs \citep{terayama2021black, wang2022bayesian}.
BayesOpt relies on a Bayesian model of the objective (a surrogate model) to select new observations (queries) that maximize the user's expected utility.
If the surrogate does not fit the objective well, then the expected utility of new queries may not correspond well at all to their \textit{actual} utility, leading to little or no improvement in the objective value after many rounds of data collection.

\begin{figure*}
    \centering
    \captionsetup[subfigure]{justification=centering}
    \begin{subfigure}{0.24\textwidth}
        \includegraphics[height=2.7cm]{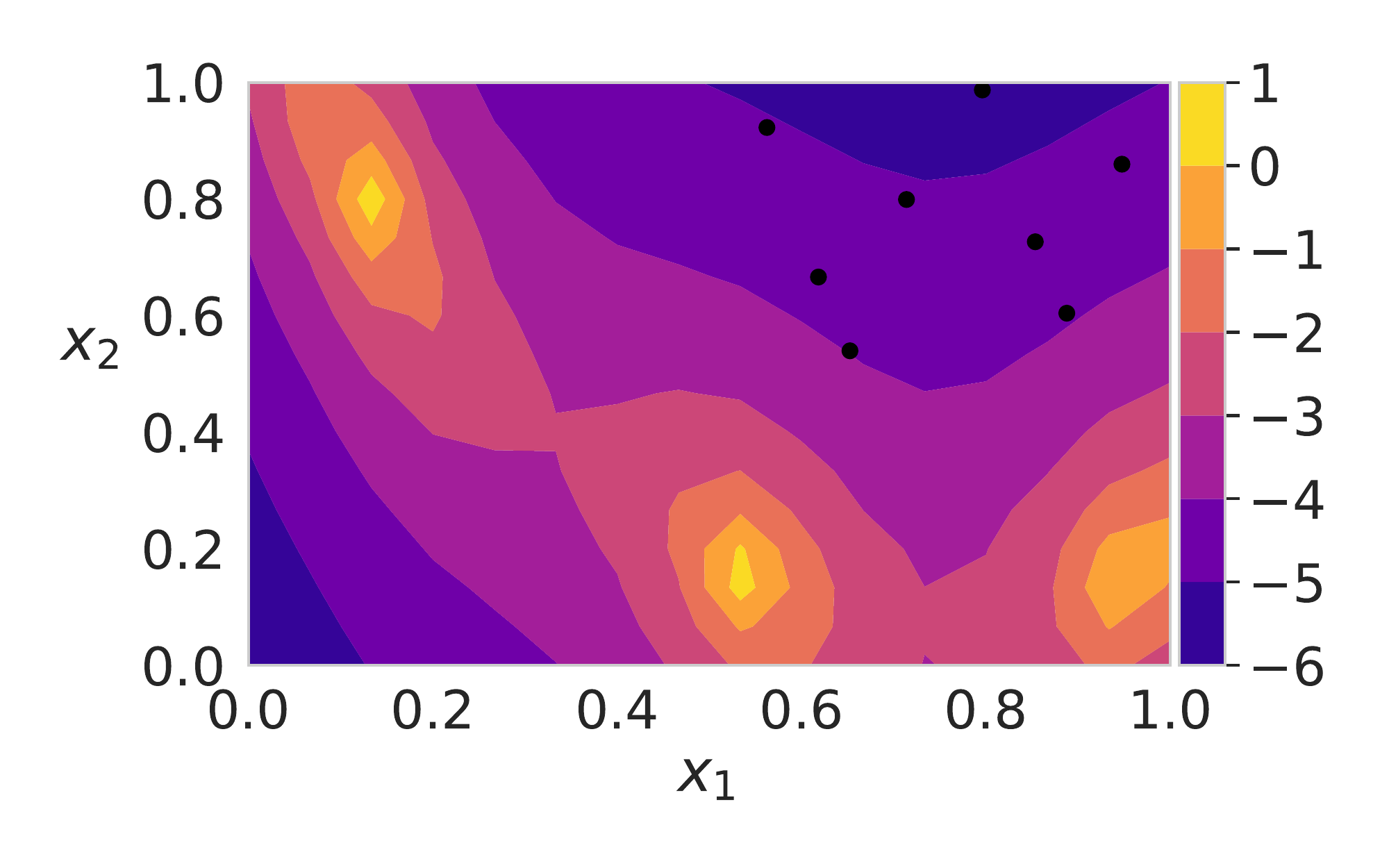}
        \caption{Objective fn.}
        \label{fig:branin_surf}
    \end{subfigure}
    \begin{subfigure}{0.24\textwidth}
        \includegraphics[height=2.7cm]{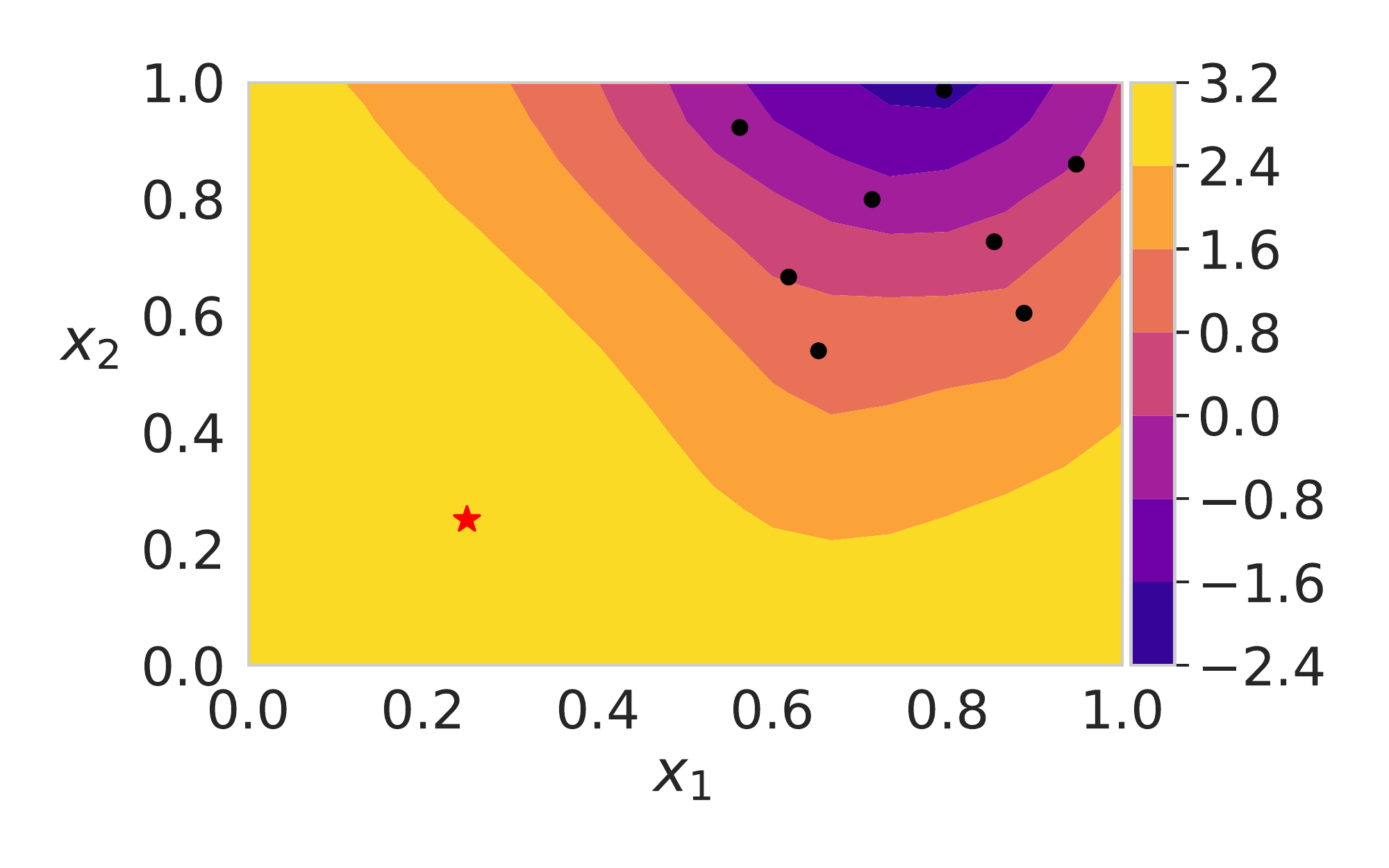}
        \caption{UCB}
        \label{fig:branin_ucb}
    \end{subfigure}
    \begin{subfigure}{0.24\textwidth}
        \includegraphics[height=2.7cm]{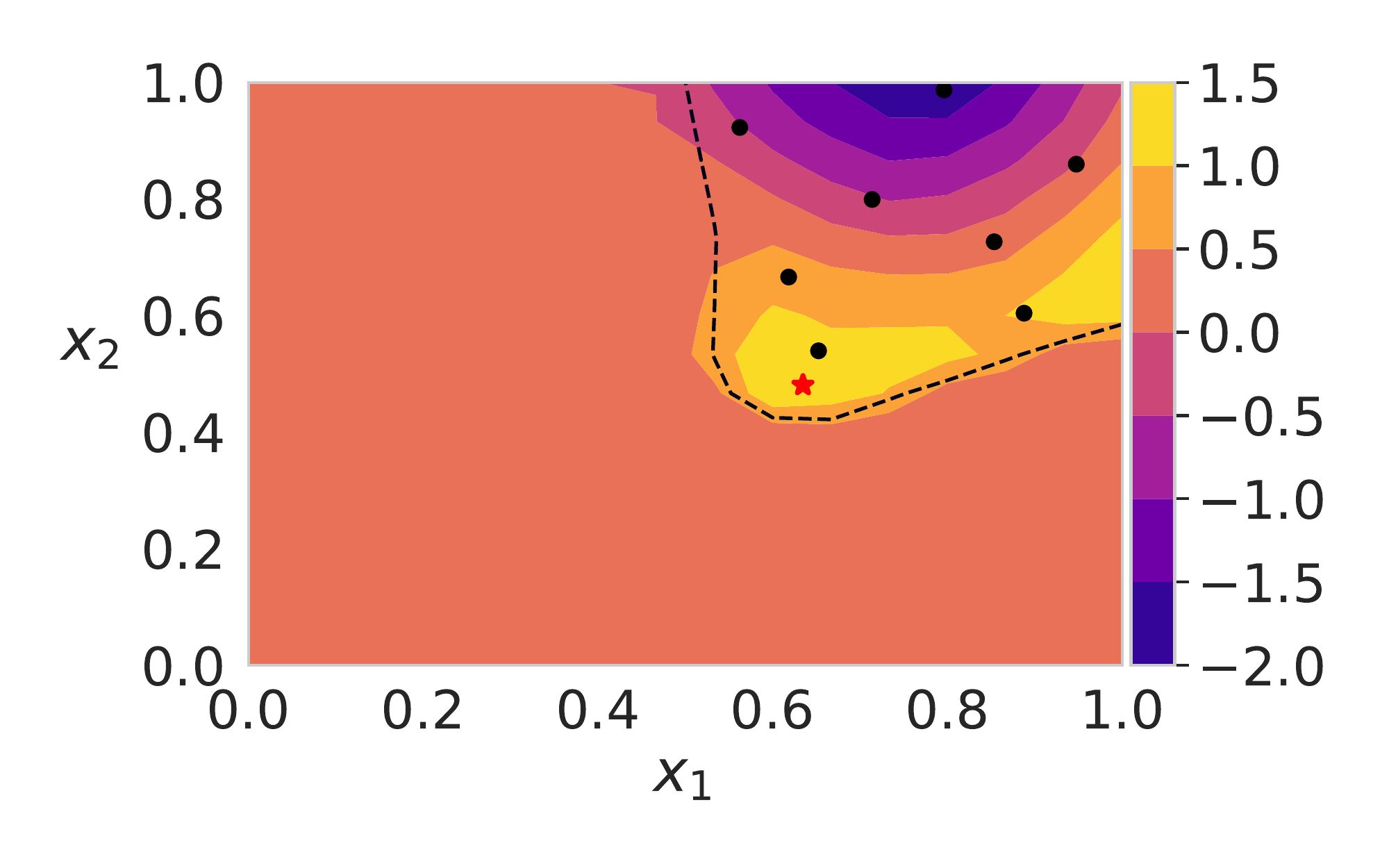}
        \caption{CUCB}
        \label{fig:branin_cucb}
    \end{subfigure}
    \begin{subfigure}{0.24\textwidth}
        \includegraphics[height=2.7cm]{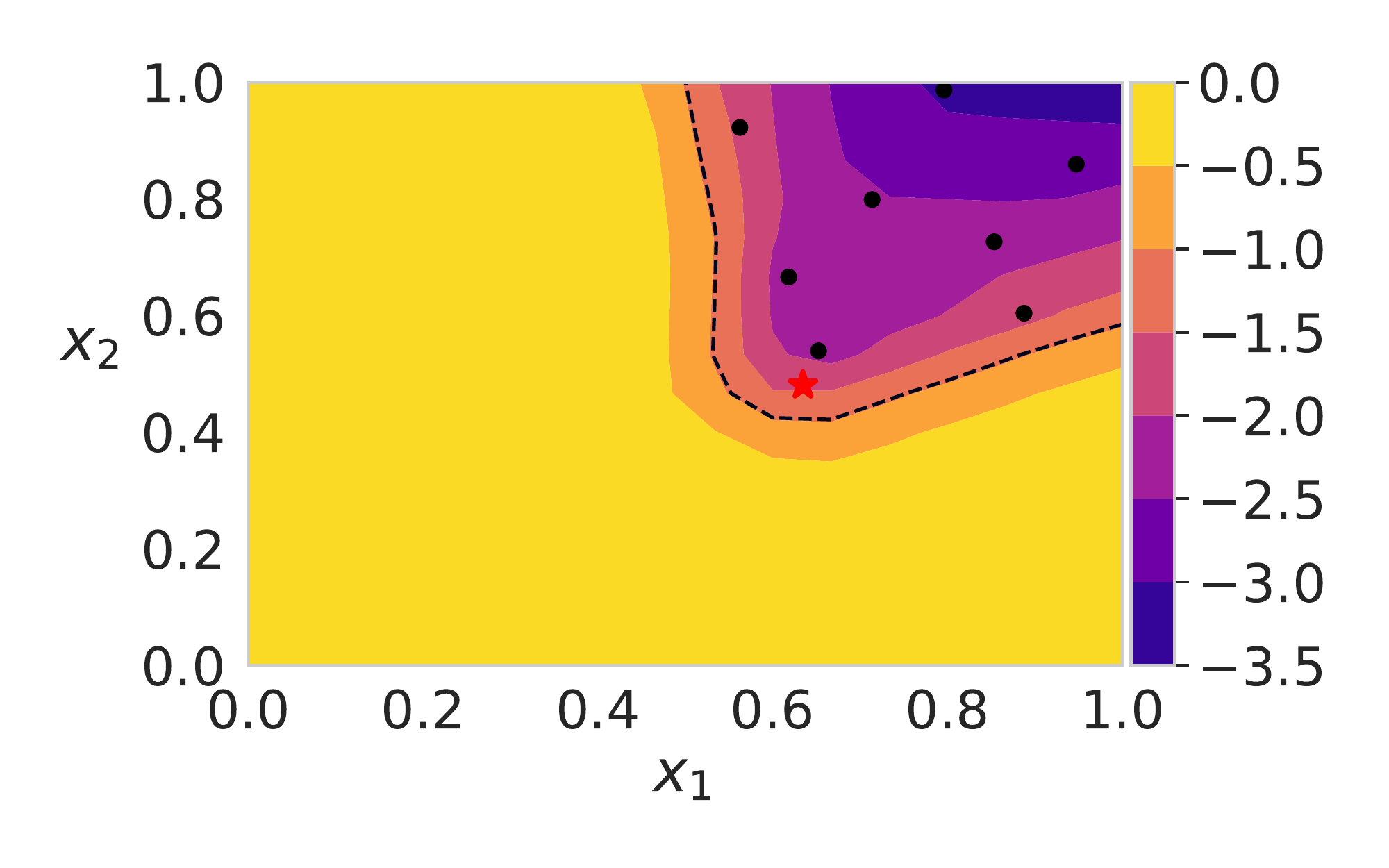}
        \caption{Normalized density ratio}
        \label{fig:branin_cucb_impweights}
    \end{subfigure}
    \caption{
    A motivating example of feedback covariate shift.
    We want $\vec x^* \in [0, 1]^2$ which maximizes the Branin objective \textbf{(a)}, starting from $8$ examples in the upper right (the black dots).
    The upper-confidence bound (UCB) acquisition function \textbf{(b)} selects the next query (the \textcolor{red}{red} star) far from any training data, where we cannot guarantee reliable predictions.
    In higher dimensions, we will exhaust our query budget long before covering the whole search space with training data.
    Given a miscoverage tolerance $\alpha = 1 / \sqrt{8}$, conformal UCB \textbf{(c)} directs the search to the region where conformal predictions are guaranteed coverage of at least $(1 - \alpha)$.
    \textbf{(d)} The dashed line is the set $\vec x$ such that $w(\vec x) \propto p_{\mathrm{query}}(\vec x) / p_{\mathrm{train}}(\vec x)$ is exactly $\alpha$.
    }
    \label{fig:branin}
\end{figure*}

The most practical way to check how well the surrogate fits the objective is to compute its accuracy on a random heldout subset of the training data.
Unfortunately such a holdout set is not at all representative of points we are likely to query since the goal is to find queries that are \textit{better} than the training data in some way.
In other words there is \textit{feedback covariate shift} between the likely query points and the existing training data which degrades the accuracy of the surrogate \citep{fannjiang2022conformal}. 
Even without covariate shift, we cannot guarantee the accuracy of the surrogate predictions at all, and instead can only hope that the predictions are accurate enough to provide a useful signal for data collection.

The crux of the issue is that the \textit{coverage} (i.e.,\ the frequency that a prediction set contains the true outcome over many repeated measurements) of Bayes credible prediction sets is directly tied to the correctness of our modeling assumptions, which we cannot entirely control \citep{datta2000bayesian, duanmu2020existence}.
We would prefer the price of assumption error to be lost \textit{efficiency} (i.e.,\ wider prediction sets), rather than poor coverage.

Conformal prediction is a distribution-free uncertainty quantification method which provides prediction sets with reliable coverage under very mild assumptions \citep{vovk2005algorithmic}.
In particular, conformal prediction can accomodate \textit{post hoc} covariate shift (i.e.,\ covariate shift that is only known after training the surrogate) and does not assume the surrogate is well-specified (e.g.,\ we could use a linear model on data following a cubic trend).
Unfortunately conformal prediction is challenging to use in a BayesOpt algorithm since it is non-differentiable, requires continuous outcomes to be discretized, and needs density ratio estimates for unknown densities.
Furthermore, because conformal prediction sets are defined over observable outcomes, they cannot distinguish between epistemic and aleatoric uncertainty, a distinction that is important for effective exploration.

In this work we present conformal Bayesian optimization with a motivating example in Figure \ref{fig:branin}.
Conformal BayesOpt adjusts how far new queries will move from the training data by choosing an acceptable miscoverage tolerance $\alpha \in (0, 1]$. 
If $\alpha = 1$ then we recover conventional BayesOpt, but if $\alpha < 1$ then the search will be directed to the region where conformal predictions are guaranteed coverage of at least $(1 - \alpha)$, keeping feedback covariate shift in check and accounting for potential error in modeling assumptions.

In summary, our contributions are as follows:
\begin{itemize}
    \item We show how to integrate conformal prediction into BayesOpt through the conformal Bayes posterior, with corresponding generalizations of common BayesOpt acquisition functions, enabling the reliable coverage of conformal prediction while still distinguishing between epistemic and aleatoric uncertainty in a principled way.
    \item An efficient, differentiable implementation of full conformal Bayes for Gaussian process (GP) regression models, which is necessary for effective query optimization, and a practical procedure to estimate the density ratio for BayesOpt query proposal distributions.
    \item Demonstrations on synthetic black-box optimization tasks and real tabular ranking tasks that conformal BayesOpt has superior sample-efficiency when the surrogate is misspecified and is comparable otherwise, while improving query coverage significantly. Note that while conformal BayesOpt has promising performance, our goal is not primarily to ``beat'' classical alternatives; rather, we show how to introduce conformal prediction into BayesOpt, and explore the corresponding empirical behaviour and results.
    \footnote{Code is available at \url{github.com/samuelstanton/conformal-bayesopt.git}}
\end{itemize}

\section{PRELIMINARIES}
\label{sec:background}

In this work we will focus on black-box optimization problems of the form $\max_{\vec x \in \mathcal{X}}(f_1^*(\vec x), \dots, f_d^*(\vec x))$, where each $f_i^*: \mathcal{X} \rightarrow \mathbb{R}$ is an unknown function of decision variables $\vec x \in \mathcal{X}$, and $d$ is the number of objectives.
We do not observe $f^*$ directly, but instead receive noisy outcomes (i.e. labels) $\vec y \in \mathcal{Y}$ according to some likelihood $p^*(\vec y | f)$.

\begin{figure*}
    \centering
    \begin{subfigure}{0.24\textwidth}
        \includegraphics[height=2.7cm]{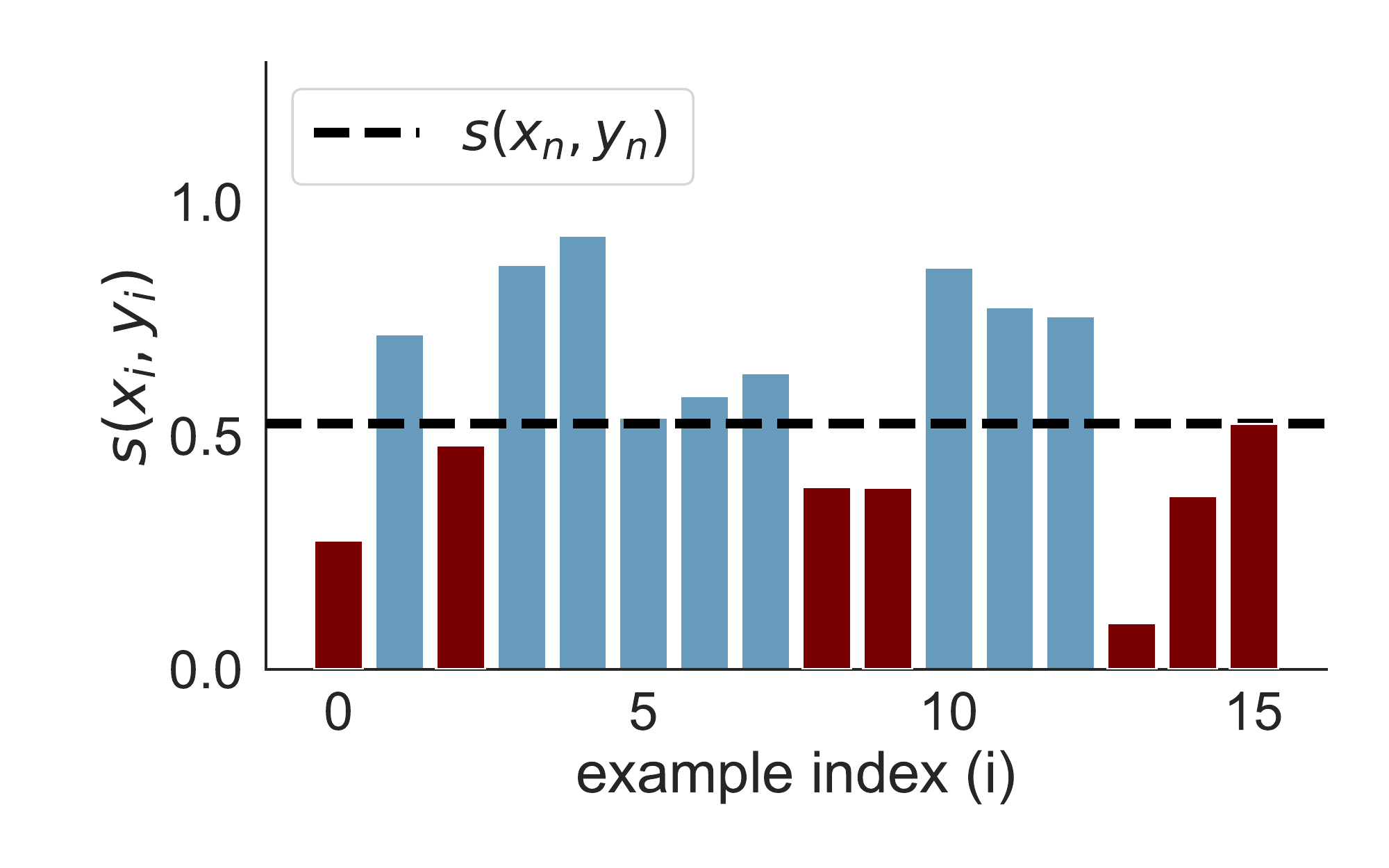}
        \caption{conformal scores $\vec s$}
    \end{subfigure}
    \hfill
    \begin{subfigure}{0.24\textwidth}
        \includegraphics[height=2.7cm]{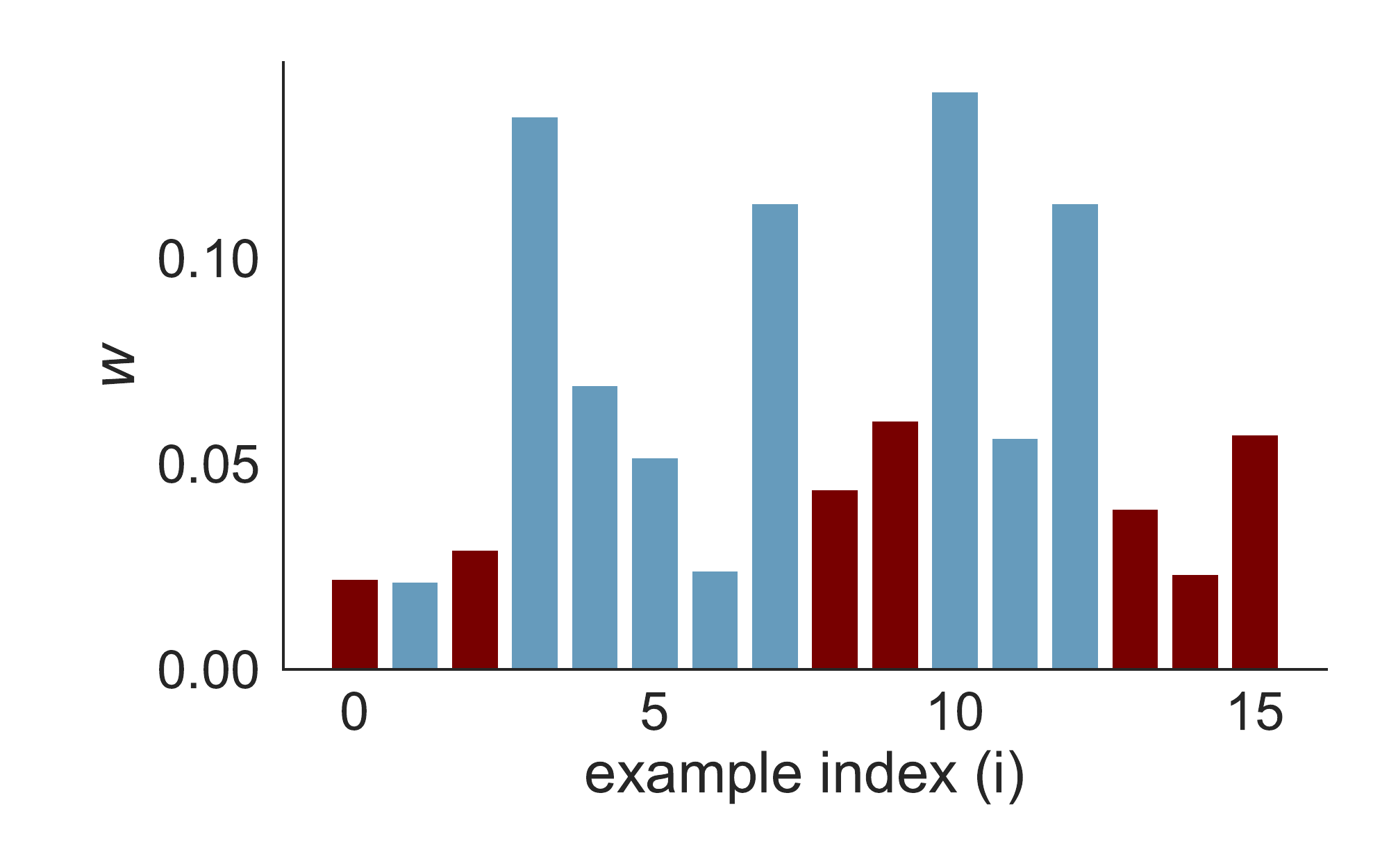}
        \caption{imp. weights $\vec w$}
    \end{subfigure}
    \hfill
    \begin{subfigure}{0.24\textwidth}
        \includegraphics[height=2.7cm]{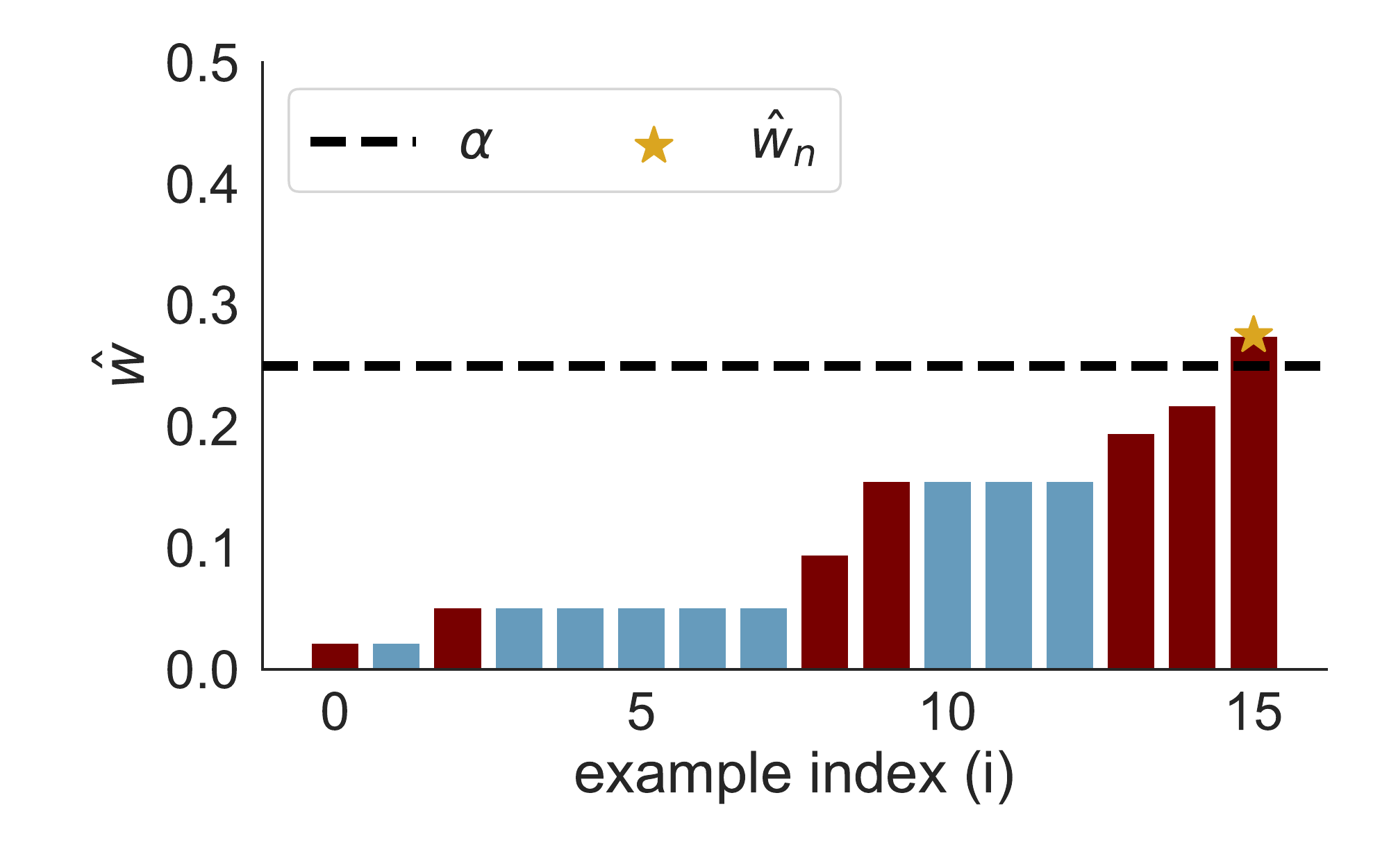}
        \caption{IW partial sums $\overline{\vec w}$}
    \end{subfigure}
    \hfill
    \begin{subfigure}{0.24\textwidth}
        \includegraphics[height=2.7cm]{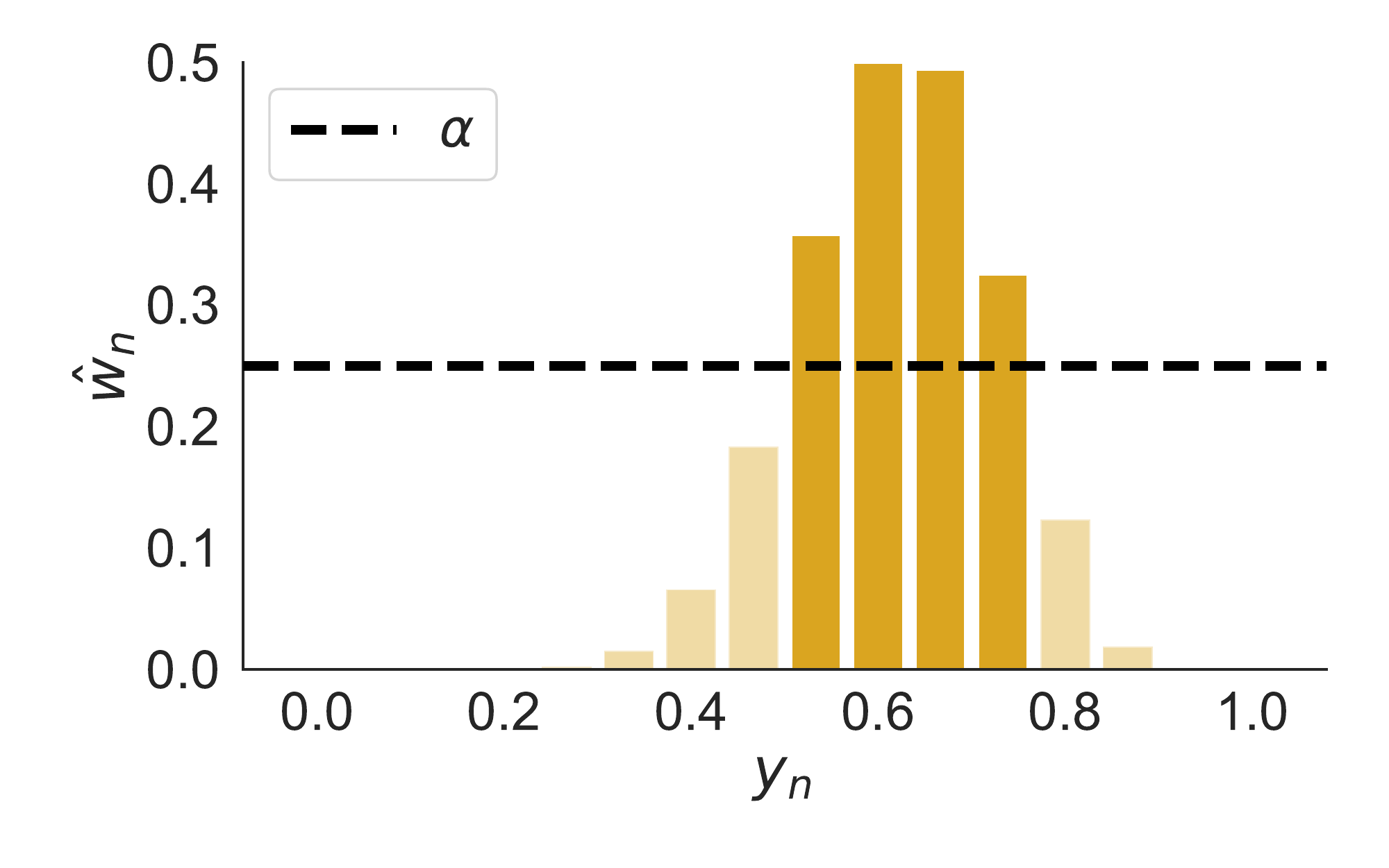}
        \caption{$\mathcal{C}_\alpha(\vec x_n)$}
    \end{subfigure}
    \caption[Visualizing conformal prediction with continuous labels]{
    Constructing a conformal prediction set $C_\alpha(\vec x_n)$ in the regression setting.
    First, \textbf{(a)} we choose some $\hat{\vec y}_n \in \mathcal{Y}$ and guess $\vec y_n = \hat{\vec y}_n$, computing conformal scores $\vec s$ of $\{(\vec x_0, \vec y_0), \dots, (\vec x_{n-1}, \vec y_{n-1}), (\vec x_n, \hat{\vec y}_n)\}$.
    \textbf{(b)} we note which examples score better than our guess (shown in \textcolor{palette5}{blue}), and mask out the corresponding importance weights $\vec w$.
    \textbf{(c)} we compute the partial sums $\overline{\vec w}$ of the masked importance weights, adding $\hat{\vec y}_n$ to $C_\alpha(\vec x_n)$ if $\overline{w}_n > \alpha$,
    \textbf{(d)} repeat steps \textbf{(a - c)} for many guesses of $\vec y_n$. Rejected and accepted guesses are shaded light and dark, respectively.
    }
    \label{fig:construct_conf_pred_set_example}
\end{figure*}

\subsection{Bayesian optimization}
\label{subsec:bayesopt_background}

BayesOpt alternates between \textit{inference} and \textit{selection}, inferring the expected utility of potential query points from available data, which then serves as a proxy objective to select the next batch of observations, which are fed back into the inference procedure, completing one iteration of a repeating loop \citep{brochu2010tutorial, frazier2018tutorial}. 
Inference consists of applying Bayes rule to a prior $p(f)$, a likelihood $p(\vec y | f)$ and dataset $\mathcal{D} = \{(\vec x_i, \vec y_i)\}_{i=0}^{n-1}$ to obtain a Bayes posterior $p(f | \mathcal{D})$.
The expected utility of $\vec x$ is given by an acquisition function $a: \mathcal{X} \rightarrow \mathbb{R}$ with the general form
\begin{align}
    a(\vec x, \mathcal{D}) = \int u(\vec x, f, \mathcal{D})p(f | \mathcal{D})df,
    \label{eq:acq_fn_general_form}
\end{align}
where $u$ is a user-specified utility function. 
For example, taking $u(\vec x, f, \mathcal{D}) = [f(\vec x) - \max_{\vec y_i \in \mathcal{D}} \vec y_i]_+$, where $[\cdot]_+ = \max(\cdot, 0)$, yields the expected improvement (EI) acquisition function \citep{jones1998efficient}.
Since $a$ is the Bayes posterior expectation of $u$, maximizers $\vec x^* = \argmax_{\vec x \in \mathcal{X}} a(\vec x, \mathcal{D})$ are \textit{Bayes-optimal} with respect to $u$.
Bayes-optimality means a decision is \textit{coherent} with our posterior beliefs about $f^*$.
We think of $\argmax_{\vec x \in \mathcal{X}} a(\vec x, \mathcal{D})$ as inducing a distribution $p_{\mathrm{query}}(\vec x) = p'(\vec x | \mathcal{D}) \propto \exp\{a(\vec x, \mathcal{D})\}$ \citep{levine2018reinforcement}.

\subsection{Bayesian inference and model misspecification}
\label{subsec:bayes_limitations}

One way to assess the quality of our posterior beliefs is to check the coverage of the corresponding Bayes $\beta$-credible prediction sets, which are subsets $\mathcal{K}_\beta(\vec x) \subseteq \mathcal{Y}$ satisfying
\begin{align}
    \beta = \int_{\vec y \in \mathcal{K}_\beta(\vec x)} \int p(\vec y | f(\vec x))p(f | \mathcal{D})df d\vec y, \label{eq:bayes_cred_set}
\end{align}
where $\beta \in (0, 1]$ is the level of subjective credibility \citep{gneiting2007probabilistic}.
$\beta$-credible sets may exhibit poor coverage, meaning the frequency of ``implausible'' events outside the set happening is much more than $1 - \beta$ \citep{BACHOC201355}.
Note that poor coverage does not necessarily imply that $\mathcal{K}_\beta(\vec x)$ was computed incorrectly, it may simply indicate a faulty assumption.

For example, in BayesOpt it is very common to assume $f^* \sim \mathcal{GP}(0, \kappa)$, where $\kappa$ is a Mat\'ern kernel.
Mat\'ern kernels support functions that are at least once differentiable, and can struggle to model objectives with discontinuities.
As another example, we typically do not know the true likelihood $p^*(\vec y | f)$, and often choose a simple homoscedastic likelihood $p(\vec y | f) = \mathcal{N}(f, \sigma^2 I_d)$ for convenience.
In reality the true noise process may be correlated with $\vec x$, across objectives, or may not be Gaussian at all \citep{assael2014heteroscedastic,griffiths2019achieving,makarova2021}.
These examples are common instances of \textit{model misspecification}.

In practice faulty assumptions are nearly inevitable, and they are not always harmful, since simplifying assumptions can confer significant practical benefits.
Indeed, theoretical convergence rates for acquisition functions like UCB \citep{srinivas2010gaussian} suggest that for BayesOpt we want to use the \textit{smoothest possible} model, subject to the constraint that we can still model $f^*$ \textit{sufficiently} well.
Similarly Gaussian likelihoods have significant computational advantages, and there is no guarantee that constructing a task-specific likelihood for every optimization problem would be worth the effort.
We propose accepting that some assumption error will always be present, and instead focus on how alter BayesOpt to accomodate imperfect models.

\subsection{Conformal prediction}
\label{subsec:conformal_prediction_background}

See \citet{shafer2008tutorial} for a complete tutorial on conformal prediction, or \citet{angelopoulos2021gentle} for a modern, accessible introduction.
Informally, a conformal prediction set $C_\alpha(\vec x_n) \subset \mathcal{Y}$ is a set of possible labels for a test point $\vec x_n$ given a sequence of observations $\mathcal{D}$. 
Candidate labels $\hat{\vec y}_n$ are included in $C_\alpha(\vec x_n)$ if the resulting pair $(\vec x_n, \hat{\vec y}_n)$ is sufficiently similar to the actual examples in $\mathcal{D}$.
The degree of similarity is measured by a score function $s$ and importance weights (IWs) $\vec w$, and the similarity threshold is determined by the miscoverage tolerance $\alpha$.
In Figure \ref{fig:construct_conf_pred_set_example} we visualize the process of constructing $C_\alpha(\vec x_n)$.

Conformal prediction is a  \textit{distribution-free} uncertainty quantification method because it does not assume $\mathcal{D}$ is drawn from any particular distribution, nor does it assume $s$ is derived from a well-specified model.
In our context the critical assumption is that $\mathcal{D} \cup \{(\vec x_n, \vec y_n)\}$ is pseudo-exchangeable.
\citet{fannjiang2022conformal} provide a formal statement of pseudo-exchangeability (Definition 2), and prove a coverage guarantee for conformal prediction sets in the special case when $\mathcal{D}$ is IID from $p(\vec x)p^*(\vec y | \vec x)$ and $p(\vec x | \mathcal{D})$ is invariant to shuffling of $\mathcal{D}$, which we restate below:

\begin{definition}
Let $\mathcal{D} \sim p(\vec x)p^*(\vec y | \vec x)$ and $(\vec x_n, \vec y_n) \sim p'(\vec x | \mathcal{D})p^*(\vec y | \vec x)$, where $p'(\vec x | \mathcal{D})$ is chosen such that $\mathcal{D} \cup \{(\vec x_n, \vec y_n)\}$ is pseudo-exchangeable.
Given $w_i \propto p'(\vec x_i | \hat{\mathcal{D}}_{-i}) / p(\vec x_i)$ s.t. $\sum_i w_i = 1$,
$\forall \alpha \in (0, 1]$, the conformal prediction set corresponding to score function $s$ is
\begin{align}
    \mathcal{C}_\alpha(\vec x_n) &:= \left\{ \hat{\vec y}_n \in \mathcal{Y}
    \; \bigg| \; \vec h^\top \vec w > \alpha
    \right\}, \label{eq:conf_pred_defn} \\
    \text{where } h_i &:= \mathds{1}\left\{s(\vec x_i, \vec y_i, \hat{\mathcal{D}}) \leq s(\vec x_n, \hat{\vec y}_n, \hat{\mathcal{D}})\right\}, \nonumber
\end{align}
\label{def:conf_bayes_pred_set}
\end{definition}
\vspace{-6mm}
$\hat{\mathcal{D}} = \mathcal{D} \cup \{(\vec x_n, \hat{\vec y}_n)\}$, and $p'(\vec x | \hat{\mathcal{D}}_{-i})$ is the query proposal density given training data $\hat{\mathcal{D}}_{-i} = \hat{\mathcal{D}} - \{(\vec x_i, \vec y_i)\}$.
The importance weights $\vec w$ account for covariate shift \citep{tibshirani2019conformal},
and $w_i = 1 / (n + 1) \; \forall i$ in the special case where $\mathcal{D} \cup \{(\vec x_n, \vec y_n)\}$ is fully exchangeable (e.g. IID).

Conformal prediction enjoys a frequentist marginal coverage guarantee on $\mathcal{C}_\alpha(\vec x_n)$ with respect to the joint distribution over $\mathcal{D} \cup \{(\vec x_n, \vec y_n)\}$,
\begin{align}
    \mathbb{P}[\vec y_n \in \mathcal{C}_\alpha(\vec x_n) ] \geq 1 - \alpha, \label{eq:marg_coverage_defn}
\end{align}
meaning if we repeatedly draw $\mathcal{D} \sim p(\vec x)p^*(\vec y | \vec x)$, and $(\vec x_n, \vec y_n) \sim p'(\vec x | \mathcal{D})p^*(\vec y | \vec x)$, $\mathcal{C}_\alpha(\vec x_n)$ will contain the observed label $\vec y_n$ with frequency at least $(1 - \alpha)$.
A prediction set with a coverage guarantee like Eq. \eqref{eq:marg_coverage_defn} is \textit{conservatively valid} at the $1 - \alpha$ level.
In Appendix \ref{subsec:randomized_conf_pred} we discuss \textit{randomized} conformal prediction which is \textit{exactly valid}, meaning the long run frequency of errors converges to exactly $\alpha$.
Marginal coverage is distinct from \textit{conditional} coverage, since it does not guarantee the coverage of $C_\alpha$ for any specific $\vec x_n$, only the average coverage over the whole domain $\mathcal{X}$.

\textbf{Full conformal Bayes} corresponds to the score function $s(\vec x_i, \vec y_i, \hat{\mathcal{D}}) = \log p(\vec y_i | \vec x_i, \hat{\mathcal{D}})$.
Conditioning an existing posterior $p(\vec y | \vec x_i, \mathcal{D})$ on the additional observation $(\vec x_n, \hat{\vec y}_n)$ is commonly referred to as ``retraining'' in the conformal prediction literature.
\textit{If} the surrogate just so happens to be correctly specified (e.g. $f^* \sim p(f)$), then $\log p(\vec y_i | \vec x_i, \hat{\mathcal{D}})$ is the optimal choice of score function, meaning it provides the most \textit{efficient} prediction sets (i.e. smallest by expected volume w.r.t. the prior $p(f)$) among all prediction sets that are valid at the $1 - \alpha$ level \citep{hoff2021bayes}.
In the typical situation where we think our model assumptions are plausible but do not really believe them, full conformal Bayes rewards us if our assumptions turn out to be correct, yet it produces valid predictions as long as the true data generation process is some pseudo-exchangeable sequence.

\textbf{BayesOpt and pseudo-exchangeability:} unfortunately if $\mathcal{D}$ is collected by some active online selection strategy such as BayesOpt, then $\mathcal{D}$ is not IID and the coverage guarantee for Defn. \ref{def:conf_bayes_pred_set} does not apply.
Note that even large offline datasets are not guaranteed to be IID, so the same issue may arise even in single-round design setting considered by \citet{fannjiang2022conformal}.
Despite the gap in theory, in this work we investigate the technical challenges associated with incorporating conformal prediction sets into BayesOpt, and find that empirically they can still improve query coverage (Section \ref{subsec:single_obj_results}).
In Appendix \ref{subsec:assumptions} we include further discussion of the assumptions and limitations of conformal prediction.

\section{RELATED WORK}
\label{sec:related_work}

\textbf{Conformal prediction:} 
Our work is related to \citet{fannjiang2022conformal}, who propose a black-box optimization method based on conformal prediction specifically to address feedback covariate shift.
However, because they assume new queries are drawn from a closed-form proposal distribution, and because exact conformal prediction is not differentiable, their approach cannot be easily extended to most BayesOpt methods.\footnote{BayesOpt proposal distributions are usually implicit, obtained through gradient-based optimization of the acquisition function.}
\citet{bai2022efficient} propose a differentiable approximation of conformal prediction, but it requires solving a minimax optimization subproblem.
\citet{stutz2021learning} independently proposed a continuous relaxation of conformal prediction, like our work, but only for fully exchangeable classification data.
We propose a more general form that allows for covariate shift, and we also provide an efficient discretization procedure for regression and show how to estimate the importance weights when the queries are drawn from an implicit density.

\textbf{Robust BayesOpt:}
There is a substantial body of work on adaptation to model misspecification in the bandit setting (i.e. discrete actions), e.g. \citet{lattimore2020learning} and \citet{foster2020adapting}, however we are primarily focused on problems with continuous decisions.
Since the seminal analysis of GP-UCB regret bounds by \citet{srinivas2010gaussian}, follow-up work has proposed UCB variants for misspecified likelihoods \citep{makarova2021}, misspecified GP priors \citep{bogunovic2021misspecified}, or to guarantee $f^*(\vec x_i) > c$ for some threshold $c \in \mathbb{R}$ \citep{sui2015safe}.
These approaches are not easy to extend to other acquisition functions, and tend to rely on fairly strong assumptions on the smoothness of $f^*$ or fix a specific kind of model misspecification.\footnote{For example, it is commonly assumed that $f^*$ has bounded RKHS norm w.r.t. the chosen GP kernel, that we know a good bound in order to set hyperparameters correctly, and that $f^*$ is Lipschitz continuous.}
\citet{wang2018regret} prove regret bounds for GP-UCB when $f^*$ is drawn from a GP with unknown mean and kernel functions, but assume we know the right hypothesis class of mean and kernel functions and have a collection of offline datasets available for pretraining.

Finally, \citet{eriksson2019scalable} propose TuRBO, which is superficially similar to conformal BayesOpt since it constrains queries to a Latin hypercube trust region around the best known local optimum.
While TuRBO can be very effective in practice, the size of the trust region is controlled by a heuristic with five hyperparameters in the single-objective case, and even more in the multi-objective case \citep{daulton2022multi}.
Despite the additional complexity, the credible set coverage on queries in TuRBO trust regions can still vary wildly (see Section \ref{subsec:single_obj_results}).
In contrast, conformal prediction provides distribution-free coverage guarantees under very mild assumptions, and our approach can be applied to any reparameterizable acquisition function \citep{wilson2017reparameterization}.
To our knowledge our approach is the first BayesOpt procedure to incorporate conformal prediction.

\section{METHOD}
\label{sec:conformal_bayesopt}

We now describe the key ideas behind conformal Bayesian optimization.
First in Section \ref{subsec:full_conformal_gps} we show how to efficiently compute $C_\alpha(\vec x_n)$, addressing differentiability and discretization of continuous outcomes.
Our procedure is summarized in Algorithm \ref{alg:differentiable_conformal_masks}.
In Section \ref{subsec:conf_acq_fns} we introduce the conformal Bayes posterior $p_\alpha(f(\vec x_n) | \mathcal{D})$, allowing us to distinguish between aleatoric and epistemic uncertainty and to combine conformal prediction with many well-known BayesOpt utility functions.
Finally in Section \ref{subsec:handling_covariate_shift} we address feedback covariate shift without requiring closed-form expressions for $p(\vec x)$ and $p'(\vec x_i | \hat{\mathcal{D}}_{-i})$.
In Appendix \ref{subsec:method_summary} we provide a detailed overview of the whole method, along with a discussion of the computational complexity in Appendix \ref{subsec:complexity}.

\begin{algorithm}[t]
\caption{Differentiable conformal prediction masks}
\label{alg:differentiable_conformal_masks}
\SetAlgoLined
    \KwData{train data $\mathcal{D} = \{(\vec x_i, \vec y_i)\}_{i=0}^{n-1}$, test point $\vec x_n$, imp. weights $\vec w$, label candidates $Y_{\mathrm{cand}}$, score function $s$, miscoverage tolerance $\alpha$, relaxation strength $\tau > 0$.}
      $m_j = 0$, $\forall j \in \{0, \dots, k - 1\}$. 
      
      \For{$\hat{\vec y}_j \in Y_{\mathrm{cand}}$}{
        $\hat{\mathcal{D}} \leftarrow \mathcal{D} \cup \{(\vec x_n, \hat{\vec y}_j)\}$
        
        $\vec s \leftarrow [s(\vec x_0, \vec y_0, \hat{\mathcal{D}}) \; \cdots \; s(\vec x_n, \hat{\vec y}_j, \hat{\mathcal{D}})]^\top$. 
        
        $\vec h \leftarrow \texttt{sigmoid}(\tau^{-1}(\vec s - s_n ))$. 
        
        $\overline{w} \leftarrow \vec h^\top \vec w$. 
        
        $m_j \leftarrow \texttt{sigmoid}(\tau^{-1} (\overline{w} - \alpha))$.
      }
    \KwResult{$\vec m$}
\end{algorithm}

\begin{figure*}
    \centering
    \captionsetup[subfigure]{justification=centering}
    \begin{subfigure}{0.3\textwidth}
        \centering
        \includegraphics[width=\textwidth,trim=1cm 1.35cm 0.25cm 1.2cm]{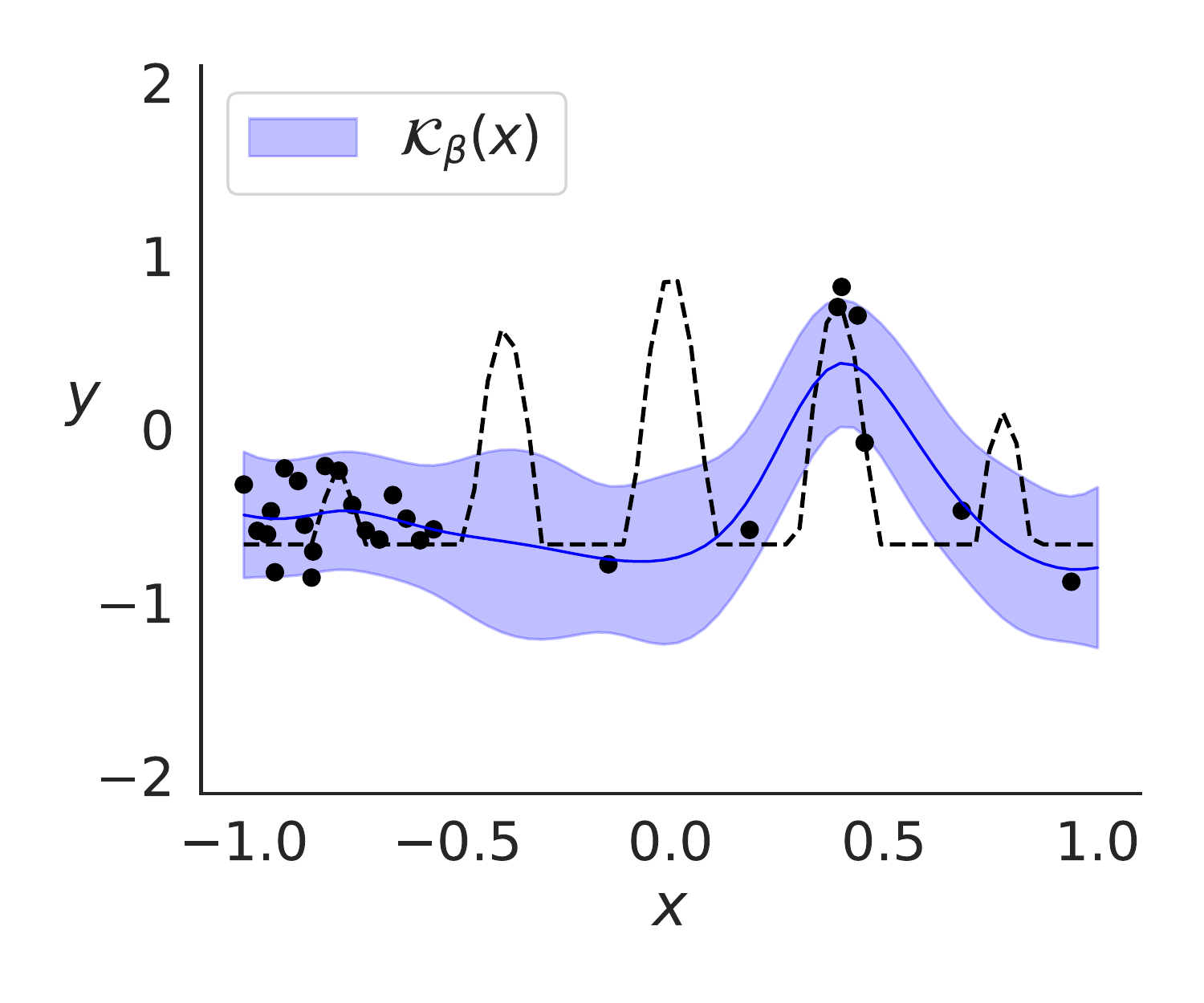}
        \caption{}
        \label{fig:synth-example_cred-set}
    \end{subfigure}
    \hfill
    \begin{subfigure}{0.3\textwidth}
        \centering
        \includegraphics[width=\textwidth,trim=1cm 1.35cm 0.25cm 1.2cm]{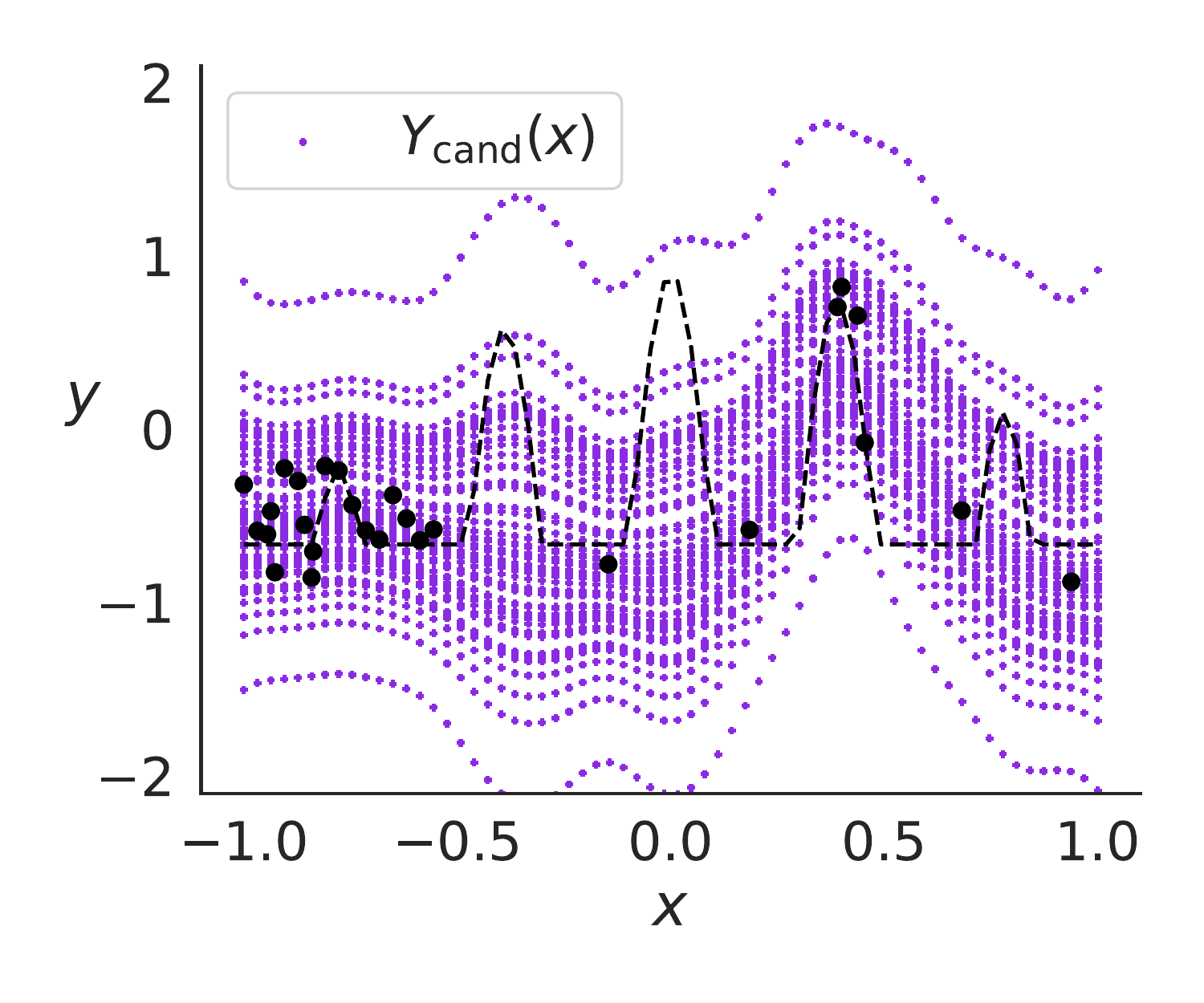}
        \caption{}
        \label{fig:synth-example_conf-grid}
    \end{subfigure}
    \hfill
    \begin{subfigure}{0.3\textwidth}
        \centering
        \includegraphics[width=\textwidth,trim=0.4cm 1cm 0.25cm 1.2cm]{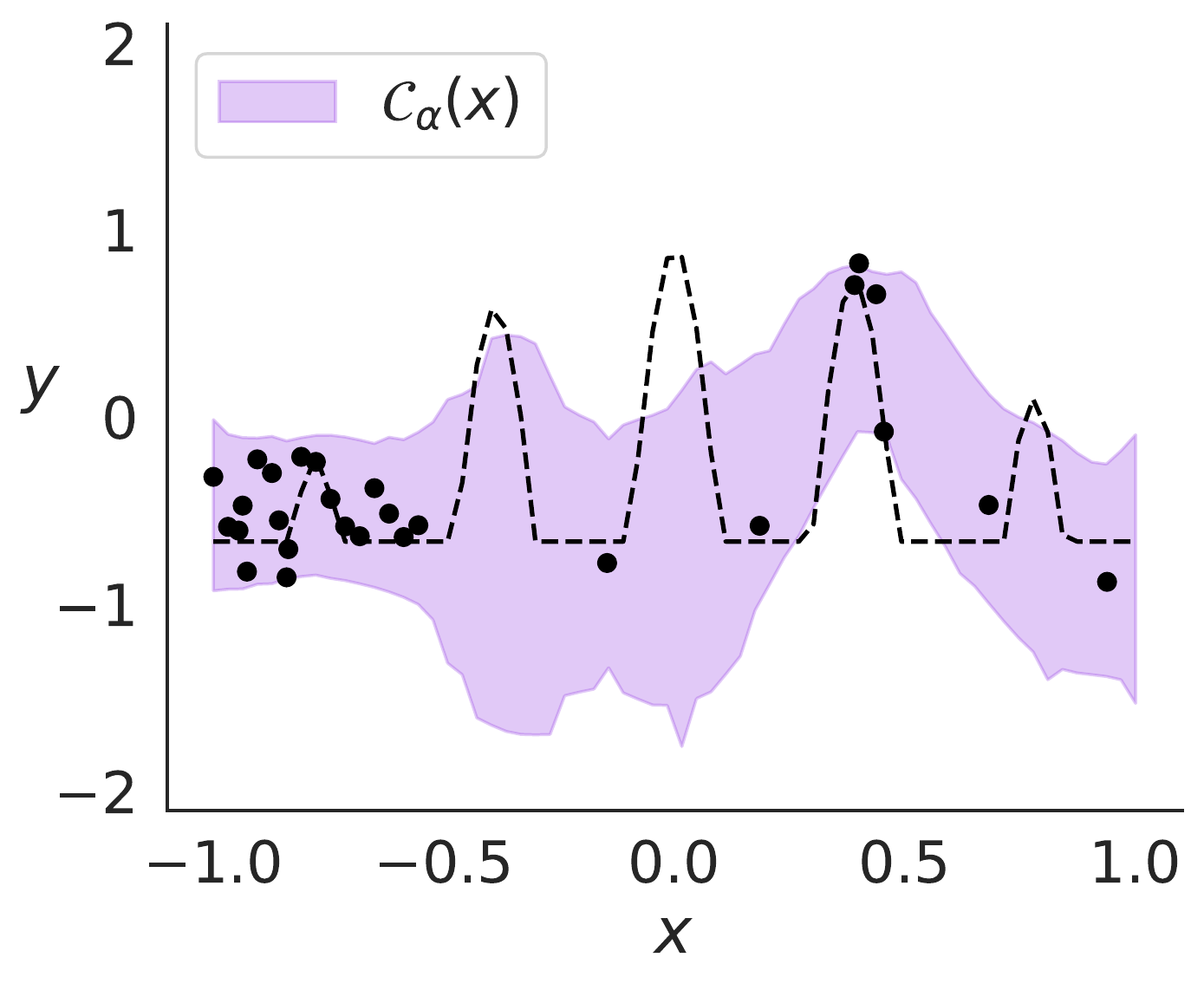}
        \caption{}
        \label{fig:synth-example_conf-set}
    \end{subfigure}
    \caption[Example of conformal regression prediction]{
    Constructing full conformal Bayes prediction sets starting with a Bayes posterior $p(\hat{\vec y}| \vec x, \mathcal{D})$.
    In this example $\mathcal{D}$ is composed of $n = 27$ noisy observations (shown as black dots) of the true objective (shown as a black dashed line) and $\alpha = 1 - \beta = 0.19$.
    In panel \textbf{(a)} we show $\mathcal{K}_\beta(\vec x)$, the $\beta$-credible prediction set.
    In panel \textbf{(b)} we show $Y_{\mathrm{cand}}$ populated by samples from $p(\hat{\vec y} | \vec x, \mathcal{D})$.
    In panel \textbf{(c)} we show $\mathcal{C}_\alpha(\vec x)$, the conformal prediction set.
    The coverage of $\mathcal{C}_\alpha(\vec x)$ is noticeably better than $\mathcal{K}_\beta(\vec x)$ in regions where there is little training data, though the nominal confidence level is the same.
    }
    \label{fig:conf_pred_discretization}
\end{figure*}

\subsection{Full conformal Bayes with Gaussian processes}
\label{subsec:full_conformal_gps}

\textbf{Efficient retraining:} full conformal Bayes requires us to compute $\log p(\vec y_i | \vec x_i, \hat{\mathcal{D}}) \; \forall i \leq n$ and $\forall \hat{\vec y}_j \in Y_{\mathrm{cand}}$, where $Y_{\mathrm{cand}}$ is some discretization of $\mathcal{Y}$.
This incremental posterior update can be done very efficiently if the surrogate is a GP regression model \citep{gardner2018gpytorch, stanton2021kernel, maddox2021conditioning}, and we will later reuse the conditioned posteriors to estimate expectations w.r.t. $p_\alpha(f(\vec x) | \mathcal{D})$.
Note that computing the GP posterior likelihood of training data can be numerically unstable, which we address in Appendix \ref{app:imp_details}.
Other Bayesian predictive posteriors (e.g. from Bayesian neural networks) are conditioned on training data via iterative methods such as gradient descent, making full conformal Bayes very expensive \citep{fong2021conformal}. 

\textbf{Differentiable prediction masks:} 
the definition of $\mathcal{C}_\alpha(\vec x_n)$ in Eq. \eqref{eq:conf_pred_defn} can be broken down into a sequence of simple vector operations interspersed with Heaviside functions.
The Heaviside function is piecewise constant, with ill-defined derivatives, so we replace it with its continuous relaxation, the \texttt{sigmoid} function (Algorithm \ref{alg:differentiable_conformal_masks}).
Informally, the output $m_j$ of the final sigmoid can be interpreted as the \textit{probabilility} of accepting some $\hat{\vec y}_j$ into $\mathcal{C}_\alpha(\vec x_n)$.
The smoothness of the relaxation is controlled by a single hyperparameter $\tau \in (0, +\infty)$.
As $\tau \rightarrow 0$ the relaxation becomes exact but the gradients become very poorly behaved.

\textbf{Efficient discretization of $\mathcal{Y}$:} now we need a good way to choose $Y_{\mathrm{cand}}$.
When $\vec y$ is low-dimensional (e.g. sequential, single-objective tasks), then $Y_{\mathrm{cand}}$ can be a dense grid, however dense grids are inefficient since they must be wide enough to capture all possible values of $\vec y$ and dense enough to pinpoint the boundary of $\mathcal{C}_\alpha(\vec x_n)$.
Even if we do not fully believe $p(\vec y | \vec x_n, \mathcal{D})$, it is still our best guess of where $\vec y | \vec x_n$ should be, so instead of a dense grid we populate $Y_{\mathrm{cand}}$ with proposals $\hat{\vec y}_j \sim p(\vec y | \vec x_n, \mathcal{D})$.
This approach not only reduces computational effort for low-dimensional $\vec y$, it also allows us to extend to tasks with multiple objectives and batched queries (Appendix \ref{subsec:practical_extensions}).
In Figure \ref{fig:conf_pred_discretization} we visualize the computation of a conformal Bayes prediction set.

\subsection{Conformal acquisition functions}
\label{subsec:conf_acq_fns}

For the sake of clarity in the following sections we will omit the subscript from $\vec x_n$.
By the sum rule of probability, we can rewrite $p(f(\vec x) | \mathcal{D})$ as an integral over all possible outcomes $\vec y | \vec x$,
\begin{align}
    p(f(\vec x) | \mathcal{D}) &= \int_{\hat{\vec y} \in \mathcal{Y}} p(f(\vec x) | \hat{\mathcal{D}})p(\hat{\vec y} | \vec x, \mathcal{D})d\hat{\vec y} \label{eq:f_bma}.
\end{align}
In other words, $p(f(\vec x) | \mathcal{D})$ can be seen as a Bayesian model average, where we condition each component model on a different potential observation $(\vec x, \hat{\vec y})$, and weight the components by $p(\hat{\vec y} | \vec x, \mathcal{D})$.

\begin{figure}[t]
    \centering
    \includegraphics[height=2.7cm]{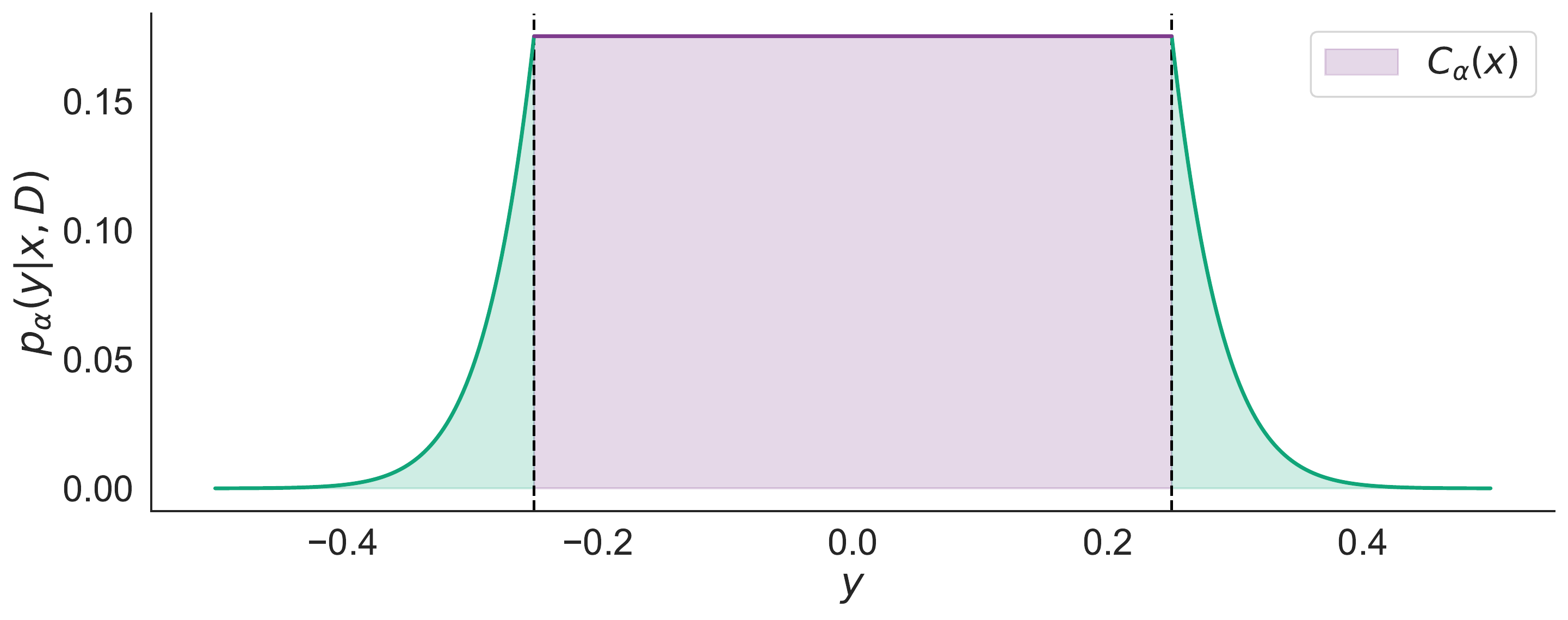}
    \caption[Conformal predictive posterior illustration]{
    An illustration of $p_\alpha(\hat{\vec y} | \vec x, \mathcal{D})$.
    The total density of all outcomes in $\mathcal{C}_\alpha(\vec x)$ is set to $1-\alpha$.
    }
    \label{fig:conf_bo_posterior}
\end{figure}

We are free to change the component weights to any other valid distribution over $\hat{\vec y}$ we like.
Now we introduce the conformal Bayes predictive posterior $p_\alpha(\hat{\vec y} | \vec x, \mathcal{D})$,
\begin{align*}
    p_\alpha(\hat{\vec y} | \vec x, \mathcal{D}) := 
    \begin{cases}
        (1 - \alpha) / Z_1 & \text{ if } \hat{\vec y} \in C_{\alpha}(\vec x), \\
        \alpha p(\hat{\vec y} | \vec x, \mathcal{D}) / Z_2 & \text{else,}
    \end{cases}
\end{align*}
where $Z_1, Z_2$ are normalization constants.
See Figure \ref{fig:conf_bo_posterior} for an illustration.
We are partitioning the outcome space into two events, either $\hat{\vec y} \in C_\alpha(\vec x)$ or it is not.
Since $C_\alpha(\vec x)$ is a valid prediction set, $\hat{\vec y} \in C_\alpha(\vec x)$ with frequency $(1 - \alpha)$, and we do not consider any particular $\hat{\vec y} \in C_\alpha(\vec x)$ to be more likely than another, since the coverage guarantee holds for $C_\alpha(\vec x)$ as a whole.\footnote{
We could also use a conformal predictive density here \citep{vovk2017nonparametric, marx2022modular}, which we leave for future work.
}
We also expect that $\hat{\vec y} \in \mathcal{Y}\setminus C_\alpha(\vec x)$ with frequency $\alpha$, and we weight each $\hat{\vec y} \notin C_\alpha(\vec x)$ by $p(\hat{\vec y} | \vec x, \mathcal{D})$ to form a proper density (i.e. a density that integrates to 1).

If we had noiseless observations (i.e. $\vec y_i = f(\vec x_i))$, we could use $p_\alpha(\hat{\vec y} | \vec x, \mathcal{D})$ directly when computing the acquisition value of new queries.
However managing the explore-exploit tradeoff with noisy outcomes requires us to distinguish between epistemic and aleatoric uncertainty.
If we do not, optimistic acquisition functions like UCB may direct us towards queries whose outcomes are uncertain due to measurement error.
Substituting $p_\alpha(\hat{\vec y} | \vec x, \mathcal{D})$ for $p(\hat{\vec y} | \vec x, \mathcal{D})$ in Eq. \eqref{eq:f_bma} results in the conformal Bayes posterior $p_\alpha(f(\vec x) | \mathcal{D})$,
\begin{align}
    p_\alpha(f(\vec x) | \mathcal{D}) &:= \frac{1 - \alpha}{Z_1} \int\limits_{\hat{\vec y} \in C_\alpha(\vec x)} p(f | \hat{\mathcal{D}} )d\hat{\vec y} \label{eq:conf_bayes_posterior} \\
    & + \frac{\alpha}{Z_2} \int\limits_{\hat{\vec y} \in \mathcal{Y} \setminus C_\alpha(\vec x)} p(f(\vec x)| \hat{\mathcal{D}})p(\hat{\vec y} | \vec x, \mathcal{D})d\hat{\vec y}. \nonumber 
\end{align}
Given $p_\alpha(f| \mathcal{D})$, we can "conformalize" any acquisition function written in the form of Eq. \eqref{eq:acq_fn_general_form} by substituting $p_\alpha(f| \mathcal{D})$ for $p(f | \mathcal{D})$.
In Appendix \ref{subsec:conf_bayes_posterior_details} we show that $p_\alpha(f | \mathcal{D})$ converges pointwise to $p(f | \mathcal{D})$ as $\alpha \rightarrow 1$, and in Appendix \ref{subsec:more_conf_acqs} we explicitly derive conformal variants of several popular BayesOpt acquisition functions.

\begin{figure*}[t]
    \centering
    \captionsetup[subfigure]{justification=centering}
    \begin{subfigure}{0.2\textwidth}
        \includegraphics[width=\textwidth]{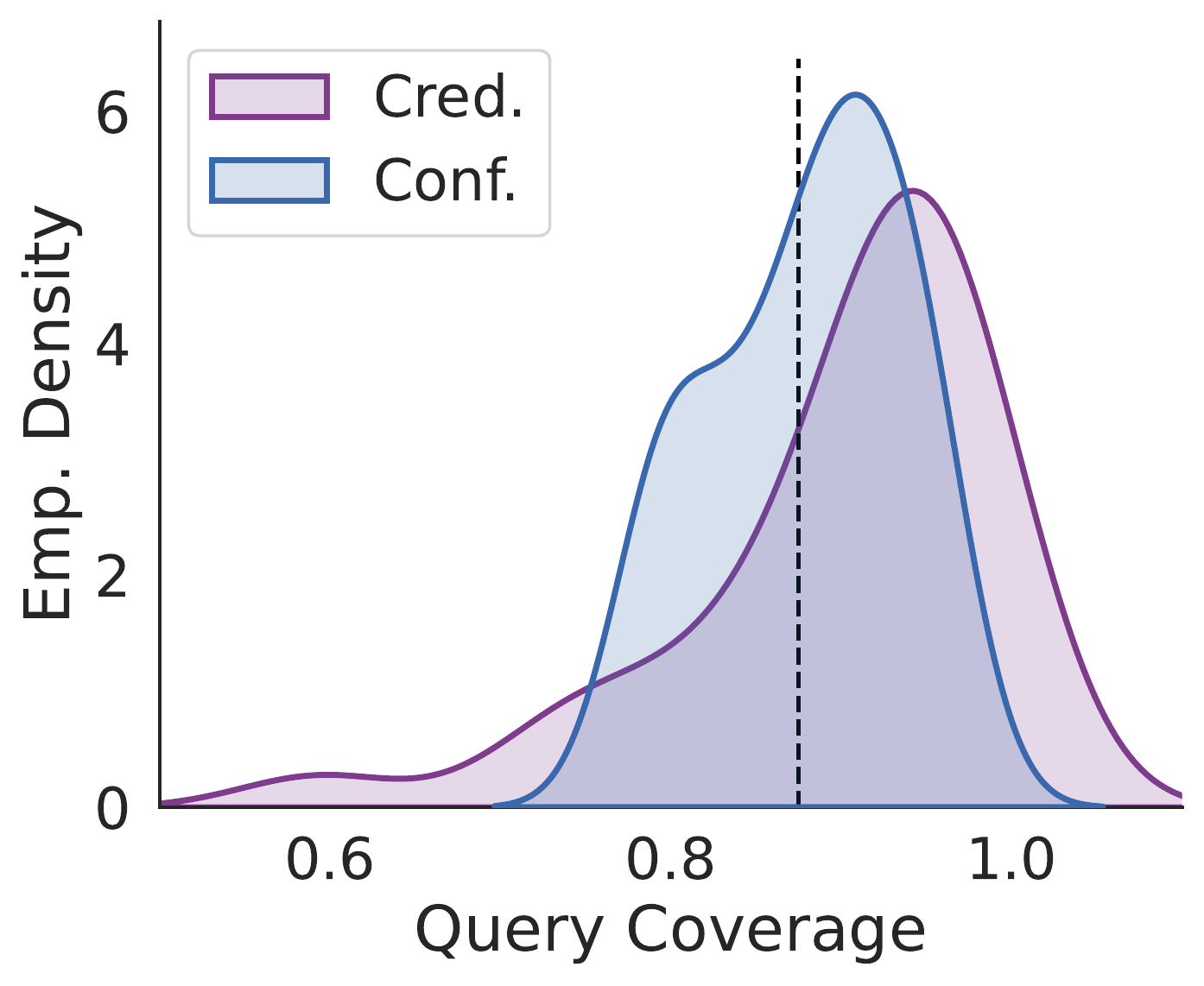}
        \caption{}
        \label{subfig:conf_cred_cvrg_comparison}
    \end{subfigure}
    \hfill
    \begin{subfigure}{0.2\textwidth}
        \includegraphics[width=\textwidth]{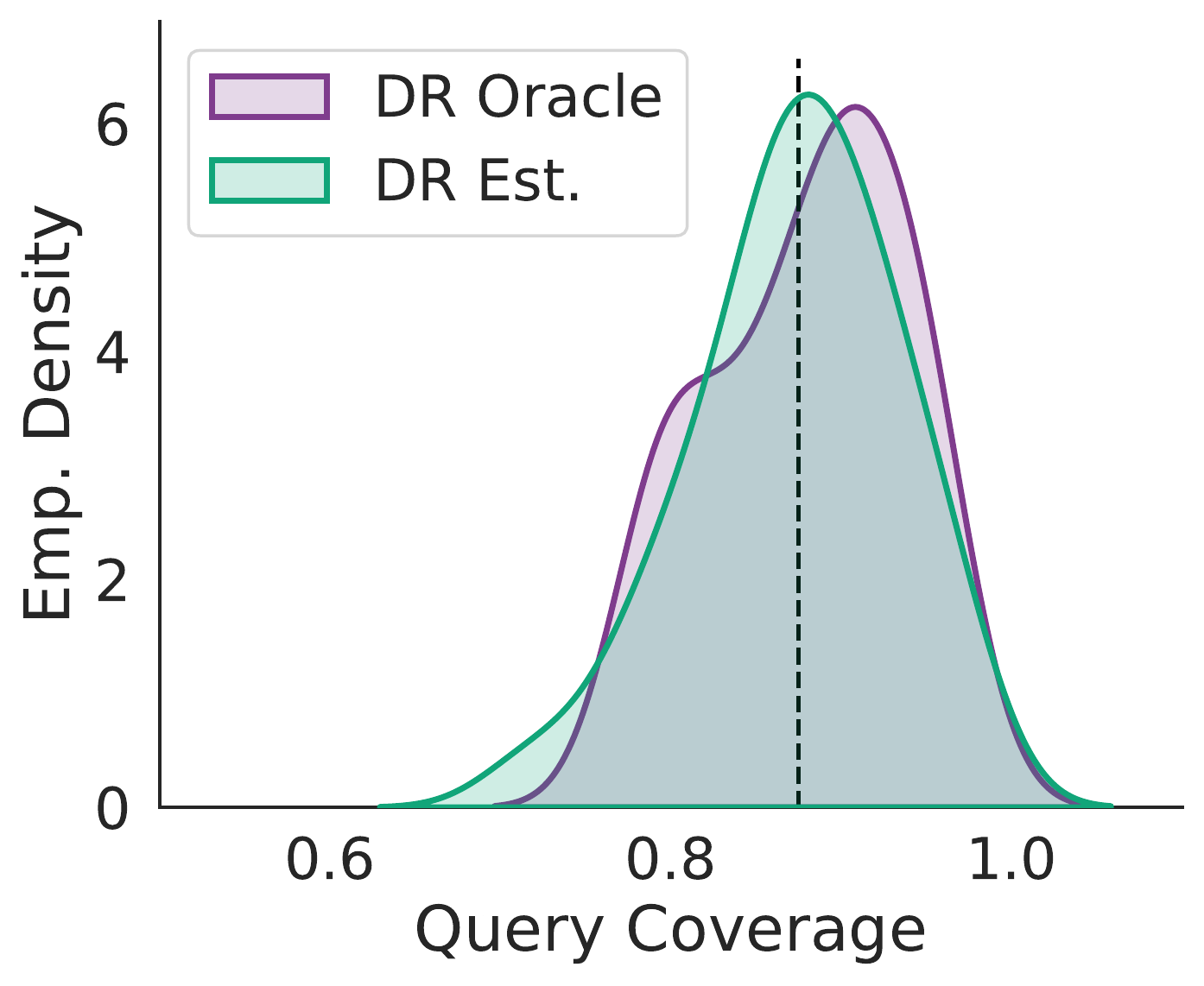}
        \caption{}
    \end{subfigure}
    \hfill
    \begin{subfigure}{0.2\textwidth}
        \includegraphics[width=\textwidth]{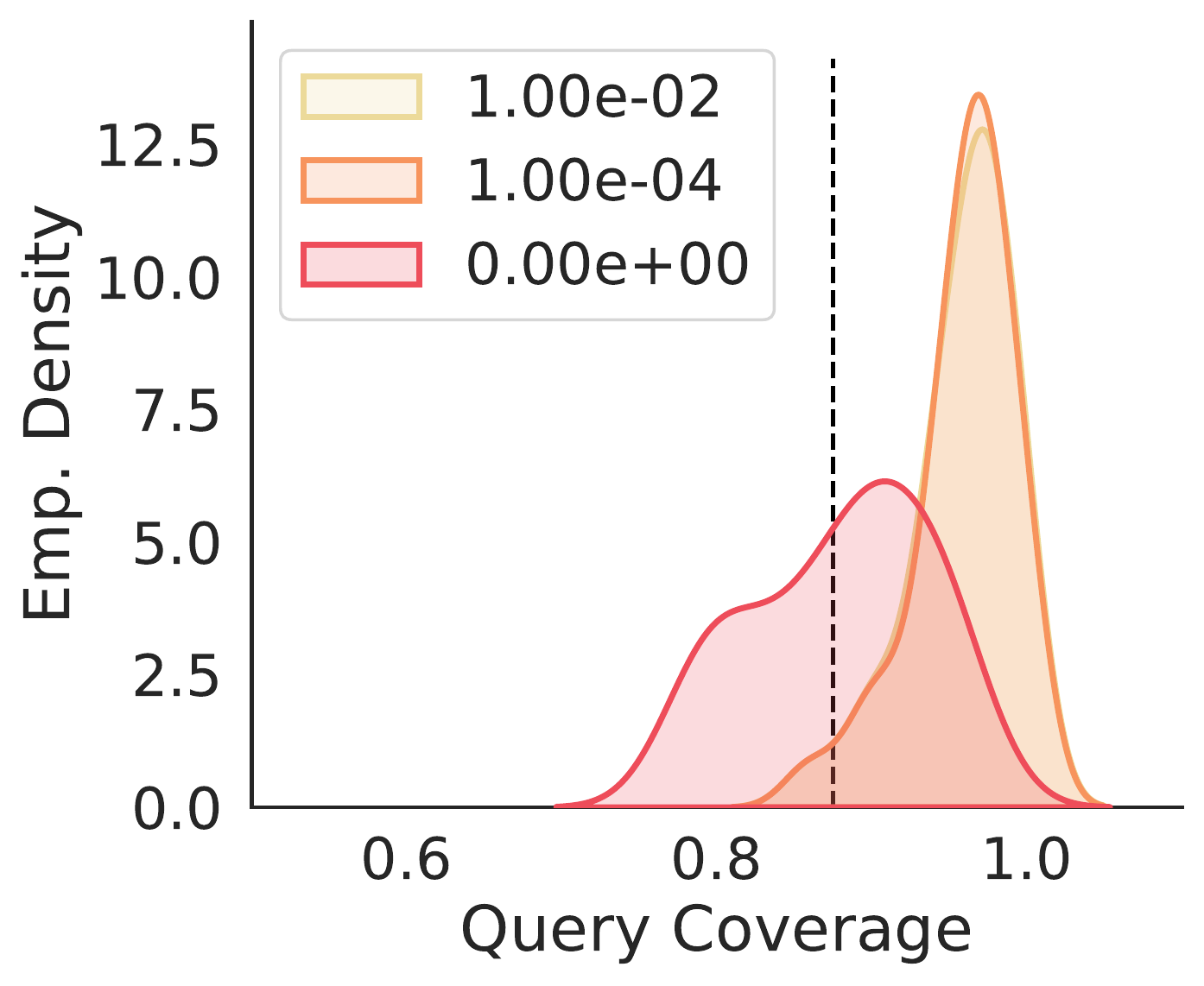}
        \caption{}
    \end{subfigure}
    \hfill
    \begin{subfigure}{0.2\textwidth}
        \includegraphics[width=\textwidth]{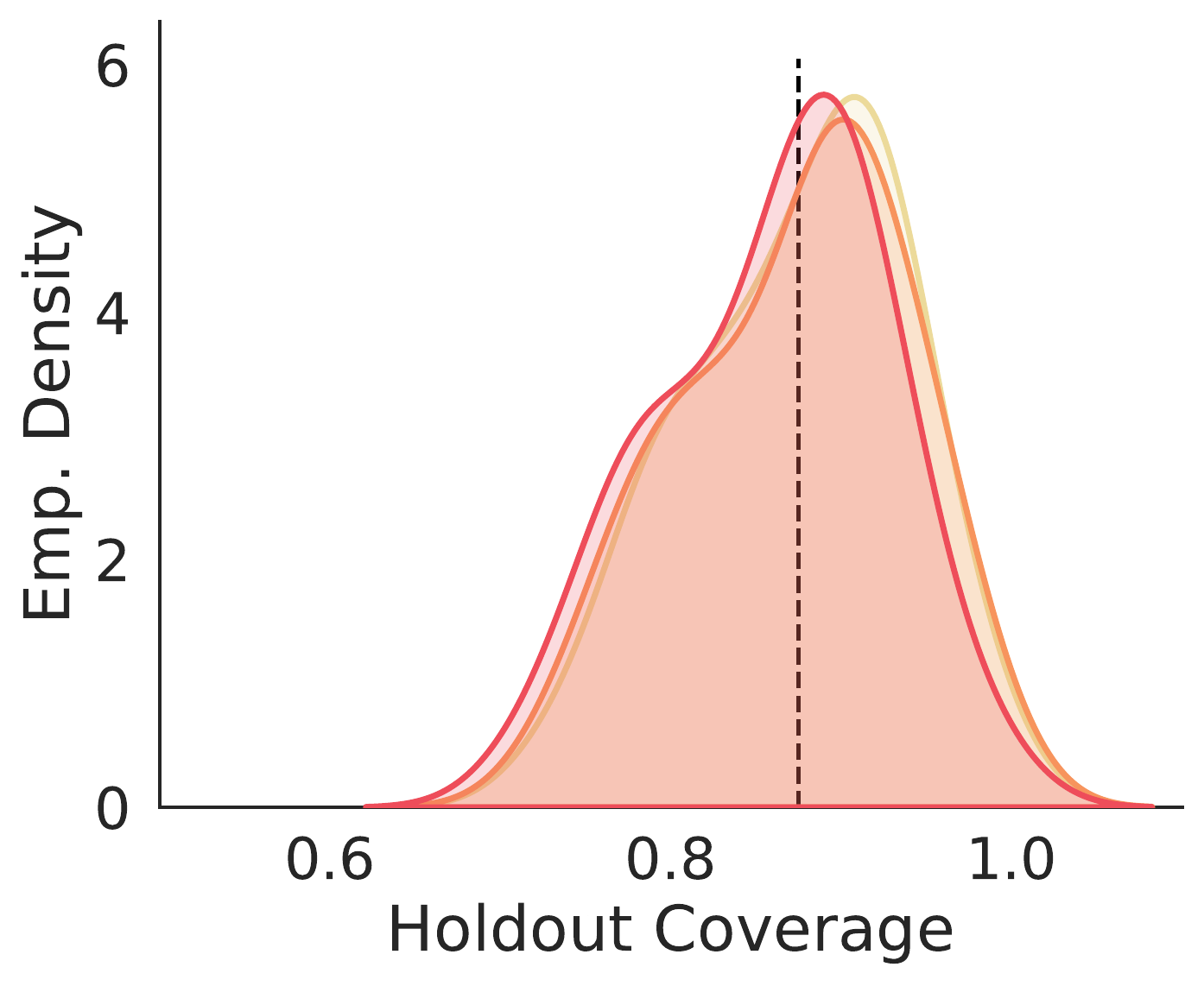}
        \caption{}
    \end{subfigure}
    \caption[Coverage of approximate conformal prediction]{
    Here we evaluate the empirical coverage of $C_\alpha(\vec x)$.
    The shaded regions in each panel depict a KDE estimate of the distribution of coverage when $n=64$ and $\alpha=0.125$, estimated from 32 independent trials.
    The black dashed line indicates $1 - \alpha$.
    In panel \textbf{(a)} we compare the coverage of Bayesian $\beta$-credible ($\beta = 1 - \alpha$) and randomized conformal prediction sets, if $\tau = 0$ and we have a density ratio oracle.
    Conformal prediction provides much more consistent coverage.
    Next in panel \textbf{(b)} replacing the density ratio oracle with learned density ratio estimates has fairly minimal effect on the coverage of the resulting conformal prediction sets.
    In panel \textbf{(c)} we investigate the effect of the sigmoid temperature $\tau$ when $p'(\vec x | \mathcal{D}) \neq p(\vec x)$.
    In panel \textbf{(d)} we investigate the effect of $\tau$ when $p'(\vec x | \mathcal{D}) = p(\vec x)$.
    Increasing $\tau$ makes the prediction sets more conservative.
    }
    \label{fig:approx_error_analysis}
\end{figure*}

\textbf{Monte Carlo estimates of conformal acquisition values:} 
in brief, given a query point $\vec x$ and utility function $u$, we first draw a candidate grid $Y_{\mathrm{cand}}$ and compute the corresponding prediction mask $\vec m$ according to Section \ref{subsec:full_conformal_gps}. Then we estimate the conformal acquisition value as follows:
\begin{align}
a_\alpha(\vec x, \mathcal{D}) &= \displaystyle \int u(\vec x, f, \mathcal{D})p_\alpha(f | \mathcal{D})df, \label{eq:conf_acq_mc_est} \\
& \approx (1 - \alpha) \vec u^\top \vec v + \alpha \vec u^\top \vec v', \nonumber \\
\text{where } \vec u &= [u(\vec x, f^{(0)}, \mathcal{D}) \; \cdots \; u(\vec x, f^{(k-1)}, \mathcal{D})]^\top, \nonumber \\
\vec v_i &= \frac{m_i}{p(\hat{\vec y}_i | \vec x, \mathcal{D})} \left (\sum_j \frac{m_j}{p(\hat{\vec y}_j | \vec x, \mathcal{D})}\right )^{-1}, \nonumber \\
\vec v'_i &= (1 - m_i)(\vec 1^\top (\vec 1 - \vec m))^{-1}, \nonumber
\end{align}
and $f^{(j)} \sim p(f | \mathcal{D} \cup \{(\vec x, \hat{\vec y}_j)\}) \; \forall \hat{\vec y}_j \in Y_{\mathrm{cand}}$, which is cheap since we already computed the conditioned posteriors when calculating $\vec m$.
See Appendix \ref{subsec:mc_integration_details} for the full derivation.

\subsection{Accounting for Feedback Covariate Shift}
\label{subsec:handling_covariate_shift}

If we were merely ranking queries exchangeable with $\mathcal{D}$, then there would be no need to correct for covariate shift. 
However, our goal is to find queries with exceptional outcomes, and the more we optimize, the more severe we can expect the resulting feedback covariate shift to be.

\textbf{Density ratio estimation:} as we saw in Section \ref{subsec:conformal_prediction_background}, adapting $\mathcal{C}_\alpha(\vec x)$ to covariate shift requires estimating importance weights $w_i \propto r(\vec x_i) = p'(\vec x_i | \hat{\mathcal{D}}_{-i})/p(\vec x_i)$,
where $p'(\vec x_i | \hat{\mathcal{D}}_{-i})$ is the proposal distribution from which we \textit{would} have drawn candidate query points if we had training data $\hat{\mathcal{D}}_{-i}$.
If we have closed-form expressions for $p(\vec x_i)$ and $p'(\vec x_i | \hat{\mathcal{D}}_{-i})$ then we can compute $r(\vec x_i)$ easily, but in general we only have samples from $p(\vec x)$.
Furthermore if we wish to optimize queries with gradient based methods then $p'(\vec x_i | \hat{\mathcal{D}}_{-i})$ is implicitly defined as the distribution over iterates $\vec x_n^{(t)}$ induced by the gradient field $\nabla_{\vec x}a_\alpha$ and an initial distribution on $\vec x_n^{(0)}$.
Fortunately we can still obtain samples from $p'(\vec x_i | \hat{\mathcal{D}}_{-i})$ by sampling from the energy distribution, $p'(\vec x_i | \hat{\mathcal{D}}_{-i}) \propto \exp\{a_\alpha(\vec x, \hat{\mathcal{D}}_{-i})\}$ via stochastic gradient Langevin dynamics (SGLD) \citep{welling2011bayesian}.
Note that this formulation requires us to run $(n + 1) \times k$ SGLD chains, one for each density $p'(\vec x_i | \hat{\mathcal{D}}_{-i})$.
Since we are already intending to use a sample-based empirical approximation of the density ratio, we make another approximation here, assuming $a_\alpha(\vec x, \hat{\mathcal{D}}_{-i}) \approx a_\alpha(\vec x, \mathcal{D})$, which allows us to rely on samples from a single SGLD chain.

\begin{figure*}
    \centering
    \captionsetup[subfigure]{justification=centering}
    \begin{subfigure}{0.3\textwidth}
        \includegraphics[height=2.25cm]{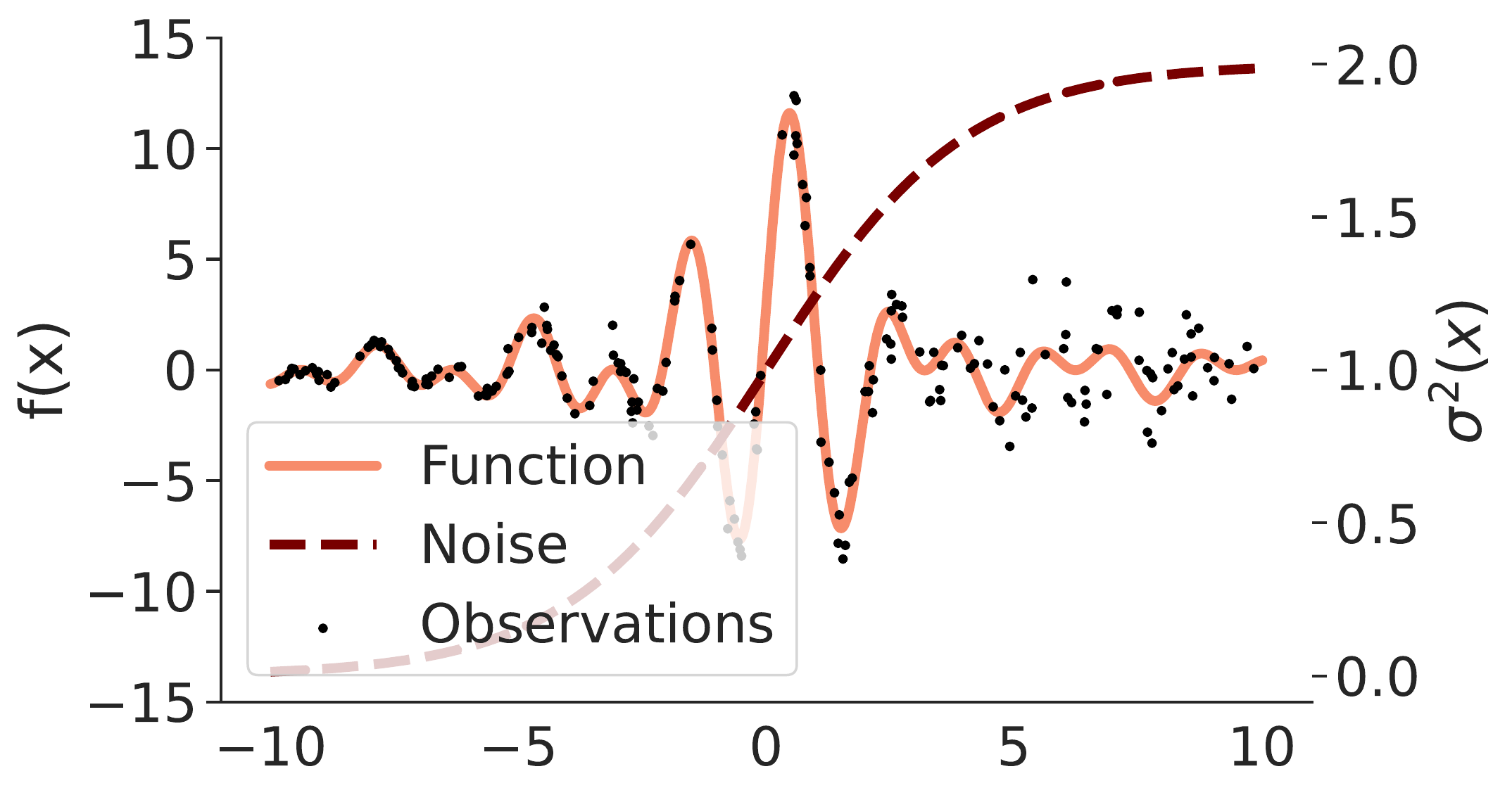}
        \caption{\texttt{sinc} objective fn.}
        \label{fig:mod_sinc}
    \end{subfigure}
    \begin{subfigure}{0.3\textwidth}
        \includegraphics[height=2.25cm,trim=0cm 0cm 0cm 1cm]{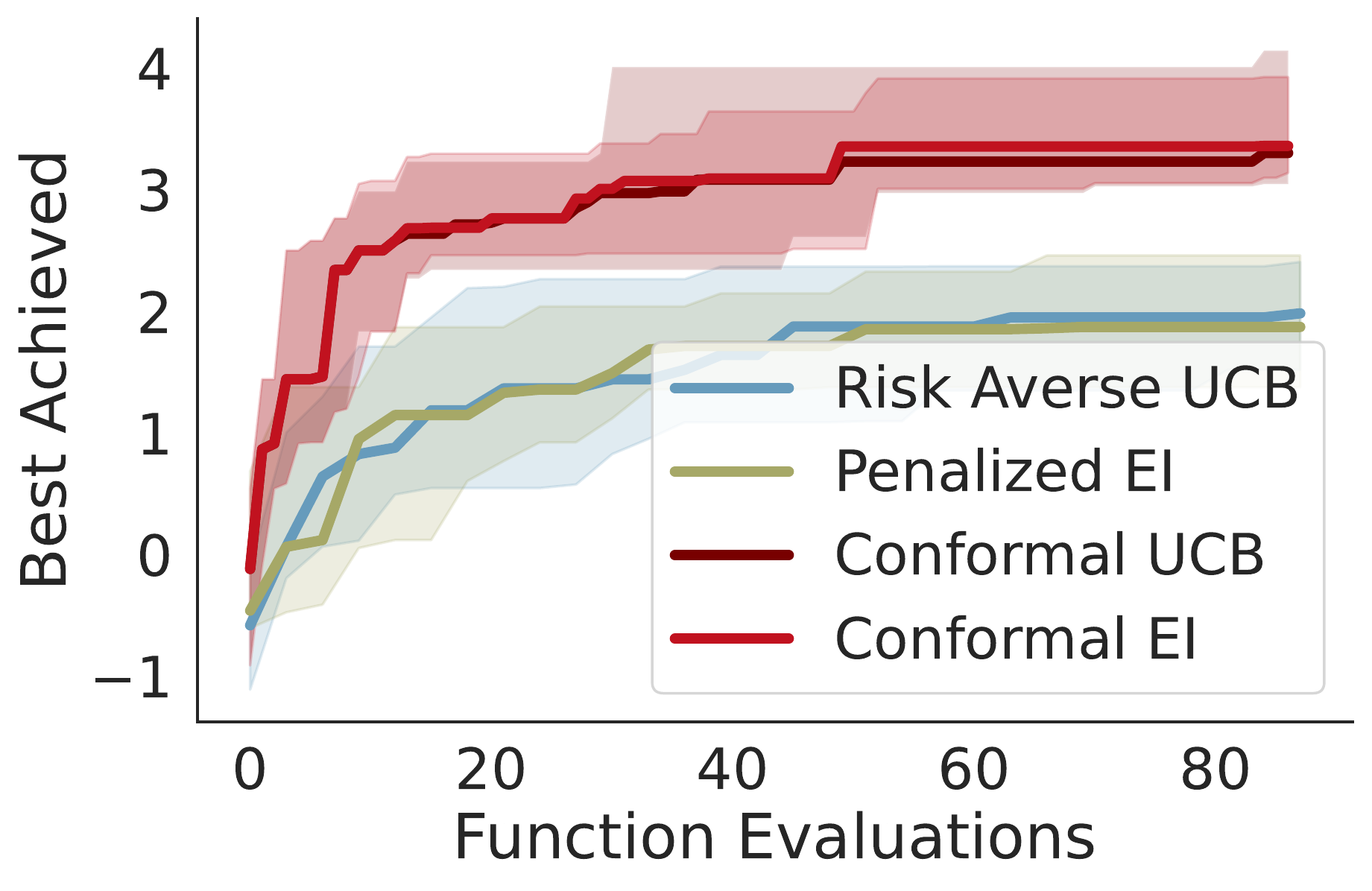}
        \caption{\texttt{sinc}}
        \label{fig:1dhetbo_performance}
    \end{subfigure}
    \begin{subfigure}{0.3\textwidth}
        \includegraphics[height=2.25cm,trim=0cm 0cm 0cm 1cm]{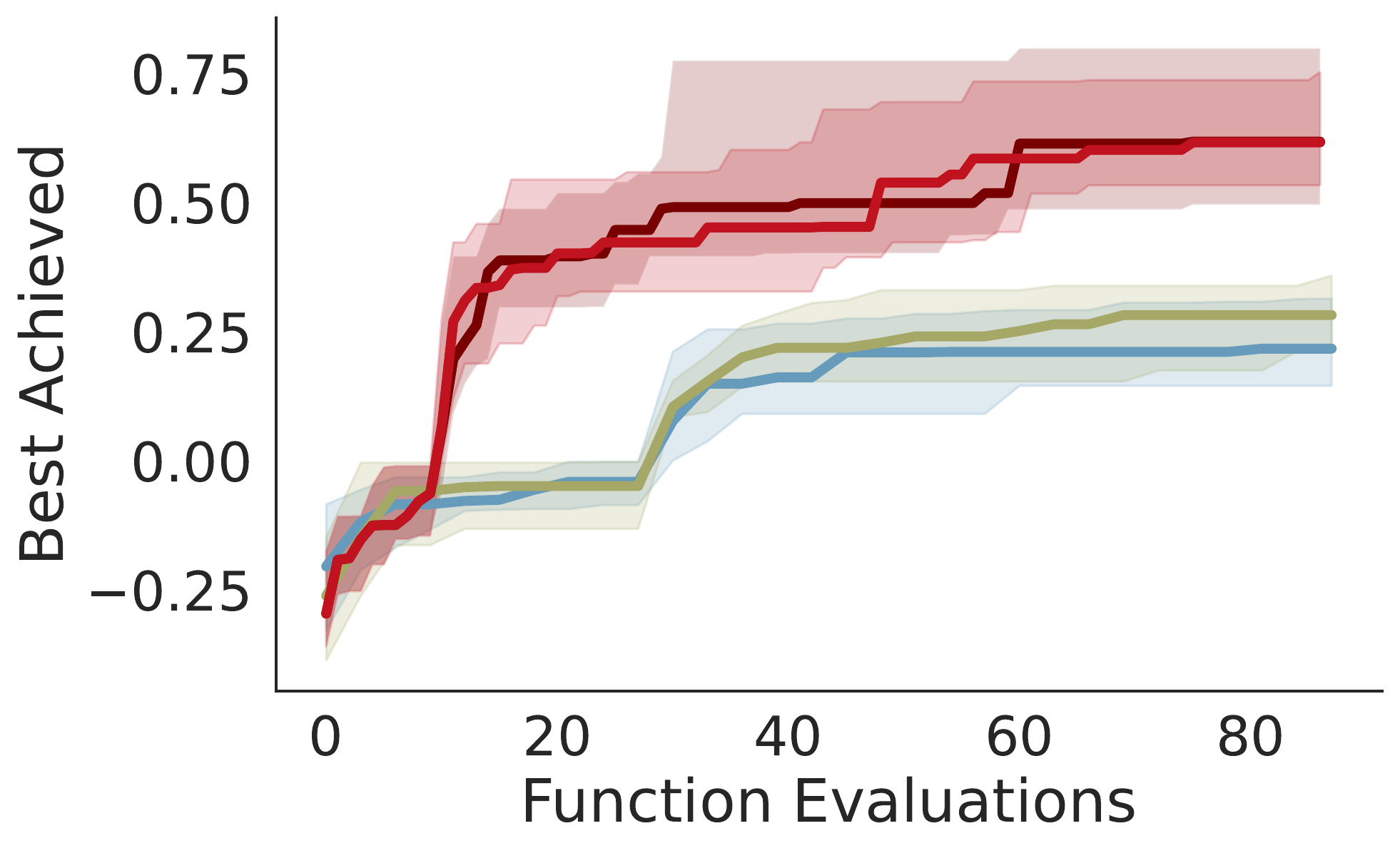}
        \caption{\texttt{double-knot}}
        \label{fig:gramacy_performance}
    \end{subfigure}
    \caption{
    BayesOpt results on heteroscedastic, single-objective tasks \texttt{sinc} and \texttt{double-knot} (reporting median and its $95\%$ conf. interval, estimated from $16$ trials).
    \textbf{(a)} $\texttt{sinc}(\vec x)$ (left $y$ axis) and $\varepsilon(\vec x)$ (right $y$ axis).
    \textbf{(b)} the \texttt{sinc} task, best objective value found by conformal BayesOpt with homoscedastic likelihoods compared to baselines, risk-averse UCB and penalized EI, both with heteroscedastic likelihoods.
    \textbf{(c)} the \texttt{double-knot} task, same experiment as in panel \textbf{(b)}.
    Conformal BayesOpt with a misspecified likelihood outperforms the specialized baselines on both tasks.
    }
    \label{fig:hetbo_1d}
\end{figure*}

Once we have samples from $p(\vec x)$ (which are already in $\mathcal{D}$) and $p'(\vec x | \mathcal{D})$, we estimate $r(\vec{x})$ with a probabilistic classifier \citep{sugiyama2012density}. 
We assign labels $z$ to the samples, corresponding to the conditional distributions $p(\vec{x}) = p(\vec{x} | z = 0)$ and $p'(\vec x | \mathcal{D}) = p(\vec{x} | z = 1)$.
By Bayes theorem, we rewrite $r(\vec{x})$,
\begin{align}
r(\vec{x}) = \frac{p(z=0)}{p(z=1)}\frac{p(z=1 \mid \vec{x})}{p(z=0\mid \vec{x})}, \label{eq:density_ratio_bayes_rule}
\end{align}
such that we need only train a probabilistic classifier $\hat p(z\mid \vec{x})$ to discriminate the sample labels. We estimate the prior ratio $p(z=0) / p(z=1)$ empirically.

\textbf{Which comes first, the acquisition function or the ratio estimator?} To estimate $r$ as just described we clearly must be able to compute $\nabla_{\vec x} a_\alpha$ to draw the required samples from $p'(\vec x | \mathcal{D}) \propto \exp\{a_\alpha (\vec x, \mathcal{D})\}$.
Here we find a second and more serious issue, since $a_\alpha$ itself depends on $r$. 
We need an estimator $\hat r$ that simultaneously induces $p'(\vec x | \mathcal{D}) \propto \exp\{a_\alpha(\vec x, \mathcal{D})\}$ \textit{and} accurately estimates $p'(\vec x | \mathcal{D}) / p(\vec x)$.
For example, we could assume $\hat r(\vec x) = 1, \forall \vec x$, but the induced $p'$ likely does not satisfy $p'(\vec x | \mathcal{D}) / p(\vec x) = 1, \forall \vec x$.

To solve this issue, we begin with an initial estimator $\hat r_0(\vec x) = 1, \forall \vec x$, and for $t \geq 0$ we sample from $p'(\vec x | \mathcal{D}) \propto \exp\{a_\alpha(\vec x, \mathcal{D})\}$ via SGLD using the current estimator $\hat r_t$, then update the classifier on those new samples to produce an updated estimator $\hat r_{t + 1}$ for the next iteration.
To keep the acquisition surface from changing too rapidly (potentially destabilizing our SGLD chain), we compute an exponential moving average of the classifier weights, and the averaged weights are used when computing gradients of $a_\alpha(\vec x, \mathcal{D})$.
Our approach is analogous to (and directly inspired by) bootstrapped deep Q-learning \citep{mnih2015human}.

\section{EXPERIMENTS}
\label{sec:experiments}

In Section \ref{subsec:evaluate_coverage} we report the empirical coverage of credible and conformal prediction sets in a simplified setting.
In Section \ref{subsec:heteroscedastic_results} we show that conformal BayesOpt is robust to a misspecified likelihood.
Finally in Section \ref{subsec:single_obj_results} we evaluate conformal BayesOpt on synthetic black-box optimization tasks, comparing the query coverage of credible and conformal prediction sets.
See Appendix \ref{app:experiment_results} for results on multi-objective synthetic tasks and real ranking tasks using drug and antibody design data,
and see Appendix \ref{app:imp_details} for all experimental details.

\begin{figure*}[!t]
    \centering
    \begin{subfigure}{0.4\textwidth}
    \centering
        \includegraphics[width=\textwidth,clip,trim=0cm 0.2cm 0cm 0cm]{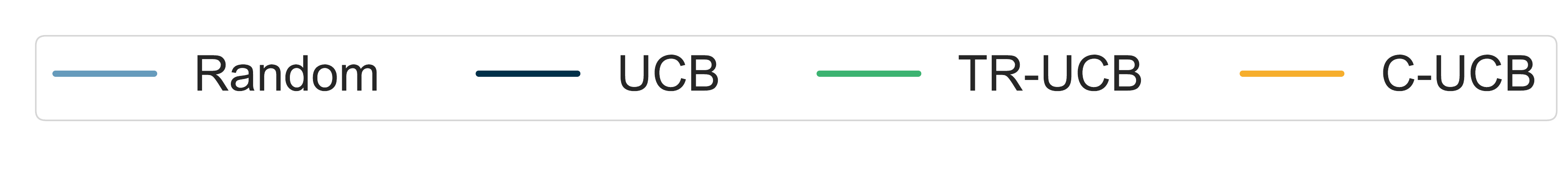}
    \end{subfigure}
    \\
    \begin{subfigure}{0.3\textwidth}
        \includegraphics[height=2.6cm]{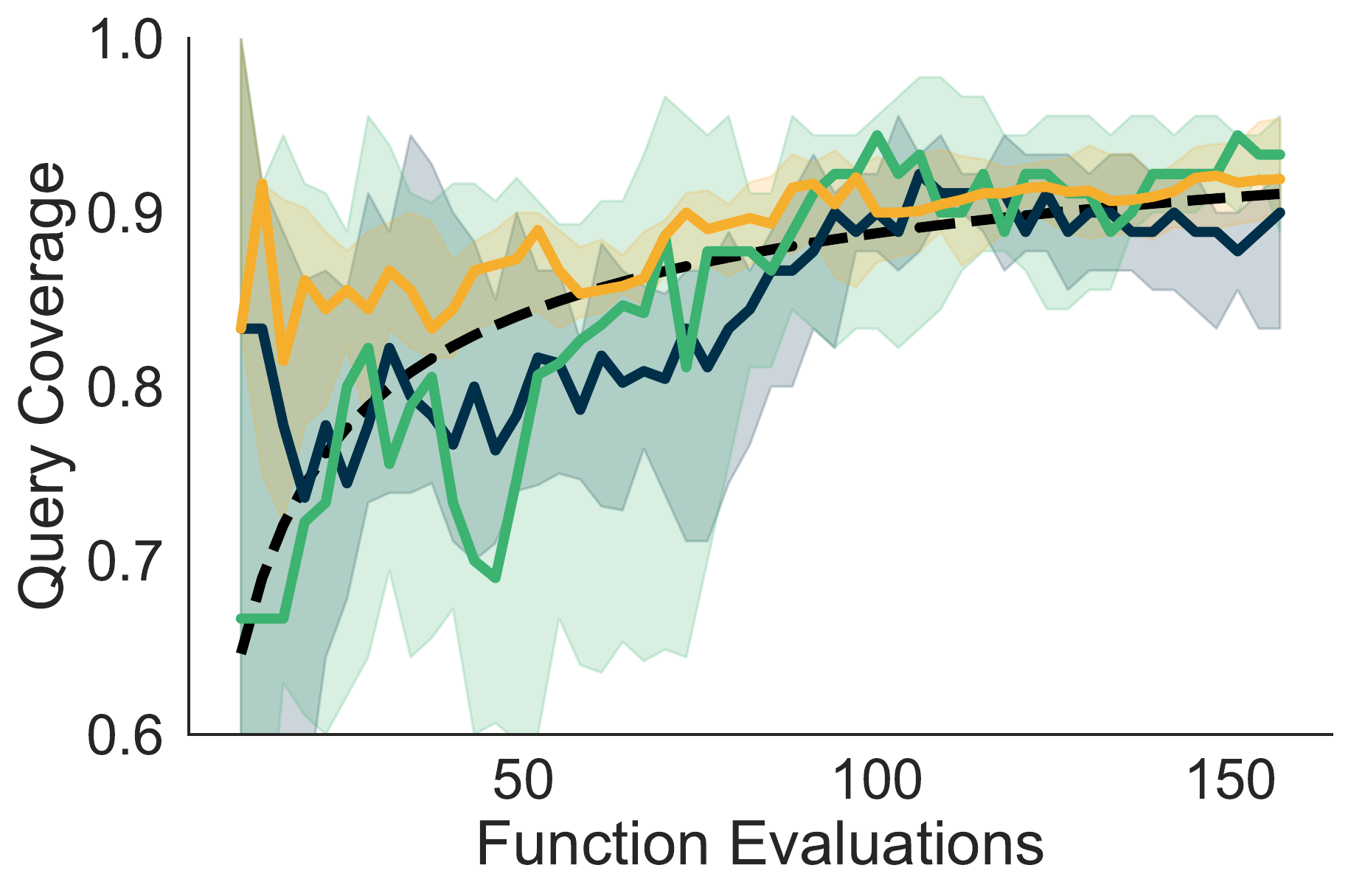}
    \end{subfigure}
    \begin{subfigure}{0.3\textwidth}
        \includegraphics[height=2.6cm]{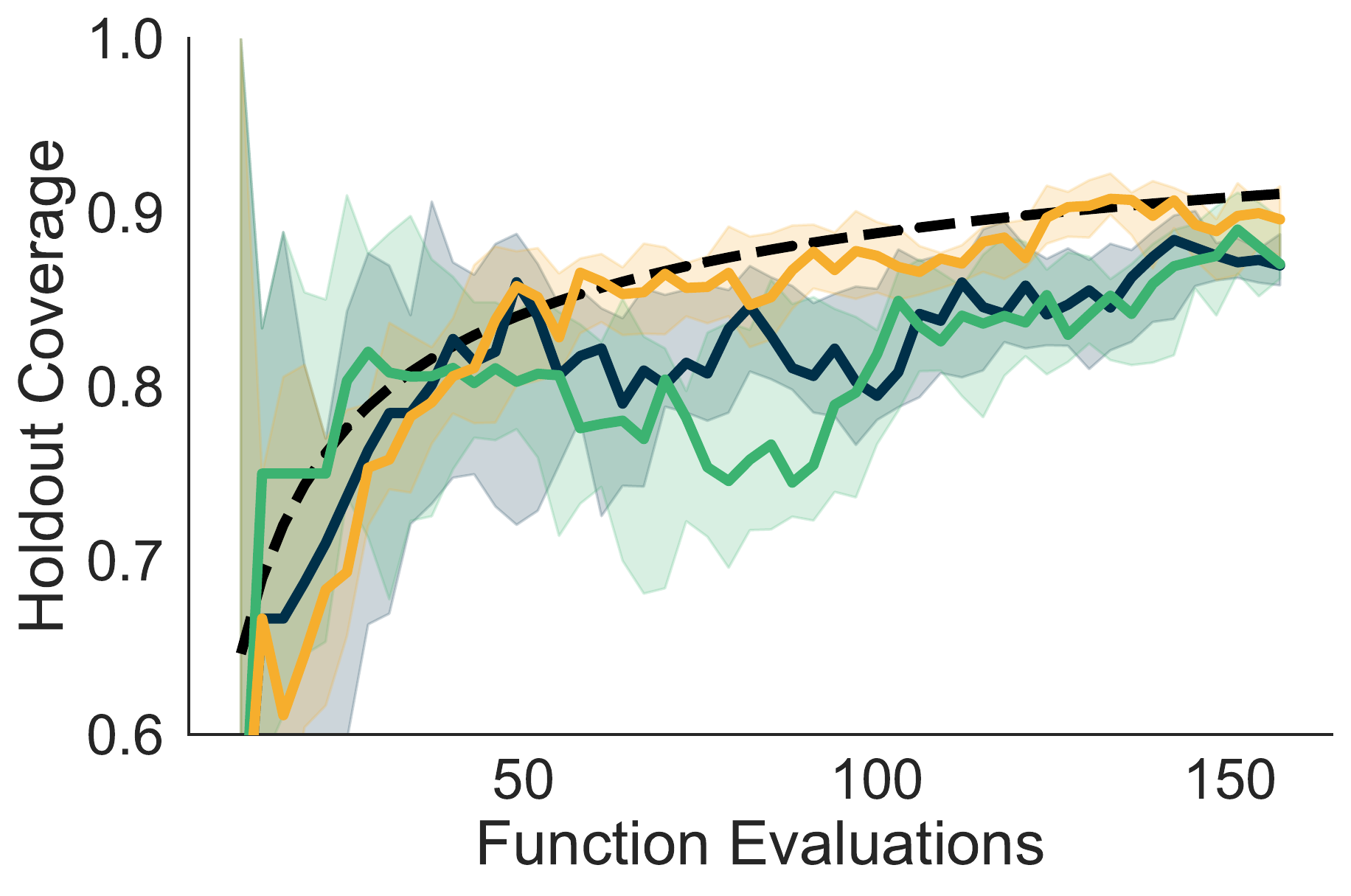}
    \end{subfigure}
    \begin{subfigure}{0.3\textwidth}
        \includegraphics[height=2.6cm]{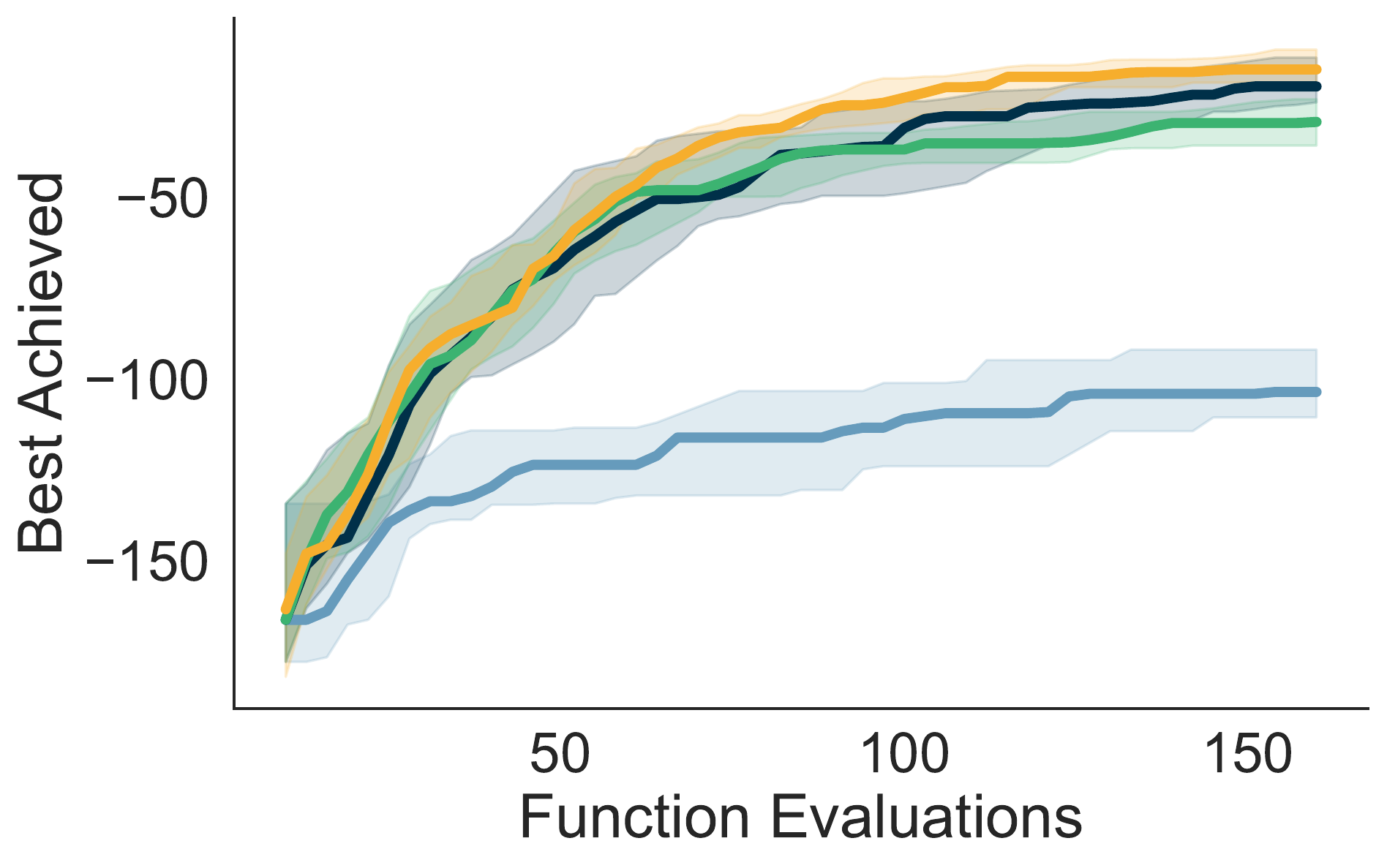}
    \end{subfigure}
    \\
    \begin{subfigure}{0.3\textwidth}
        \includegraphics[height=2.6cm]{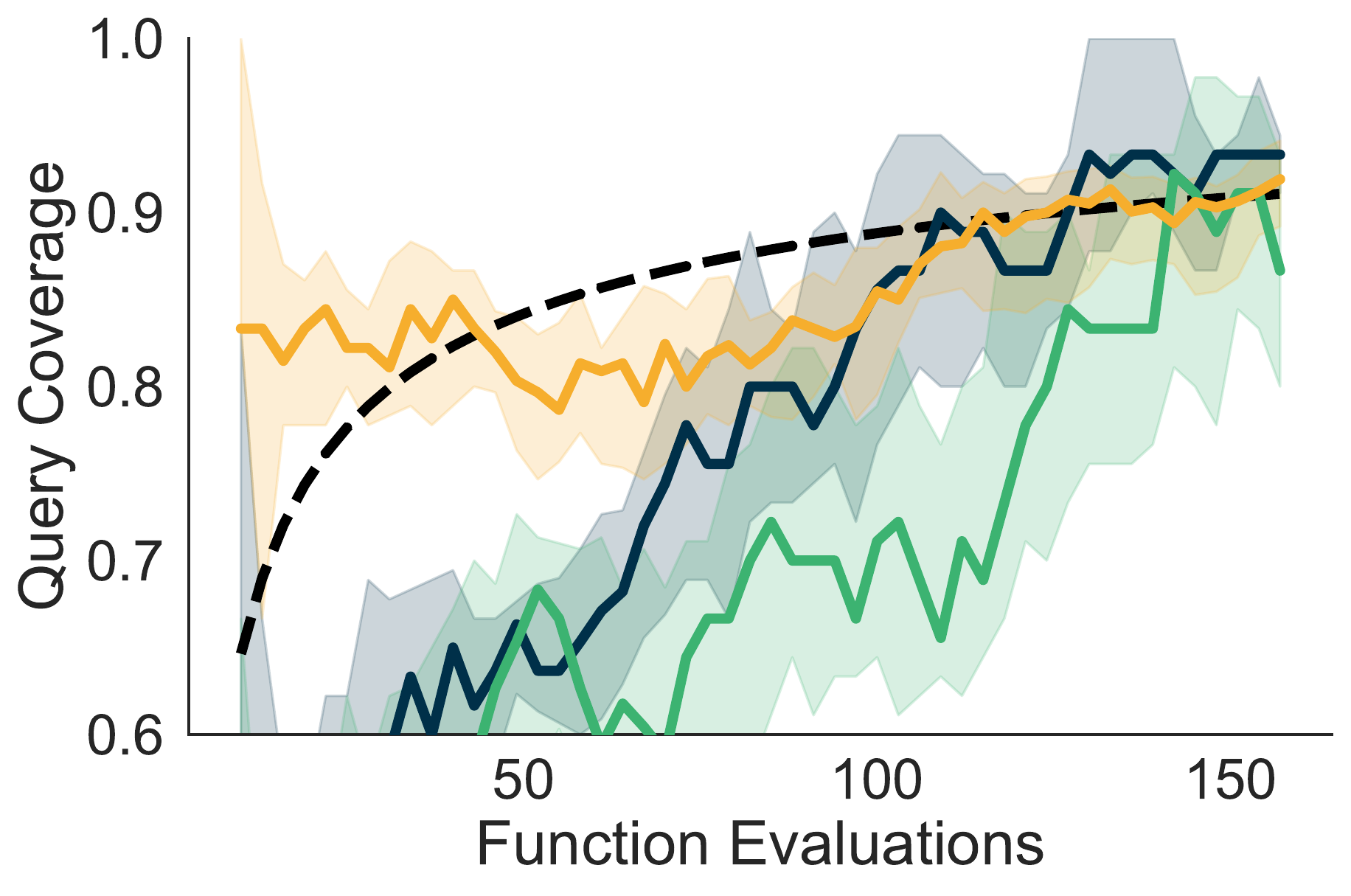}
    \end{subfigure}
    \begin{subfigure}{0.3\textwidth}
        \includegraphics[height=2.6cm]{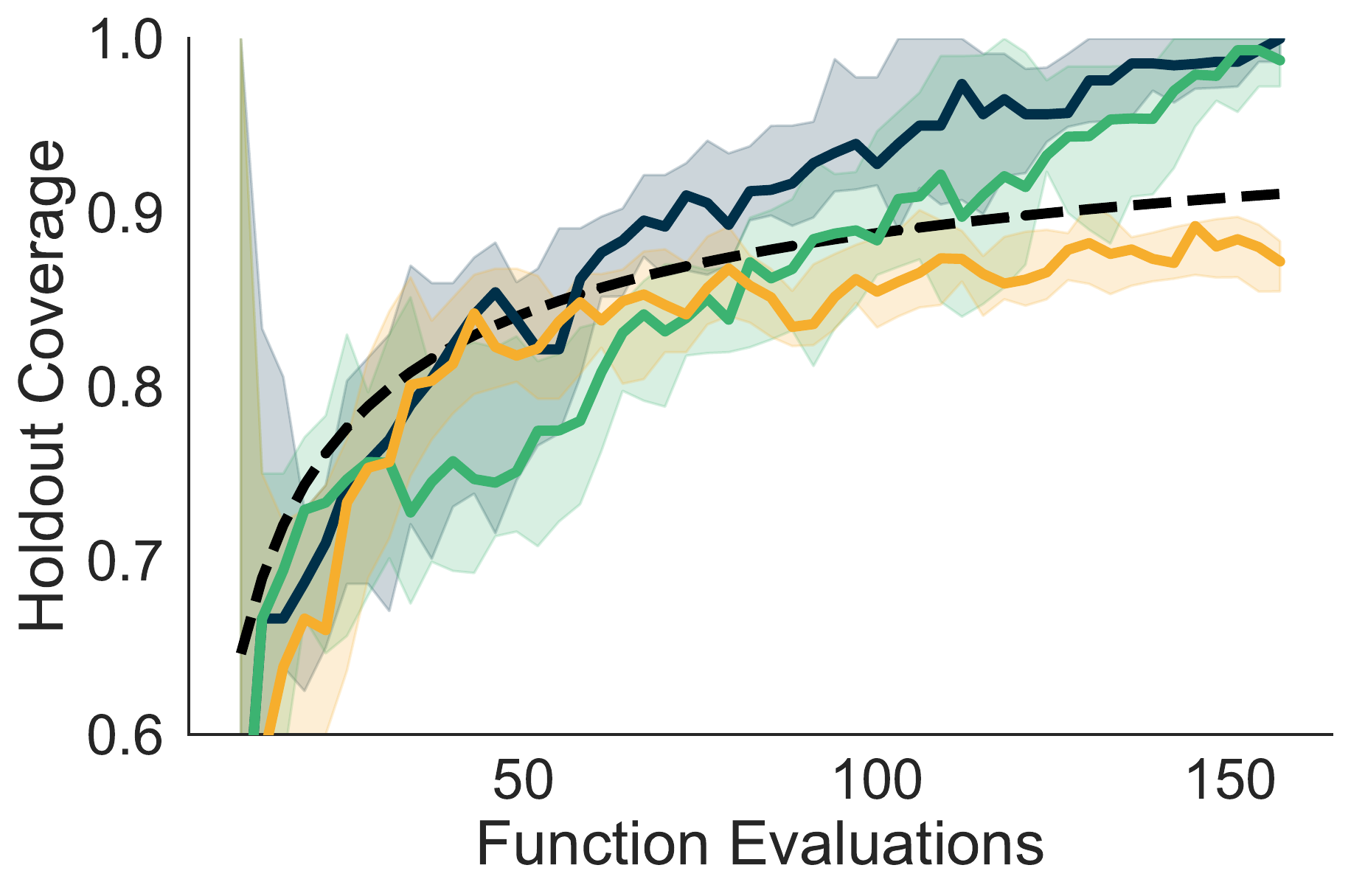}
    \end{subfigure}
    \begin{subfigure}{0.3\textwidth}
        \includegraphics[height=2.6cm]{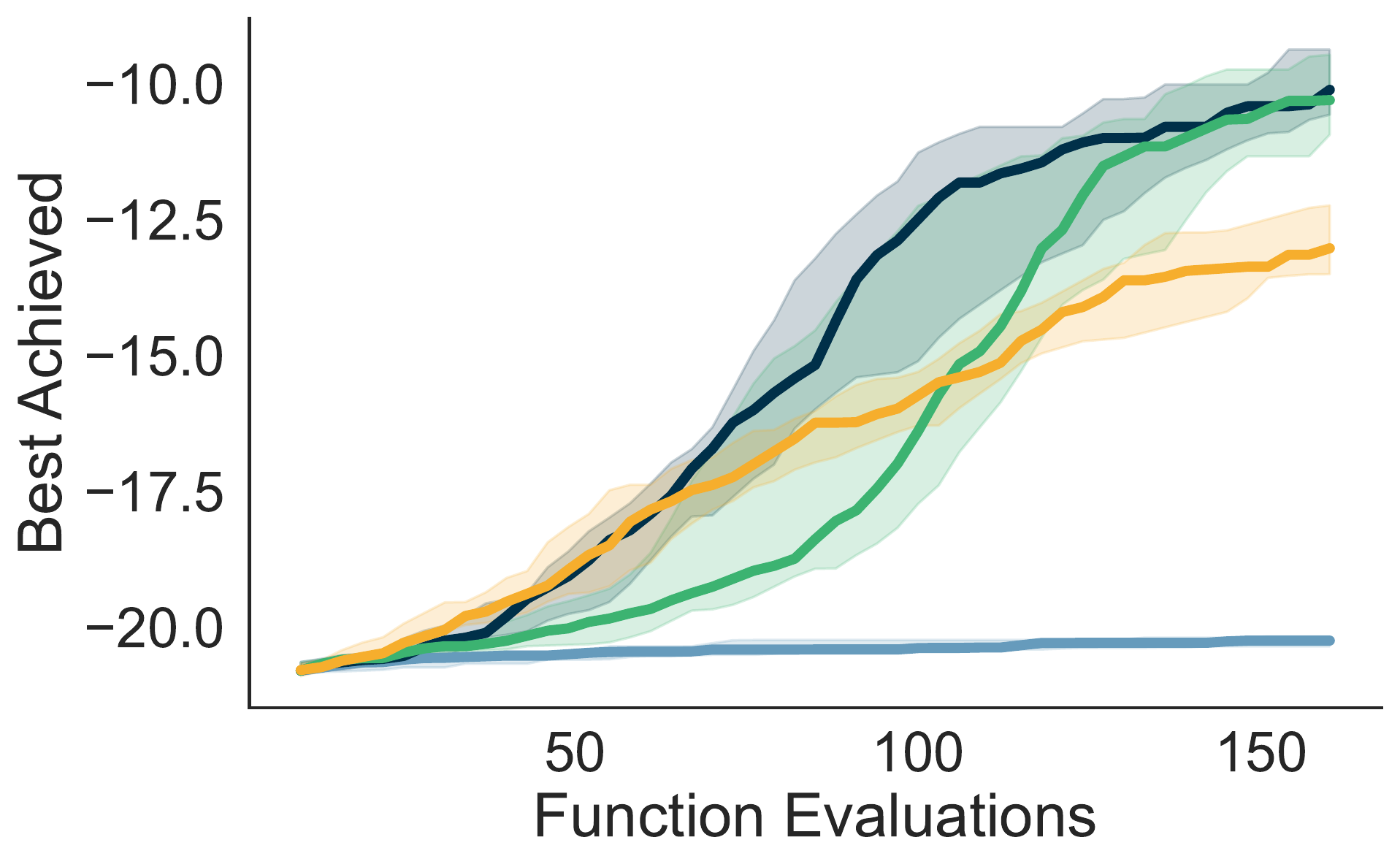}
    \end{subfigure}
\caption{
Using the same base acquisition function, we compare standard BayesOpt (UCB), TuRBO (TR-UCB), and conformal BayesOpt (C-UCB) optimizing \texttt{levy-20d} \textbf{(top row)} and \texttt{ackley-20d} \textbf{(bottom row)}.
The midpoint, lower, and upper bounds of each curve depict the 50\%, 20\%, and 80\% quantiles, estimated from 25 trials.
In the \textbf{left column} we see credible set coverage varying significantly, despite reasonable coverage on a random holdout subset of the training data \textbf{(center column)}.
The \textbf{right column} shows $\max_{0\leq i \leq n} f^*(\vec x_i)$ as $n$ increases, and we see the methods have comparable sample-efficiency.
Conformal BayesOpt improves the objective value while predicting query outcomes much more reliably.
}
\label{fig:std_bayesopt_rewrite}
\end{figure*}

\subsection{Do Our Approximations Impact Coverage?}
\label{subsec:evaluate_coverage}

First we compare the empirical coverage of Bayes credible sets and randomized conformal prediction sets, and evaluate the sensitivity of conformal prediction to continuous relaxation and density ratio estimation.
We consider a simplified offline regression setting where $p(\vec x)$ and $p'(\vec x | \mathcal{D})$ are known 3D spherical Gaussian distributions with different means, $f^*$ is the 3D Hartmann function, and $\vec y | \vec x \sim \mathcal{N}(f^*(\vec x), \sqrt{.05})$.
If the exact validity guarantee of randomized conformal prediction holds, then over many trials the coverage should concentrate around $(1 - \alpha)$.
Some deviation is to be expected due to sample variance and discretization error.
In Figure \ref{subfig:conf_cred_cvrg_comparison} we see when we have the density ratio oracle and $\tau = 0$, that the distribution of conformal coverage is indeed concentrated around $(1 - \alpha)$, especially relative to the distribution of credible coverage.
In the other panels of Figure \ref{fig:approx_error_analysis} we show that empirical density ratio estimates and the continuous relaxation do not compromise validity.
In particular increasing $\tau$ makes the corresponding prediction sets more conservative, which is consistent with the limiting case $\lim_{\tau \rightarrow \infty} C_\alpha(\vec x) = \mathcal{Y}, \; \forall \vec x$.

\subsection{Model Misspecification and Sample-Efficiency}
\label{subsec:heteroscedastic_results}

Recall from Section \ref{subsec:bayes_limitations} that BayesOpt surrogates often use a homoscedastic likelihood $p(\vec y | f) = \mathcal{N}(f, \sigma^2)$, where $\sigma^2$ is a learned constant.
In Figure \ref{fig:mod_sinc} we plot $f^*(\vec x) = \texttt{sinc}(\vec x) := (10 \sin(\vec x)+1) \sin(3\vec x)/\vec x$ on $[-10,10]$ with $\vec y | \vec x \sim \mathcal{N}(f^*(\vec x), \varepsilon(\vec x))$ and $\varepsilon(\vec x) = 2/(1 + \exp\{\vec x/2\})$
In Figure \ref{fig:1dhetbo_performance} we compare conformal BayesOpt with $p(\vec y | f) = \mathcal{N}(f, \sigma^2)$ to two baselines specifically designed for tasks with heteroscedastic noise, risk-averse UCB \citep{makarova2021} and penalized EI
 \citep{griffiths2019achieving}, which both use heteroscedastic likelihoods.
 Both baselines require multiple replicates of each query to update their likelihoods, which significantly reduces sample efficiency.
In Figure \ref{fig:gramacy_performance}, we repeat the same experiment on a second heteroscedastic task $f^*(\vec x) = \texttt{double-knot}(\vec x) := -\vec x_1 \exp\{-\vec x_1 - \vec x_2\}$ on $[-2, 6]^2$, with $\vec y | \vec x \sim \mathcal{N}(f^*(\vec x), \varepsilon(\vec x))$ and $\varepsilon(\vec x) = ||\vec x||_2)$ \citep{gramacy2005bayesian}.
Despite having a simpler, misspecified noise model, conformal BayesOpt finds a better solution with fewer queries.

\subsection{Good Query Coverage and Good Sample- Efficiency Are Not Mutually Exclusive}
\label{subsec:single_obj_results}

We use the batch UCB acquisition function ($q=3$) to optimize two synthetic functions \texttt{levy} and \texttt{ackley}, taking $\mathcal{X} \subset \mathbb{R}^{20}$.
For this experiment $\vec y | \vec x \sim \mathcal{N}(f^*(\vec x), (\sigma^*)^2)$.
To simulate the covariate shift that occurs in many applied problems, we sampled the initial training data from a random orthant of the input space.
In Figure \ref{fig:std_bayesopt_rewrite} we compare the sample efficiency and coverage of standard BayesOpt, TuRBO \citep{eriksson2019scalable}, and conformal BayesOpt.
Each method is comparable in terms of sample efficiency, and the credible set coverage for standard BayesOpt and TuRBO looks reasonable on a random subset of the training data, if a bit unpredictable.
However if we look at the \textit{query coverage} we see that the credible set coverage varies wildly.
The difference between coverage on a random holdout set and coverage on the query set is due to feedback covariate shift.
In contrast, we see that the conformal set coverage for both random holdout points and query points very consistently tracks $(1 - \alpha)$, where $\alpha = 1 / \sqrt{n}$.
In other words, of the methods considered conformal BayesOpt is the only approach that improves the objective while reliably predicting the query outcomes.

\begin{figure}[h]
    \centering
    \begin{subfigure}{0.16\textwidth}
        \includegraphics[width=\textwidth,clip,trim=0cm 0.2cm 0cm 0cm]{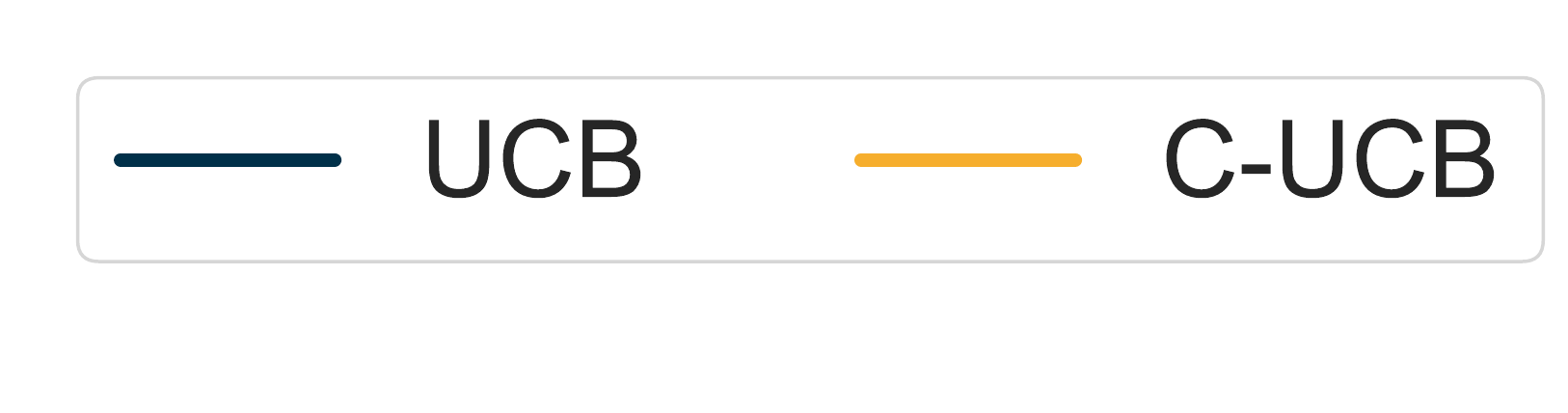}
    \end{subfigure} \\
    \begin{subfigure}{0.24\textwidth}
        \includegraphics[width=\textwidth]{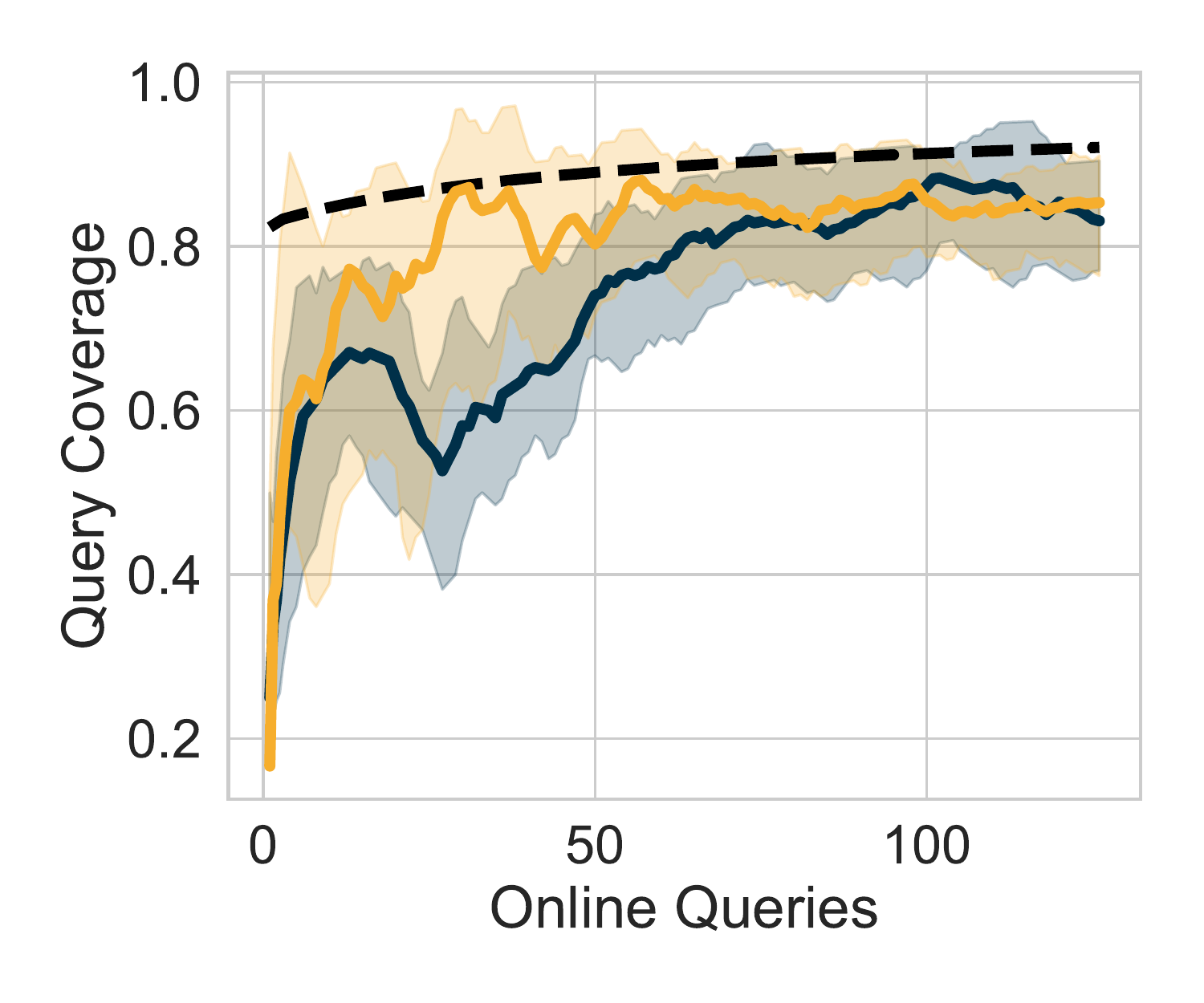}
    \end{subfigure}
    \caption{
        We used C-UCB to select small-molecule compounds for dopamine receptor binding affinity, showing that conformal acquisitions improve query coverage without harming sample efficiency.
    }
    \label{fig:tab_ranking_q_cvrg}
\end{figure}

In Figure \ref{fig:tab_ranking_q_cvrg} we preview results showing we can also improve query coverage in tabular ranking tasks related to drug design.
See Appendix \ref{subsec:tab_ranking_full} for the full experiment.

\section{DISCUSSION}
\label{sec:discussion}

We have shown that a combination of model misspecification and optimization-induced covariate shift can make Bayesian credible sets unreliable exactly where they are needed most --- where the queries concentrate.
Conformal prediction provides a principled solution to these issues, with distribution-free validity guarantees that help ensure robustness to model misspecification, and a natural mechanism to correct for covariate shift.
To use conformal prediction inside of BayesOpt, we have developed differentiable and efficiently discretizable conformal prediction sets and coupled these with a practical density ratio estimation procedure, addressing key technical challenges in the conformal prediction literature.
With the introduction of the conformal Bayes posterior, we have derived conformal generalizations of many popular acquisition functions, allowing us to accommodate features of practical tasks including batched queries, noisy observations, and multiple objectives.
Empirically we find the combination of conformal prediction and BayesOpt to be very promising, since it improves query coverage and has sample-efficiency comparable to methods with no coverage guarantees at all.

Looking forward, although we focus on GP surrogates in the low-$n$ regime, we expect many of these ideas to transfer to much larger models and datasets by replacing full conformal Bayes with split conformal Bayes, and either augmenting GPs with deep kernel learning, or by replacing GPs entirely with Bayesian linear models operating on pretrained representations learned by large self-supervised models.
Extending conformal BayesOpt to discrete optimization, specifically biological sequence design, is a particularly exciting direction for future work.
There are also intriguing theoretical directions, such as analyzing the effect of continuous relaxation and the effect of learned density ratio estimates on conformal coverage guarantees, and investigating whether the regret of conformal BayesOpt can be analyzed with milder assumptions than algorithms like GP-UCB \citep{srinivas2010gaussian}.

As machine learning systems are deployed for increasingly impactful applications, we must confront the reality that machine learning models \textit{will} be built on faulty assumptions, and those models \textit{will} be asked to rank potential decisions without sufficient training data. 
The solution is not to blind ourselves to the error in our assumptions, nor is it to paralyze ourselves in pursuit of a perfect model.
Instead we should develop methods that can gracefully accommodate imperfect models, balancing internal coherence with external validity.

\clearpage
\subsubsection*{Acknowledgments}
The authors thank Sanyam Kapoor for his SGLD implementation, and Anastasios Angelopoulos, Matthias Seeger, Greg Benton, Andres Potapczynski, and Wanqian Yang for helpful discussions. 
This research is supported by NSF CAREER IIS-2145492, NSF I-DISRE 193471, NIH R01DA048764-01A1, NSF IIS-1910266, NSF 1922658 NRT-HDR: FUTURE Foundations, Translation, and Responsibility for Data Science, Meta Core Data Science, Google AI Research, BigHat Biosciences, Capital One, and an Amazon Research Award.

\bibliography{references}
\bibliographystyle{apalike}

\clearpage

\onecolumn

\appendix
\appendixpage

The appendices are structured as follows:
\begin{itemize}
    \item In Appendix \ref{app:limitations}, we describe the assumptions, limitations, and broader impacts of this work.
    \item In Appendix \ref{app:proofs}, we provide detailed derivations of the randomized differentiable conformal prediction, conformal Bayes posterior and of conformal acquisition functions.
    \item In Appendix \ref{app:experiment_results}, we include more experimental results, in particular multi-objective black-box optimization and single-objective tabular ranking tasks with real data.
    \item In Appendix \ref{app:imp_details}, we give implementation details for all experiments.
\end{itemize}

\clearpage
\section{ASSUMPTIONS, LIMITATIONS, AND BROADER IMPACTS}
\label{app:limitations}

\subsection{Assumptions}
\label{subsec:assumptions}

The assumptions underlying the coverage guarantee for conformal prediction are strikingly mild.
All else equal, any real-valued, measurable score function will produce a valid prediction set \citep{vovk2005algorithmic}.
There are trivial examples that produce trivially valid prediction sets $\mathcal{C}_\alpha(\vec x) = \mathcal{Y}, \; \forall \vec x, \; \forall \alpha$.
In general if we choose $s$ poorly we pay a price in terms of \textit{efficiency} (i.e. the volume of the prediction sets), but validity is still maintained.

The critical assumption is that $\{(\vec x_0, \vec  y_0), \dots, (\vec x_n, \vec  y_n)\}$ are pseudo-exchangeable.\footnote{
Note that every IID sequence of random variables is exchangeable, but not every exchangeable sequence is IID.
Similarly pseudo-exchangeability does not mean every element of the sequence except for the last is IID.}
A sequence of random variables is pseudo-exchangeable if the joint density can be factored into terms that only depend on the values of the sequence, not the ordering \citep{fannjiang2022conformal}.
Informally, we can see that BayesOpt satisfies pseudo-exchangeability because the likelihood of the training data is just the mixture of all the previous query likelihoods, and the query likelihoods do not depend on the order of the past observations.
Because we make no assumptions about the data distribution beyond pseudo-exchangeability, conformal prediction belongs to a class of methods known as \textit{distribution-free} uncertainty quantification.

\subsection{Limitations}

\textbf{Marginal vs. conditional coverage guarantees}: full and split conformal prediction sets have marginal coverage guarantee that is easy to confuse with conditional coverage guarantees \citep{angelopoulos2021gentle}. 
Marginal coverage guarantees must be interpreted with the same frequentist mindset as other frequentist measures of uncertainty, such as confidence intervals and $p$-values, with similar risks of misinterpretation by inexperienced users.
We have attempted to make clear in the main text that the full conformal prediction coverage guarantee is only realized in the aggregate, as the average of coverages observed in many independent, parallel experiments.
Coverage observed within any specific trial for any specific input can (and does) vary substantially from the aggregate tendency.
There is very recent work which seeks to provide a stronger conditional validity guarantee that can be expected to hold for some $(1 - \delta)$ fraction of trials, which we hope to apply to conformal BayesOpt in future work \citep{bates2021testing}.

\textbf{Approximation error:} we have introduced some necessary approximations in this work, notably the discretization of continuous labels and the continuous relaxation of conformal prediction sets.
While we have given empirical evidence that the error introduced by these approximations does not appear to be too severe, practitioners should be aware that some deviation from the expected coverage level may occur, as we discuss in Section \ref{sec:experiments}.
This limitation is analogous to the numeric limitations of linear algebra implemented with floating point arithmetic.
We may be able to make use of \citet{ndiaye2022stable} to avoid discretizing continuous outcomes entirely, which we leave for future work.

\subsection{Broader Impacts:}

\textbf{Potential negative social impacts:} black-box optimization algorithms are application-agnostic.
The same algorithms that are being used to design new therapeutics could in theory be used to discover new toxins for bioterror or biowarfare.
Similarly, the same algorithms used to design new materials for scientific discovery could be used to design new weapons or rocket fuels.
Our work is not particularly vulnerable to misuse relative to the large body of existing work on black-box optimization algorithms.

\textbf{Machine learning research:} phenomena like model misspecification and covariate shift are often blamed on complexity in the external world, but they are also induced by our own behavior, such as choosing a convenient likelihood for a model (even when a more sophisticated option is available) or actively selecting new training data.
We hope this work spurs more interest in understanding how to reliably interact with the models we have \textit{today}, in addition to work on ``better'' models for tomorrow.

\textbf{Experimental design:} applications like materials science and drug discovery require the coordination of large, interdisciplinary teams of scientists and engineers. If machine learning systems are to play a central role in that coordination, they must be reliable, in the sense that the systems should have stable behavior and consistently valid predictions.
That kind of reliability requires more than faith in an ad hoc collection of modeling assumptions with limited experimental validation.
This work is a step towards machine learning systems with interpretable certificates of reliability that can serve as the foundation on which to build teams which push the boundaries of experimental science.

\clearpage
\section{PROOFS AND DERIVATIONS}
\label{app:proofs}

\subsection{Smoothed conformal prediction}
\label{subsec:randomized_conf_pred}

\begin{algorithm}[!h]
\caption{Randomized differentiable conformal prediction masks}
\label{alg:randomized_differentiable_conformal_masks}
\SetAlgoLined
    \KwData{train data $\mathcal{D} = \{(\vec x_i, \vec y_i)\}_{i=0}^{n-1}$, test point $\vec x_n$, imp. weights $\vec w$, label candidates $Y_{\mathrm{cand}}$, score function $s$, miscoverage tolerance $\alpha$, relaxation strength $\tau$.}
    
      $m_j = 0$, $\forall j \in \{0, \dots, k - 1\}$. 
      
      \For{$\hat{\vec y}_j \in Y_{\mathrm{cand}}$}{
        $\hat{\mathcal{D}} \leftarrow \mathcal{D} \cup \{(\vec x_n, \hat{\vec y}_j)\}$
        
        $\vec s \leftarrow [s(\vec x_0, \vec y_0, \hat{\mathcal{D}}) \; \cdots \; s(\vec x_n, \hat{\vec y}_j, \hat{\mathcal{D}})]^\top$. 
        
        $\vec h \leftarrow \texttt{sigmoid}(\tau^{-1}(\vec s - s_{n}))$. 
        
        $\overline{w} \leftarrow \vec h^\top \vec w$. 
        
        $\theta \leftarrow \texttt{clip}(w_n^{-1} (\overline{w} - \alpha), 0, 1)$.
        
        $\eta \sim \mathrm{Bernoulli}(\theta)$.
        
        $\overline{w}' = \overline{w} - (1 - \eta)w_n$.
        
        $m_j \leftarrow \texttt{sigmoid}(\tau^{-1}(\overline{w}' - \alpha))$.
      }
    \KwResult{$\vec m$}
\end{algorithm}

As discussed in Section \ref{subsec:conformal_prediction_background} of the main text, standard conformal prediction (Definition \ref{def:conf_bayes_pred_set}) is \textit{conservatively valid}, meaning in the long run the coverage of conformal prediction sets is \textit{at least} $1 - \alpha$.
If the prediction sets are too conservative, they may be too wide to be helpful for decision-making.
In the BayesOpt context we want prediction sets that are \textit{exactly valid}, neither underconfident nor overconfident.
Fortunately with a small change (i.e. randomization) conformal prediction sets can be made exactly valid.

Informally, exact validity only requires that we treat an edge case more carefully (see ``smoothed conformal predictors'' in \citet{vovk2005algorithmic} for more details).
Specifically there will be some candidate labels $\hat{\vec y}_j$ that are right on the boundary of the prediction set, and we will introduce randomness to sometimes include such points, and sometimes not, depending on exactly how close to the boundary the points are.

More precisely, there are occasions when
\begin{align*}
    \sum_{i=0}^{n-1} h_i w_i < \alpha < \sum_{i=0}^n h_i w_i = \overline{w}.
\end{align*}
In standard conformal prediction the corresponding label $\hat{\vec y}_j$ would always be accepted into the prediction set.
To make a smoothed conformal predictor, we flip a coin with bias $\theta = \texttt{clip}(w_n^{-1} (\overline{w} - \alpha)), 0, 1)$.
We call the outcome of the flip $\eta$.
If $\eta = 1$, then we accept $\hat{\vec y}_j$, similarly if $\eta = 0$ we reject $\hat{\vec y}_j$.
Note if $\overline{w} - \alpha < 0$ then $\hat{\vec y}_j$ is always rejected, similarly if $\overline{w} - \alpha > w_n$ then $\hat{\vec y}_j$ is always accepted.
Informally we are checking to see if the contribution of $w_n$ is responsible for pushing $\overline{w}$ over the threshold $\alpha$, and if so we are probabilistically accepting $\hat {\vec y}_j$ depending on how close $\overline{w} - w_n$ is to $\alpha$.
We give the continuous relaxation of smoothed conformal prediction in Algorithm \ref{alg:randomized_differentiable_conformal_masks}.

\subsection{Characterizing the conformal Bayes posterior}
\label{subsec:conf_bayes_posterior_details}

All conditional distributions are also conditioned on $\mathcal{D}$, which we omit from the notation for the sake of clarity.
Recall that the conformal Bayes posterior is written as
\begin{align*}
    p(f(\vec x) | \vec x) &= \int\limits_{\mathcal{Y}} p(f(\vec x) | \vec x, \vec y)p(\vec y | \vec x)d\vec y, \\
    &= \int\limits_{C_\alpha(\vec x)} p(f(\vec x) | \vec x, \vec y)p(\vec y | \vec x)d\vec y +
    \int\limits_{\mathcal{Y} \setminus C_\alpha(\vec x)} p(f(\vec x) | \vec x, \vec y)p(\vec y | \vec x)d\vec y.
\end{align*}
Now we define a new conformal Bayes posterior distribution as a mixture distribution over $\vec y$,
\begin{align}
    p_\alpha(\vec y | \vec x) &= (1 - \alpha) q_1(\vec y | \vec x) + \alpha q_2(\vec y | \vec x)& \label{eq:conf_bayes_posterior_mixture}\\
    Z_1 &= \int\limits_{C_\alpha(\vec x)} 1 d\vec y,
    &q_1(\vec y | \vec x) =
    \begin{cases}
     1 / Z_1 & \text{if } \vec y \in C_\alpha(\vec x), \\
    0 & \text{else},
    \end{cases} \nonumber \\
    Z_2 &= \int\limits_{\mathcal{Y} \setminus C_\alpha(\vec x)} p(\vec y | \vec x)d\vec y,
    &q_2(\vec y | \vec x) = \begin{cases}
     0 & \text{if } \vec y \in C_\alpha(\vec x), \\
    p(\vec y | \vec x) / Z_2 & \text{else},
    \end{cases} \nonumber
\end{align}
where the normalizing constants $Z_1, Z_2$ ensure that $\int p_\alpha(\vec y | \vec x)d\vec y = 1$ (assuming $C_\alpha(\vec x)$ is bounded and non-empty, so $Z_1$ is non-zero and finite). If $\mathcal{C}_\alpha(\vec x) = \mathcal{Y}$ and $\mathcal{Y}$ is unbounded then $p_\alpha(\vec y | \vec x)$ is not a proper density.

The corresponding conformal Bayes posterior distribution over $f$ is
\begin{align*}
    p_\alpha(f(\vec x) | \vec x) &= \int\limits_{\mathcal{Y}} p(f(\vec x) | \vec x, \vec y)p_\alpha(\vec y | \vec x)d\vec y \\
    &= \frac{1 - \alpha}{Z_1} \int\limits_{C_\alpha(\vec x)} p(f(\vec x) | \vec x, \vec y)d\vec y +
    \frac{\alpha}{Z_2} \int\limits_{\mathcal{Y} \setminus C_\alpha(\vec x)} p(f(\vec x) | \vec x, \vec y)p(\vec y | \vec x)d\vec y
\end{align*}
Finally we can rewrite both integrals over all $\mathcal{Y}$ by introducing a binary mask,
\begin{definition}
\label{def:conf_bayes_posterior}
\begin{align*}
    p_\alpha(f(\vec x) | \vec x) &:= \frac{1 - \alpha}{Z_1} \int m_\alpha(\vec x, \vec y) p(f(\vec x) | \vec x, \vec y)d\vec y \\
    & \hspace{4mm} + \frac{\alpha}{Z_2} \int (1 - m_\alpha(\vec x, \vec y)) p(f(\vec x) | \vec x, \vec y)p(\vec y | \vec x)d\vec y, \\
    m_\alpha(\vec x, \vec y) &:= \begin{cases}
    1 & \text{if } \vec y \in C_\alpha(\vec x), \\
    0 & \text{else}.
    \end{cases}
\end{align*}
\end{definition}

\begin{proposition}
\label{prop:conf_bayes_posterior_limit}
Let $n > 1$ and $p_\alpha(f | D)$ be defined according to Definition \ref{def:conf_bayes_posterior}.
Then $p_\alpha(f | D)$ converges pointwise in $\vec x$ to $p(f(\vec x) | \vec x, D)$ as $\alpha \rightarrow 1$,
\begin{align*}
    \lim_{\alpha \rightarrow 1} p_\alpha(f(\vec x) | \vec x) = p(f | \vec x).
\end{align*}
\end{proposition}

\textbf{Proof:} \\

Let $\varepsilon > 0$, $n > 2$, and define $\alpha_k = 1 - 1 / (k + 1)$ for $k \in \mathbb{N}$.
\begin{align*}
    |p_{\alpha_k}(f(\vec x) | \vec x) - p(f(\vec x) | \vec x)| &= |\Delta_1 + \Delta_2|, \\
    &\leq |\Delta_1| + |\Delta_2|,
\end{align*}
where
\begin{align*}
    \Delta_1 &= \frac{1 - \alpha_k}{Z_1} \int\limits_{C_{\alpha_k}(\vec x)} p(f(\vec x) | \vec x, \vec y)d\vec y
    - \int\limits_{C_{\alpha_k}(\vec x)} p(f(\vec x) | \vec x, \vec y)p(\vec y | \vec x)d\vec y, \\
    \Delta_2 &= \frac{\alpha_k}{Z_2} \int\limits_{\mathcal{Y} \setminus C_{\alpha_k}(\vec x)} p(f(\vec x) | \vec x, \vec y)p(\vec y | \vec x)d\vec y - \int\limits_{\mathcal{Y} \setminus C_{\alpha_k}(\vec x)} p(f(\vec x) | \vec x, \vec y)p(\vec y | \vec x)d\vec y.
\end{align*}

Recalling the definition of $C_\alpha(\vec x)$ (Def. \ref{def:conf_bayes_pred_set}), we observe that $C_{\alpha_k}(\vec x) \supset C_{\alpha_{k+1}}(\vec x), \forall k \in \mathbb{N}$.\footnote{$A \supset B$ indicates that $A$ is a strict superset of $B$.} 
Furthermore we see that since the importance weights $\vec w$ must sum to 1 that $\lim_{k \rightarrow \infty} C_{\alpha_k}(\vec x) = \emptyset$.

\textbf{Bounding $|\Delta_1|$:}\\
\begin{align*}
    |\Delta_1| &\leq |\mathcal{O}(1 - \alpha_k) - \mathcal{O}(1 - \alpha_k)|, \\
    \Rightarrow |\Delta_1| &\leq c_1(1 - \alpha_k).
\end{align*}

\textbf{Bounding $|\Delta_2|$:}\\
\begin{align*}
    |\Delta_2| &\leq |(\alpha_k - 1)\mathcal{O}(1)|, \\
    \Rightarrow |\Delta_2| &\leq c_2(1 - \alpha_k).
\end{align*}

Choose $k \in \mathbb{N}$ large enough that $(c_1 + c_2)(1 - \alpha_k) < \varepsilon$. \hfill $\blacksquare$

\subsection{Monte Carlo integration of conformal acquisition functions}
\label{subsec:mc_integration_details}

We want to integrate acquisition functions of the form
\begin{align*}
    a(\vec x, \mathcal{D}) &= \int u(\vec x, f, \mathcal{D})p_\alpha(f | \mathcal{D})df \\
    &= \frac{1 - \alpha}{Z_1} \int \int u(\vec x, f, \mathcal{D}) m_\alpha(\vec x, \vec y) p(f(\vec x) | \vec x, \vec y)d\vec ydf \\
    & \hspace{4mm} + \frac{\alpha}{Z_2} \int \int u(\vec x, f, \mathcal{D}) (1 - m_\alpha(\vec x, \vec y)) p(f(\vec x) | \vec x, \vec y)p(\vec y | \vec x)d\vec ydf,
\end{align*}

Suppose we have sampled $Y_{\mathrm{cand}} = \{\hat{\vec y}_j\}_{j=0}^{k-1}$, with $\hat{\vec y}_j \sim p(\vec y | \vec x, \mathcal{D})$, and $f^{(j)} \sim p(f | \mathcal{D} \cup \{(\vec x, \hat{\vec y}_j)\})$.
Starting with the first term in the sum, we have
\begin{align*}
    \frac{1 - \alpha}{Z_1} \int \int u(\vec x, f, \mathcal{D}) m_\alpha(\vec x, \vec y) p(f(\vec x) | \vec x, \vec y)d\vec ydf &\approx \frac{1 - \alpha}{Z_1 k} \sum_{j=0}^{k-1} \frac{m_\alpha(\vec x, \hat{\vec y}_j)}{p(\hat{\vec y}_j | \vec x)} u(\vec x, f^{(j)}, \mathcal{D}).
\end{align*}
We estimate the normalization constant $Z_1$ as follows:
\begin{align*}
    Z_1 = \int\limits_{C_\alpha(\vec x)} 1 d\vec y
    &= \int m_\alpha(\vec x, \vec y)d\vec y \\
    &\approx \frac{1}{k}\sum_{j=0}^{k-1} \frac{m_\alpha(\vec x, \hat{\vec y}_j)}{p(\hat{\vec y}_j | \vec x)}
\end{align*}

By similar logic the second term in the sum is estimated as follows:
\begin{align*}
    \frac{\alpha}{Z_2} \int \int u(\vec x, f, \mathcal{D}) (1 - m_\alpha(\vec x, \vec y)) p(f(\vec x) | \vec x, \vec y)p(\vec y | \vec x) d\vec ydf &\approx \frac{\alpha}{Z_2 k} \sum_{j=0}^{k-1} (1 - m_\alpha(\vec x, \hat{\vec y}_j)) u(f^{(j)}, \mathcal{D}),
\end{align*}
where
\begin{align*}
    Z_2 = \int\limits_{\mathcal{Y} \setminus C_\alpha(\vec x)} p(\vec y | \vec x)d\vec y
    &= \int (1 - m_\alpha(\vec x, \vec y))p(\vec y | \vec x)d\vec y \\
    &\approx \frac{1}{k}\sum_{j=0}^{k-1} (1 - m_\alpha(\vec x, \hat{\vec y}_j))
\end{align*}
Upon further inspection we see that $k$ drops out of the equations, and in effect we are simply computing weighted sums, where the weights have been normalized to sum to 1, i.e.
\begin{align}
    a(\vec x, \mathcal{D}) \approx \hat a(\vec x, \mathcal{D}) &= (1 - \alpha) \vec u^\top \vec v + \alpha \vec u^\top \vec v', \nonumber \\
    \vec u &= [u(\vec x, f^{(0)}, \mathcal{D}) \; \cdots \; u(\vec x, f^{(k-1)}, \mathcal{D})]^\top,  \nonumber \\
    \vec v_j &= \frac{m_\alpha(\vec x, \hat{\vec y}_j)}{p(\hat{\vec y}_j | \vec x)} \left(\sum_{i=0}^{k-1} \frac{m_\alpha(\vec x, \hat{\vec y}_i)}{p(\hat{\vec y}_i | \vec x)}\right)^{-1}, \nonumber \\
    \vec v'_j &= (1 - m_\alpha(\vec x, \hat{\vec y}_j)) \left(\sum_{i=0}^{k-1} (1 - m_\alpha(\vec x, \hat{\vec y}_i))\right)^{-1}. \nonumber
\end{align}

\subsection{Conformalized Single Objective Acquisitions}
\label{subsec:more_conf_acqs}

\textbf{Conformal NEI:} rather than taking $u(\vec x, f, \mathcal{D}) = [f(\vec x) - \max_{\vec y_i \in \mathcal{D}} \vec y_i]_+$ (which corresponds to EI), take \begin{align}
    u_{\mathrm{NEI}}(\vec x, f, \mathcal{D}) = [f(\vec x) - \max_{\vec x'_i \in \mathcal{D}} f(\vec x'_i)]_+.
\end{align}
Note that $u$ is now a function of the joint collection of function evaluations $\{f(\vec x), f(\vec x'_0), \dots, f(\vec x'_{n-1})\}$ \citep{letham2019constrained}.

\textbf{Conformal UCB:}
the reparameterized form of UCB was originally derived in \citet{wilson2017reparameterization} as follows:
\begin{align*}
\mathrm{UCB}(\vec x) = \int u_{\mathrm{UCB}}(\vec x, f, \mathcal{D})\mathcal{N}\left(f \bigg| \mu, \frac{\lambda \pi}{2} \Sigma\right)df,
\end{align*}
where $\lambda > 0$ is a hyperparameter balancing the explore-exploit tradeoff, $\mu, \Sigma$ are the mean and covariance of $p(f | \mathcal{D})$, and 
$u_{\mathrm{UCB}}(\vec x, f, \mathcal{D}) = \mu + \frac{\lambda \pi}{2} |\mu - f|$.
Because UCB is optimistic, the conformalization procedure is a little different than the previous acquisition functions.
When marginalizing out the outcomes $\vec y$ to obtain the conformal Bayes posterior, we integrate over the restricted outcome space $\mathcal{Y}_\mu = \{\vec y \in \mathcal{Y} | \vec y \geq \mu\}$.
Hence we derive conformal UCB as
\begin{align*}
\mathrm{CUCB}_\alpha(\vec x) = \int \int\limits_{\mathcal{Y}_\mu} u_{\mathrm{UCB}}(\vec x, f, \mathcal{D})\mathcal{N}\left(f \bigg| \mu(\vec y), \frac{\lambda \pi}{2} \Sigma(\vec y) \right)p_\alpha(\vec y | \vec x, \mathcal{D}) d\vec ydf,
\end{align*}
where $\mu(\vec y), \Sigma(\vec y)$ are the predictive mean and covariance of $p(f | \mathcal{D} \cup \{(\vec x, \vec y)\})$.

\subsection{Conformalized Multi-Objective Acquisition Functions}
\label{subsec:multi_objective_bo}

When there are multiple objectives of interest, a single best design $\vec x^*$ may not exist.
Suppose there are $d$ objectives, $f^*: \mathcal{X} \rightarrow \mathbb{R}^d$.
The goal of multi-objective optimization (MOO) is to identify the set of \textit{Pareto-optimal} solutions such that improving one objective within the set leads to worsening another. 
We say that $\vec x$ dominates $\vec x'$, or $f^*(\vec{x}) \succ f^*(\vec{x}')$, if $f^*_k(\vec{x}) \geq f^*_k(\vec{x}')$ for all $k \in \{1, \dotsc, d\}$ and $f^*_k(\vec x) > f^*_k(\vec x')$ for some $k$.
The set of \textit{non-dominated} solutions $\mathscr{X}^*$ is defined in terms of the Pareto frontier (PF) $\mathcal{P}^*$,
\begin{align} 
\label{eq:pareto}
\mathscr{X}^\star = \{\vec{x}: f(\vec{x}) \in \mathcal{P}^\star\}, \hspace{4mm} \text{where } \mathcal{P}^\star = \{f(\vec{x}) \: : \: \vec x \in \mathcal{X}, \;  \nexists \: \vec{x}' \in \mathcal{X} \textit{ s.t. } f(\vec{x}') \succ f(\vec{x}) \}.
\end{align}

MOO algorithms typically aim to identify a finite approximation to $\mathscr{X}^\star$, which may be infinite, within a reasonable number of iterations.
One way to measure the quality of an approximate PF $\mathcal{P}$ is to compute the hypervolume ${\rm HV}(\mathcal{P} | \vec{r}_{\rm ref})$ of the polytope bounded by $\mathcal{P} \cup \{\vec r_{\mathrm{ref}}\}$, where $\vec r_{\mathrm{ref}} \in \mathbb{R}^d$ is a user-specified \textit{reference point}.
\begin{align} 
    u_{\mathrm{EHVI}}(\vec x, f, \mathcal{D}) = {\rm HVI}(\mathcal{P}', \mathcal{P} | \vec{r}_{\rm ref}) = [{\rm HV}(\mathcal{P}' | \vec{r}_{\rm ref}) - {\rm HV}(\mathcal{P} | \vec{r}_{\rm ref})]_+, \label{eq:ehvi}
\end{align}
where $\mathcal{P}' = \mathcal{P} \cup \{\hat f(\vec x)\}$ \citep{emmerich2005single,emmerich2011hypervolume, daulton2020differentiable}.
If our measurements of $f$ are noisy we cannot compute $\mathrm{HV}$ exactly and instead must substitute $\hat f \sim p(f | \mathcal{D})$, i.e. 
\begin{align} 
    u_{\mathrm{NEHVI}}(\vec x, f, \mathcal{D}) = {\rm HVI}(\hat{\mathcal{P}}_t', \hat{\mathcal{P}}_t | \vec{r}_{\rm ref}) \label{eq:nehvi},
\end{align}
where $\hat{\mathcal{P}}_t = \{\hat f(\vec{x}) \: : \: \vec x \in \mathcal{X}_t, \;  \nexists \: \vec{x}' \in \mathcal{X}_t \textit{ s.t. } \hat f(\vec{x}') \succ \hat f(\vec{x}) \}$ and $\hat{\mathcal{P}}' = \hat{\mathcal{P}} \cup \{\hat{f}(\vec{x})\}$ \citep{daulton2021parallel}.

Our derivations hold for so-called composite acquisitions as well, so we could also extend to qParEGO and qNParEGO variants for multi-objective optimization \citep{daulton2020differentiable,daulton2022robust}.

\subsection{Conformalizing Batch Acquisitions}\label{subsec:practical_extensions}

In general batch acquisitions have the form
\begin{align}
a(\vec x_0, \dots, \vec x_{q-1}, \mathcal{D}) = \int \max_{i < q} u(\vec x_i, f, \mathcal{D}) p(f | \mathcal{D})df. \label{eq:batch_acqs}
\end{align}
Note that $f(\vec x_0), \dots, f(\vec x_{q-1})$ are sampled jointly when estimating Eq. \eqref{eq:batch_acqs} with Monte Carlo methods.
Increasing the query batch size to $q$ increases the dimensionality of the outcome to $q \times d$, where $d$ is the number of objectives.
Our importance-sampling MC integration procedure introduced in
Section \ref{subsec:full_conformal_gps} scales gracefully with higher outcome dimensionality, we simply sample the elements of $Y_{\mathrm{cand}}$ from $p(y(\vec x_0), \dots, y(\vec x_{q-1}) | \vec x_{0:q-1}, \mathcal{D})$.

The bigger challenge arises in computing the conformal masks for batched query outcomes.
In our current implementation we compute the conformal scores (the Bayes posterior log-likelihood) pointwise for each query batch element, with corresponding pointwise conformal prediction masks.
We apply the pointwise masks before computing $\max_{i < q}u$ across query batch elements.
The alternative would be to compute a joint conformal score across all query batch elements (similarly computing joint scores for each of the previous query batches in the training data).
Note that this second approach essentially reduces to replacing each datum ($\vec x_i, \vec y_i$) in Eq. \eqref{eq:conf_pred_defn} with $(X_i, Y_i) = ([\vec x_0 \cdots \vec x_{q-1}]^\top, [\vec y_0 \cdots \vec y_q]^\top)$.
We leave the implementation of this second approach for future work.

\subsection{Out-of-distribution queries}

If $p'(\vec x | \mathcal{D}) \neq p(\vec x)$, and $w(\vec x, \mathcal{D}) > \alpha$, then every candidate label is automatically accepted to the prediction set and $\mathcal{C}_\alpha(\vec x) = \mathcal{Y}$, which makes $p_\alpha(\vec x | \mathcal{D})$ an improper density.
Intuitively the issue is there are not enough points in $\mathcal{D}$ close enough to $\vec x$ to guarantee a miscoverage rate of at most $\alpha$ unless every possible label is included in the prediction set.
Our solution is to set any conformal acquisition value $a_\alpha(\vec x, \mathcal{D})$ to 0 if $w(\vec x, \mathcal{D}) > \alpha$.
In practice we achieve this effect by introducing a second mask $m' = \texttt{sigmoid}(\tau^{-1}(w_n - \alpha))$ which we apply to Eq. \eqref{eq:conf_acq_mc_est}.
For conformal EI and other similar acquisition functions, this mask simply means we will favor any point close enough to the dataset to guarantee coverage if it has positive expected utility over any out-of-distribution point.
For conformal UCB, this mask means all out-of-distribution points are assigned the value $1/n \sum_{\vec y_i \in \mathcal{D}} \vec y_i$ (assuming the labels have been mean-centered during preprocessing).

\subsection{Generating target functions with known RKHS norm}
\label{subsec:gen_target_fn}

\begin{figure}
    \centering
    \begin{subfigure}{0.4\textwidth}
        \includegraphics[width=\textwidth]{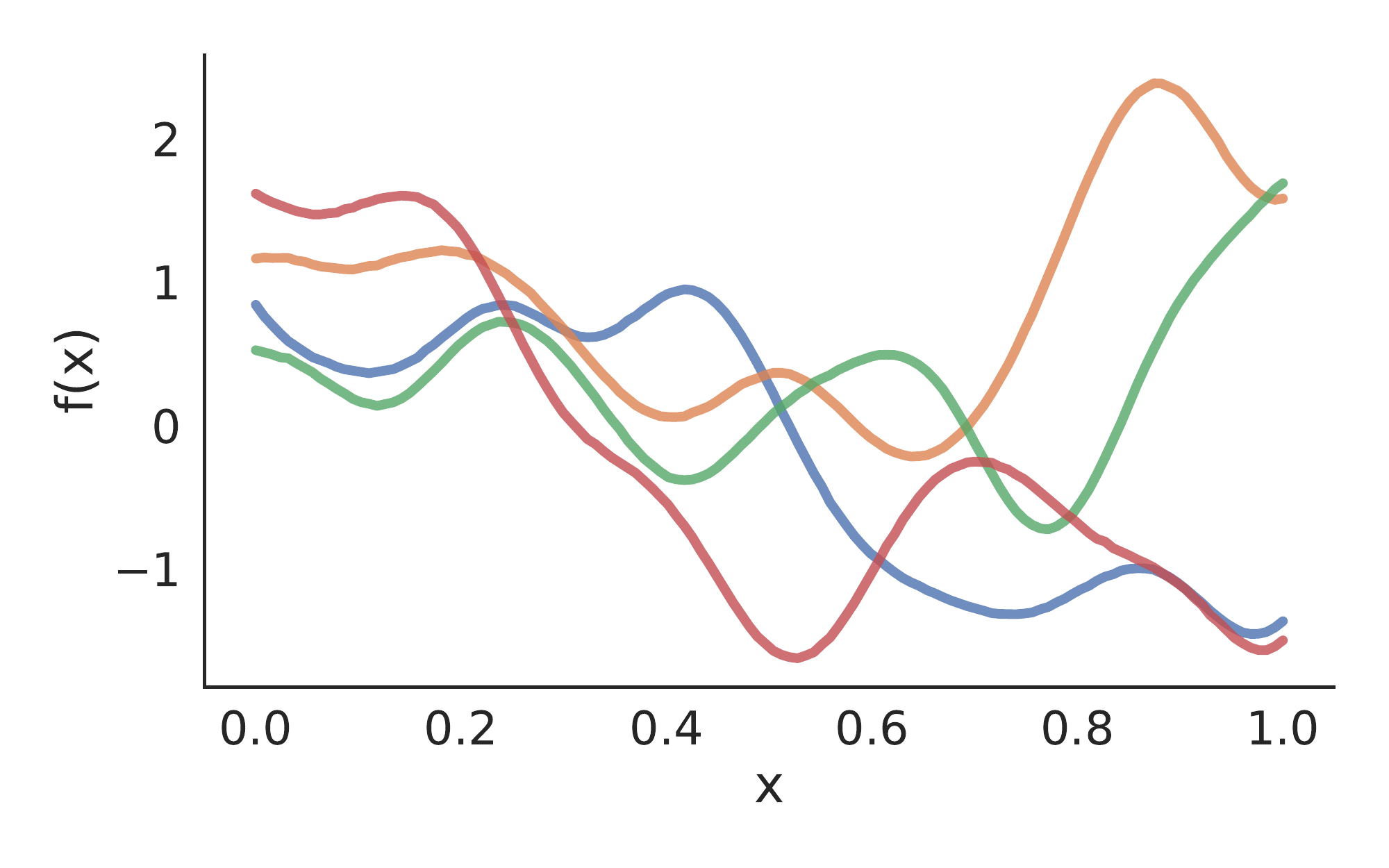}
        \caption{}
        \label{subfig:matern_gp_prior_draws}
    \end{subfigure}
    \begin{subfigure}{0.4\textwidth}
        \includegraphics[width=\textwidth]{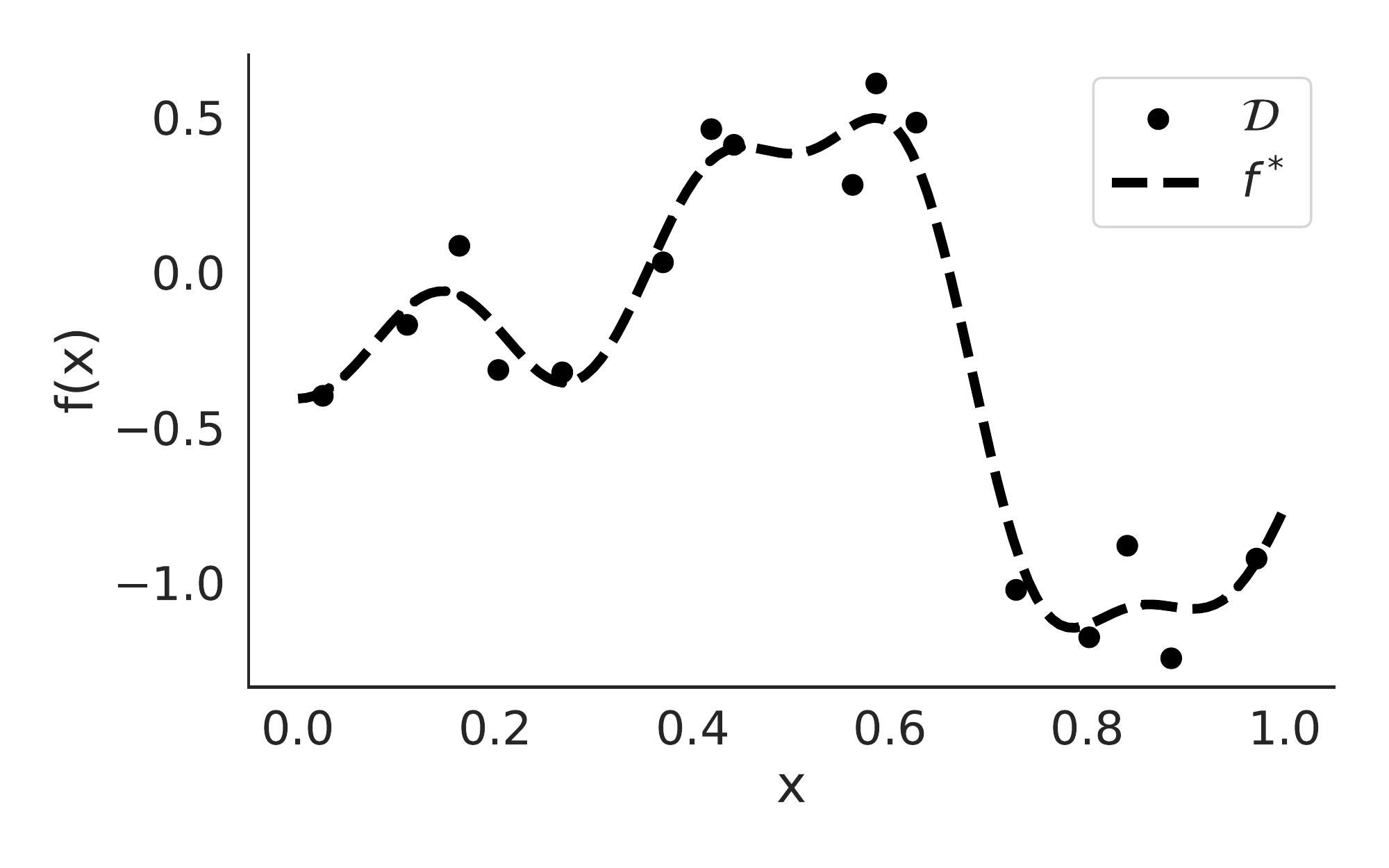}
        \caption{}
        \label{subfig:target_fn_known_norm}
    \end{subfigure}
    \caption{
        Here we demonstrate how to generate target functions with known RKHS norm w.r.t. a given kernel $\kappa$, in this case an RBF kernel with lengthscale $\ell = 0.1$ on $\mathcal{X} = [0, 1]$.
        In the \textbf{left} panel we show prior draws $f^{(i)} \sim \mathcal{GP}(0, \kappa)$.
        In the \textbf{right} panel we show a synthetic target function $f^*$ with corresponding RKHS norm $||f^*||_\kappa = 2.0934$.
        We produced $f^*$ according to Algorithm \ref{alg:gen_fn_known_norm}, using $n=16$ basis points with dimension $d=1$ and noise variance $\sigma^2 = 0.04$.
    }
    \label{fig:generating_target_fns}
\end{figure}

As we noted in Section \ref{sec:related_work}, many formal regret bounds for BayesOpt rest on the assumption that 1) the RKHS norm of $f^*$ corresponding to the choice of kernel $\kappa$ is bounded (i.e. $||f^*||_\kappa < \infty$) and 2) that we know a reasonable upper bound on the RKHS norm.
Although these assumptions almost certainly do \textit{not} hold for every $f^*$ encountered in practice, it is worthwhile to examine the behavior of conformal acquisition functions when the idealized assumptions do hold, to see what price we pay for robustness. 

Although it is enough to know a bound on $||f^*||_\kappa$ in this section we will optimize target functions for which we know the RKHS norm exactly, to remove slack in the bound as a potential experimental confounder. 
A simple approach, which we summarize in Algorithm \ref{alg:gen_fn_known_norm}, is to choose a random set of $n$ basis points $X = \{\vec x_0, \dots \vec x_{n-1}\}$, draw random function values from the GP prior $\vec u \sim \mathcal{N}(\vec 0, \kappa(X, X))$ (Figure \ref{subfig:matern_gp_prior_draws}), draw noisy labels $\vec y \sim \mathcal{N}(\vec u, \sigma^2 I_n)$, and take $f^*$ to be the GP posterior mean $\mu_{f | \mathcal{D}}$ conditioned on the synthetic dataset $\mathcal{D} = (X, \vec y)$ (Figure \ref{subfig:target_fn_known_norm}).

\begin{algorithm}[t]
\caption{Generating target functions with known RKHS norm}
\label{alg:gen_fn_known_norm}
\SetAlgoLined
    \KwData{kernel $\kappa$, \# basis points $n$, input dimension $d$, noise variance $\sigma^2$}
    
    Sample basis points $X = \{\vec x_0, \dots, \vec x_{n-1}\} \sim \texttt{SobolSequence}(n, d)$
    
    Sample $\vec u = \{u_0, \dots, u_{n-1}\} \sim \mathcal{N}(\vec 0, \kappa(X, X))$ \hfill (GP prior draw with kernel $\kappa$)
    
    Sample $\vec y = \{y_0, \dots, y_{n-1}\} \sim \mathcal{N}(\vec u, \sigma^2 I_n)$
    
    Compute $\vec c = (K_{XX} + \sigma^2 I)^{-1}\vec y$
    
    $f^*(\cdot) = \kappa(\cdot, X)\vec c$ \hfill (GP posterior mean with kernel $\kappa$ conditioned on $\mathcal{D}$)
    
    $B = \sqrt{\vec c^\top K_{XX} \vec c}$ \hfill (RHKS norm $||f^*||_\kappa$)
    
    \KwResult{$f^*, B$}
\end{algorithm}

The proof that $\mu_{f | \mathcal{D}} \in \mathcal{H}_\kappa(X)$ and $||\mu_{f | \mathcal{D}}||_\kappa = \sqrt{\vec c^\top K_{XX} \vec c}$ follows directly from the definition of an RKHS (see Chapter 2.2 in \citet{scholkopf2018learning}).
Although these are well-established results, we sketch the proof here for the reader's convenience.

\textbf{Proof:}
Recall that by definition, the RKHS corresponding to $\kappa$ with basis points $X$ can be written as 
\begin{align}
    \mathcal{H}_\kappa(X) = \overline{\{f \; : \: f(\cdot) = \sum_{i=0}^{n-1} c_i \kappa(\cdot, \vec x_i) \text{ for some } \vec c \in \mathbb{R}^n \}} \label{eq:rkhs_kernel_expansion},
\end{align}
where $\overline{\mathcal{S}}$ is the closure of $\mathcal{S}$.

Since the GP posterior mean conditioned on $\mathcal{D}$ is given by
\begin{align*}
    \mu_{f | \mathcal{D}}(\cdot) = \kappa(\cdot, X)(K_{XX} + \sigma^2 I)^{-1}\vec y,
\end{align*}
and $\vec c = (K_{XX} + \sigma^2 I)^{-1}\vec y$ is in $\mathbb{R}^n$, we've verified that $\mu_{f | \mathcal{D}} \in \mathcal{H}_\kappa(X)$.

For any $f(\cdot) = \sum_{i=0}^{n-1} c_i \kappa(\cdot, \vec x_i)$ for some $\vec c \in \mathbb{R}^n$ we can derive the corresponding RKHS norm as follows:
\begin{align}
    ||f||_{\kappa}^2 &= \langle f, f \rangle_\kappa, \nonumber \\
    &= \left \langle \sum_{i=0}^{n-1} c_i \kappa(\cdot, \vec x_i), \sum_{j=0}^{n-1} c_j \kappa(\cdot, \vec x_j) \right \rangle_\kappa, \nonumber \\
    &= \sum_{i,j=0}^{n-1} c_i c_j \left \langle \kappa(\cdot, \vec x_i), \kappa(\cdot, \vec x_j) \right \rangle_\kappa, \nonumber \\
    &= \sum_{i,j=0}^{n-1} c_i c_j \kappa(\vec x_i, \vec x_j), \nonumber \\
    \Rightarrow ||f||_\kappa &= \sqrt{\vec c^\top K_{XX} \vec c}.
\end{align}
The second line follows from the linearity of $\langle \cdot, \cdot \rangle_\kappa$, and the third line follows from the reproducing property of $\mathcal{H}_\kappa(X)$ and symmetry of $\kappa$. $\hfill \blacksquare$

\textbf{Remark:} we are not required to use $\mu_{f | \mathcal{D}}$, once we have chosen $\kappa$ and $X$ we could in principle take any linear combination of $\{k(\cdot, \vec x_0), \dots, k(\cdot, \vec x_{n-1})\}$, e.g. by sampling $\vec c \sim \mathcal{N}(\vec 0, I)$.
However $\mu_{f | \mathcal{D}}$ is convenient and sufficient for our purposes.

When generating target functions in this way, it is important to account for the dimensionality of the input $d$ when choosing the number of basis points $n$. 
Due to the curse of dimensionality, the number of points needed to produce ``interesting'' functions (i.e. functions that are not flat almost everywhere) grows exponentially with the dimension of the input when using default GP kernels (e.g. Mat\'ern).

\clearpage
\section{ADDITIONAL EXPERIMENTAL RESULTS}
\label{app:experiment_results}

\subsection{Single-Objective Black-Box Optimization}
\label{subsec:bbo_supp_results}

\begin{figure}[!h]
    \centering
    \begin{subfigure}{0.8\textwidth}
    \centering
    \includegraphics[width=\textwidth,clip,trim=0cm 0.2cm 0cm 13.75cm]{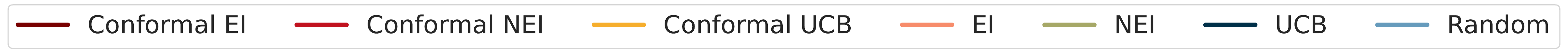}
    \end{subfigure}
    \\
    \centering
    \begin{subfigure}{0.24\textwidth}
\includegraphics[width=\textwidth]{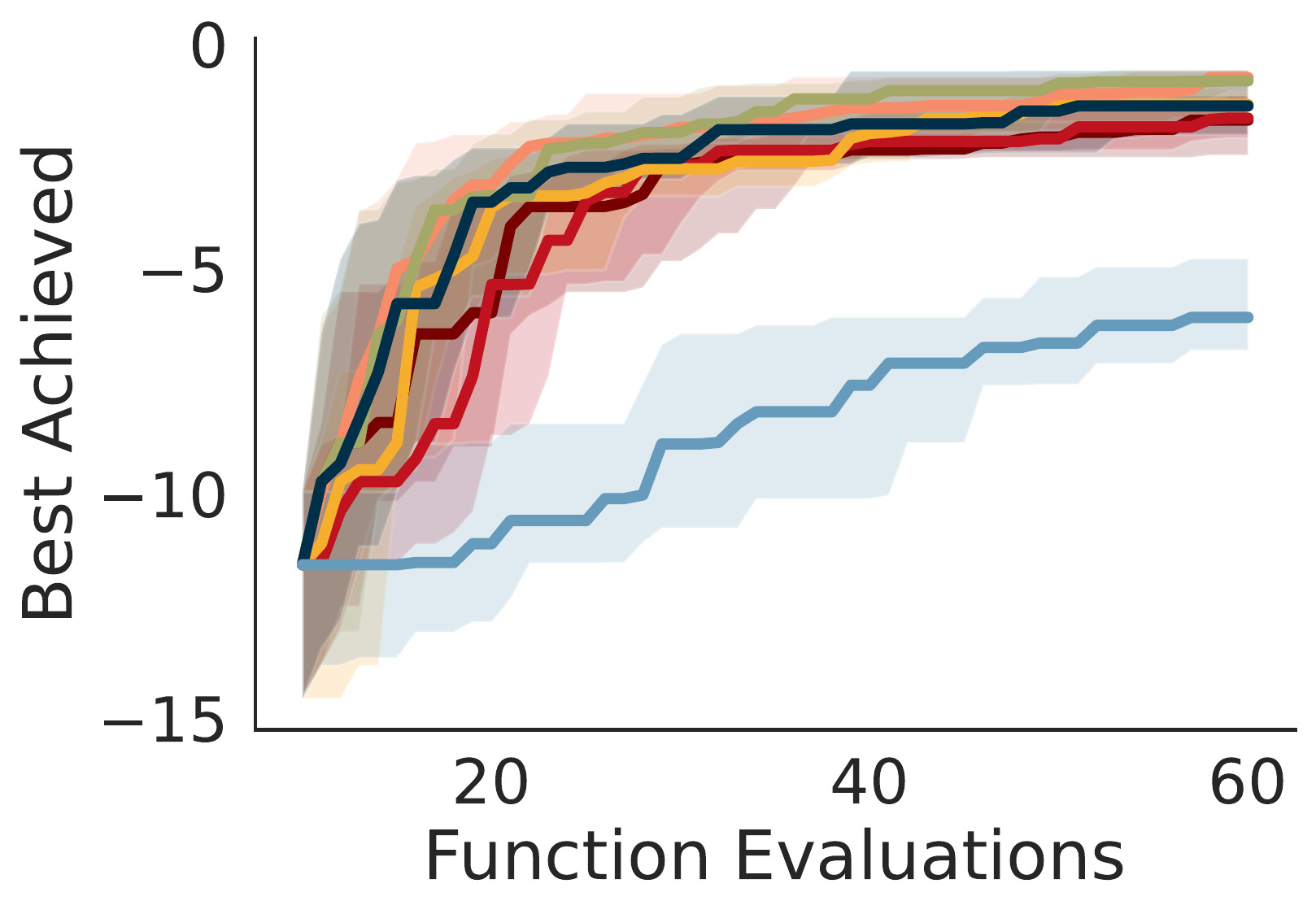}
    \caption{Levy-$5$, $q=1.$}
    \label{fig:levy5q1}
    \end{subfigure}
        \begin{subfigure}{0.24\textwidth}
\includegraphics[width=\textwidth]{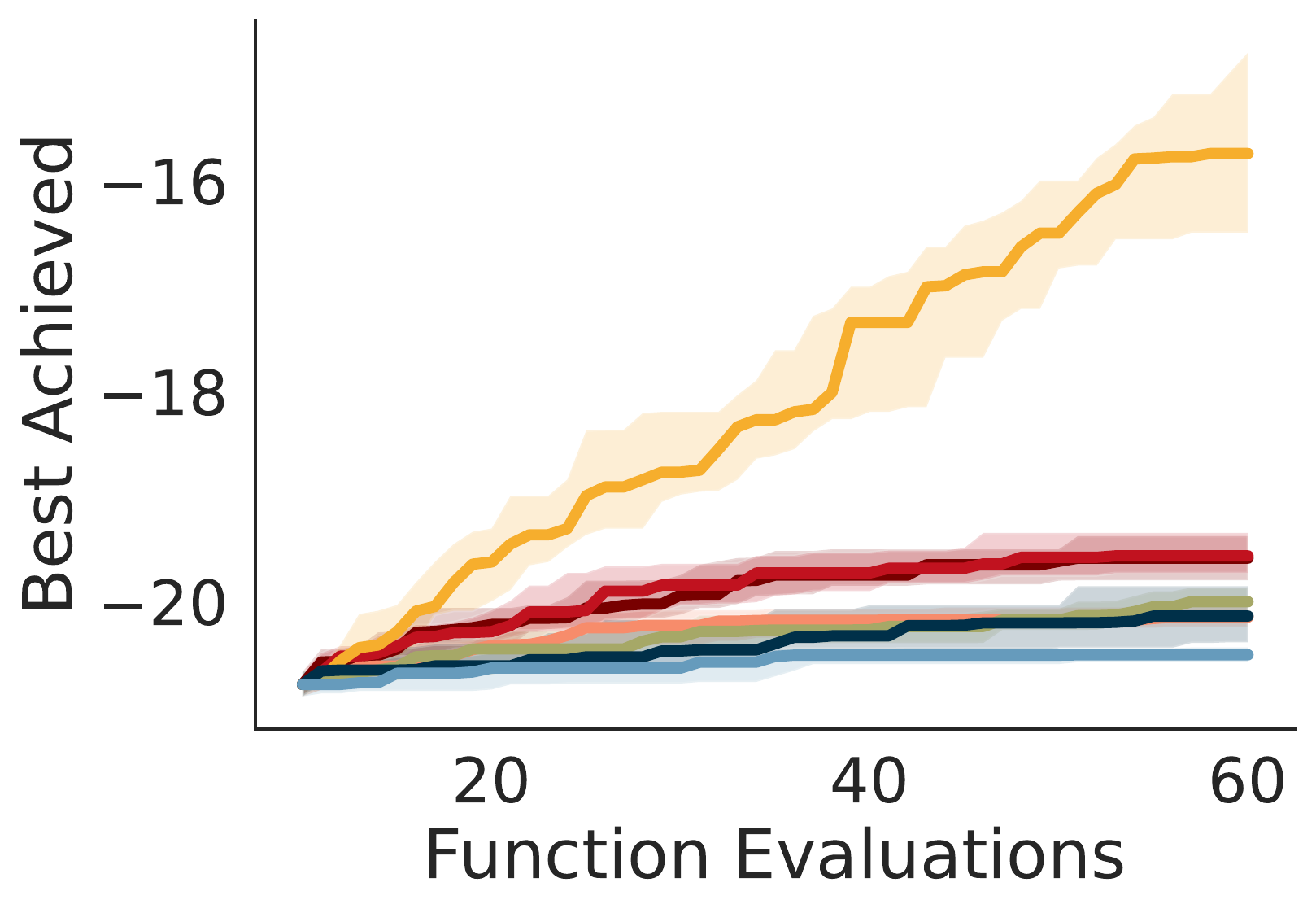}
    \caption{Ackley-$20$, $q=1.$}
    \label{fig:ackley20q1}
    \end{subfigure}
        \begin{subfigure}{0.24\textwidth}
\includegraphics[width=\textwidth]{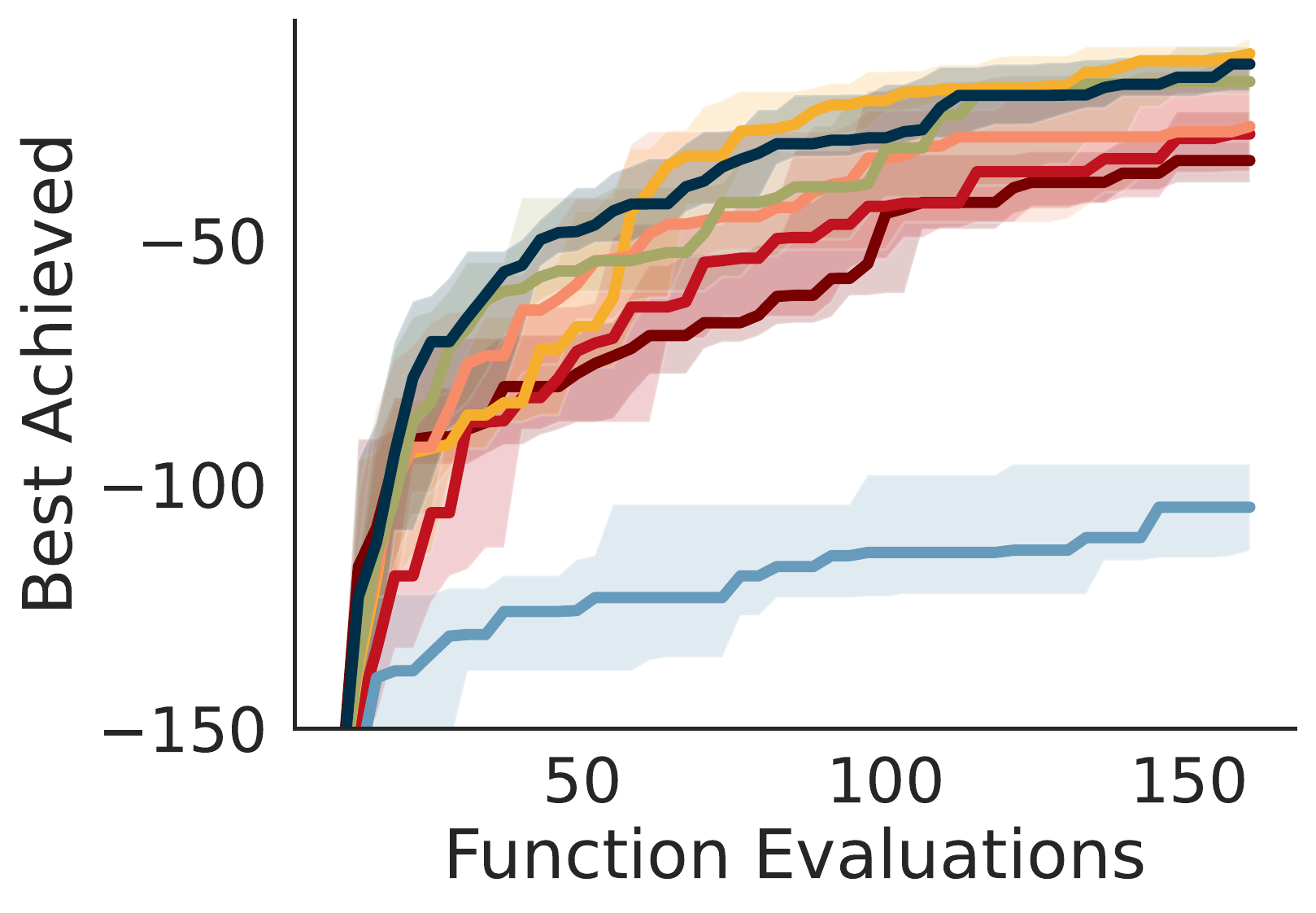}
    \caption{Levy-$20$, $q=3.$}
    \label{fig:6c}
    \end{subfigure}
        \begin{subfigure}{0.24\textwidth}
\includegraphics[width=\textwidth]{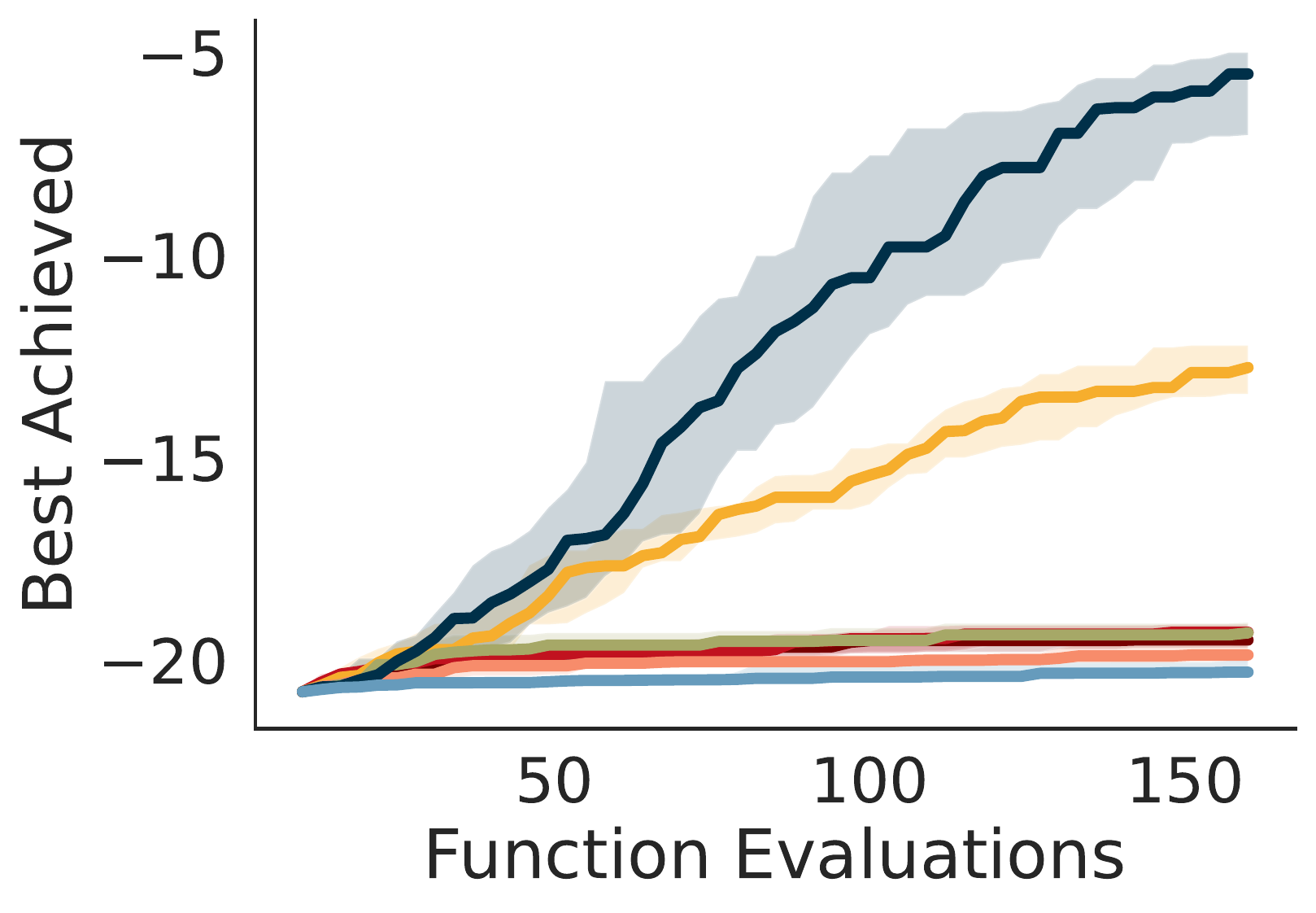}
    \caption{Ackley-$20$, $q=3.$}
    \label{fig:6d}
    \end{subfigure}
    \caption{
    BayesOpt best objective value found with conformal and standard acquisition functions on single-objective tasks Levy-$d$ and Ackley-$d$ (reporting median and its $95\%$ conf. interval, estimated from $25$ trials).
    qEI, qNEI, conformal qEI, and conformal qNEI all perform similarly, conformal qUCB is best everywhere except Ackley-20, where it comes second after qUCB.
    }
    \label{fig:std_bayesopt_oldversion}
\end{figure}

\begin{figure}[!h]
    \centering
    \begin{subfigure}{\textwidth}
    \centering
    \includegraphics[width=0.96\textwidth]{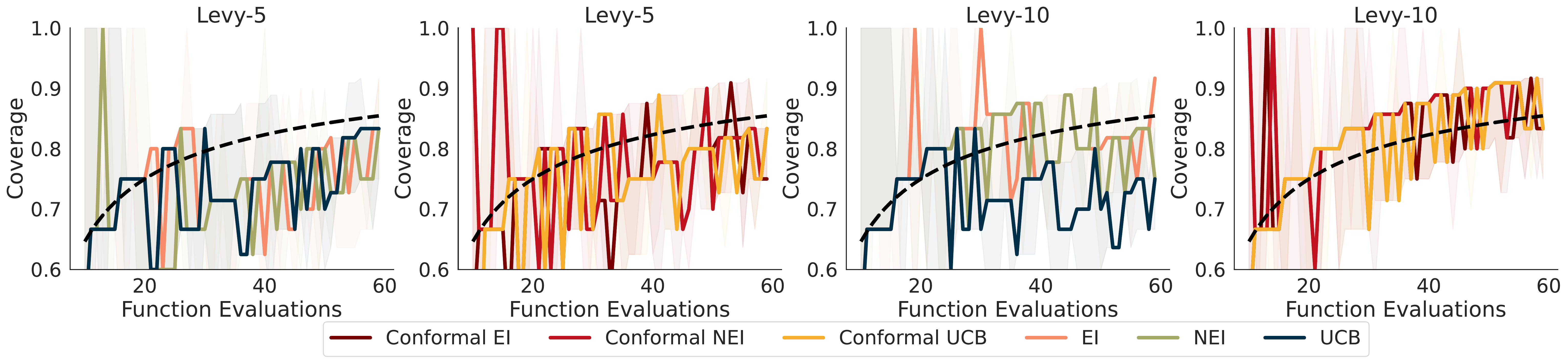}
    \caption{Levy $5$ and $10$, $q=1.$}
    \end{subfigure}
    \begin{subfigure}{\textwidth}
    \centering
    \includegraphics[width=0.96\textwidth]{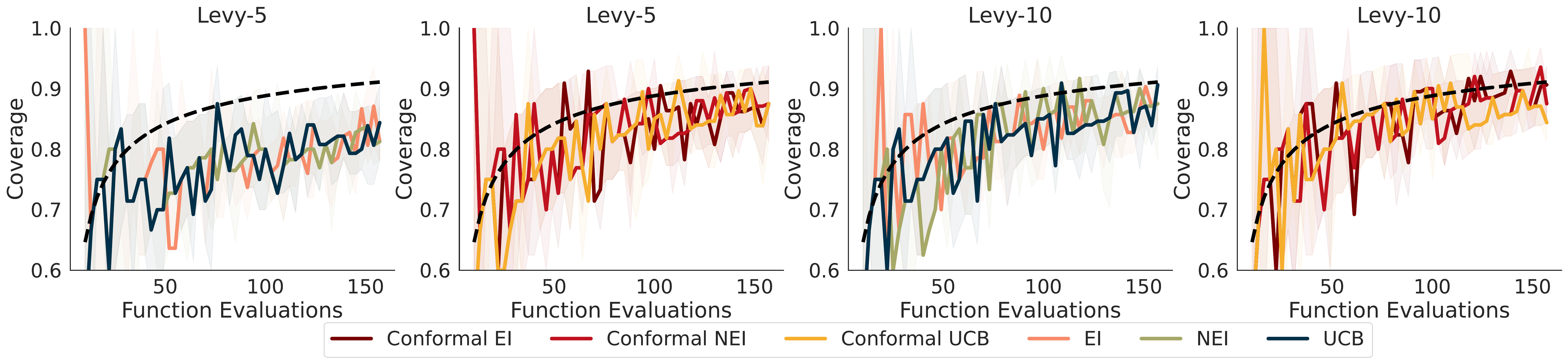}
    \caption{Levy $5$ and $10$, $q=3.$}
    \end{subfigure}
    \caption{
    BayesOpt empirical coverage of conformal and credible prediction sets
    evaluated on holdout data from single-objective task Levy-$d$ (reporting median and its $95\%$ conf. interval, estimated from $25$ trials).
    The conformal coverage curves track the target $1-\alpha$ (black dashed line) well, significantly
    better than the credible curves, which tend to be overconfident.
    Median w/ $95\%$ confidence interval is shown.
    }
    \label{fig:levy_coverages_q1_oldversion}
\end{figure}

In Figure \ref{fig:std_bayesopt_oldversion} we investigate the effect of the choice of acquisition function on sample-efficiency, comparing conventional and conformal versions.
In particular we consider expected improvement (EI), noisy expected improvement (NEI) and upper confidence bound (UCB) alongside their conformal counterparts.
No clear ranking emerges here, however UCB and conformal UCB both perform well in general.

In Figure \ref{fig:levy_coverages_q1_oldversion}, we investigate the sensitivity of coverage on random holdout data to the query batch size $q$ and the dimensionality of the inputs $d$.
Here, we plot the median and its $95\%$ confidence interval as shading, finding that the conformal sets are better calibrated in a frequentist sense than Bayesian credible sets.

\newpage
\subsection{Multi-Objective Black-Box Optimization}

\begin{figure}[!h]
    \centering
    \includegraphics[width=0.96\textwidth]{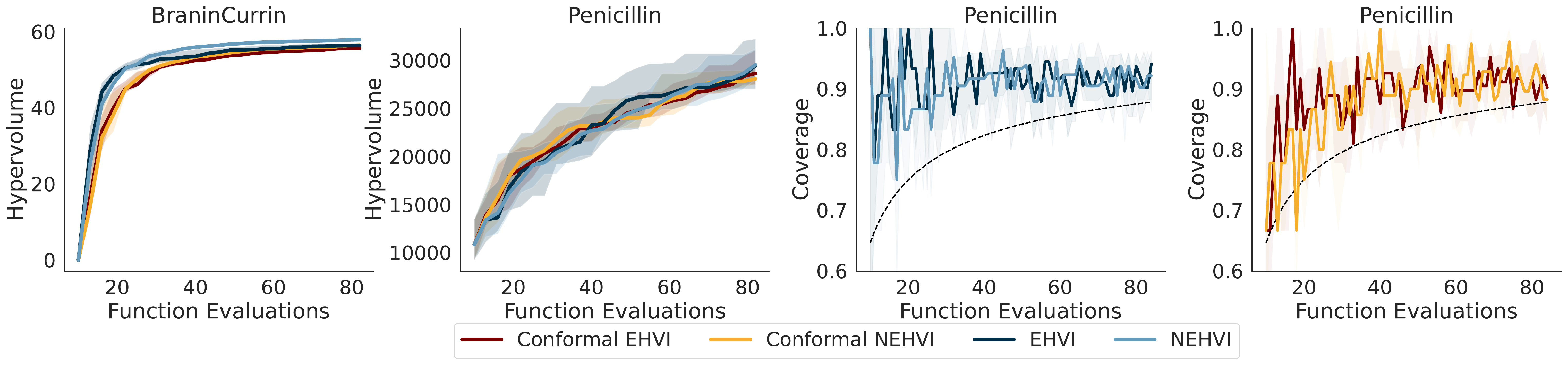}
    \caption{
    BayesOpt results on multi-objective tasks \texttt{branin-currin} and \texttt{penicillin} (reporting median and its $95\%$ conf. interval, estimated from $25$ trials).
    \textbf{Left two panels: }  Both conformal and standard acquisitions find solution sets with similar hypervolumes. 
    \textbf{Right two panels: } Credible and conformal empirical coverage curves. The conformal curves track the target $1 - \alpha$ (black dashed line) better than the credible curves, but both are underconfident.
    }
    \label{fig:std_mobo}
\end{figure}

To demonstrate that our approach scales to multi-objective tasks, we consider two tasks, \texttt{branin-currin} ($d=2$) and \texttt{penicillin} ($d=3$) \citep{liang2021scalable}.
The goal is not to find a single $\vec x^*$, but rather to find the set of all non-dominated solutions, the Pareto front (Appendix \ref{subsec:multi_objective_bo}).
By non-dominated, we mean the set of solutions with the property that the objective value cannot increase in one dimension without decreasing in another.
We report results using the expected hypervolume improvement (EHVI) \citep{emmerich2005single,emmerich2011hypervolume,daulton2020differentiable} and noisy expected hypervolume improvement (NEHVI) \citep{daulton2021parallel} as the base acquisition functions in Figure \ref{fig:std_mobo}.
Like the single-objective case conformal BayesOpt is comparable in terms of sample-efficiency as quantified by the solution hypervolume relative to a common reference point \citep{beume2009complexity}, and conformal set coverage tracks $(1 - \alpha)$ more closely than credible set coverage.
All black-box functions used in this paper are synthetic with implementations coming from BoTorch \citep{balandat_botorch_2020}. 
The Penicillin function was originally proposed by \citet{liang2021scalable}.

In this setting the performance of conventional BayesOpt and conformal BayesOpt is very similar in terms of solution quality, and the improvement to coverage is fairly small (Figure \ref{fig:std_mobo}).
The root issue appears to be that it is simply much more difficult to accurately characterize conformal prediction sets in multiple dimensions, since intervals become polyhedra \citep{johnstone2022exact}.
Although our Bayesian discretization scheme avoids the exponential memory usage of dense grids, its ability to pinpoint the boundary of conformal prediction sets appears to degrade as the dimensionality of $\vec y$ increases.

\newpage 
\subsection{Tabular Ranking with Real-World Drug and Antibody Data}
\label{subsec:tab_ranking_full}

\begin{figure}[!h]
    \centering
    \begin{subfigure}{0.2\textwidth}
    \centering
    \includegraphics[width=\textwidth,clip,trim=0cm 0.2cm 0cm 0cm]{figures/tab_bandits_cum-regret_legend_v0.0.4.pdf}
    \end{subfigure}
    \\
        \begin{subfigure}{0.21\textwidth}
\includegraphics[width=\textwidth]{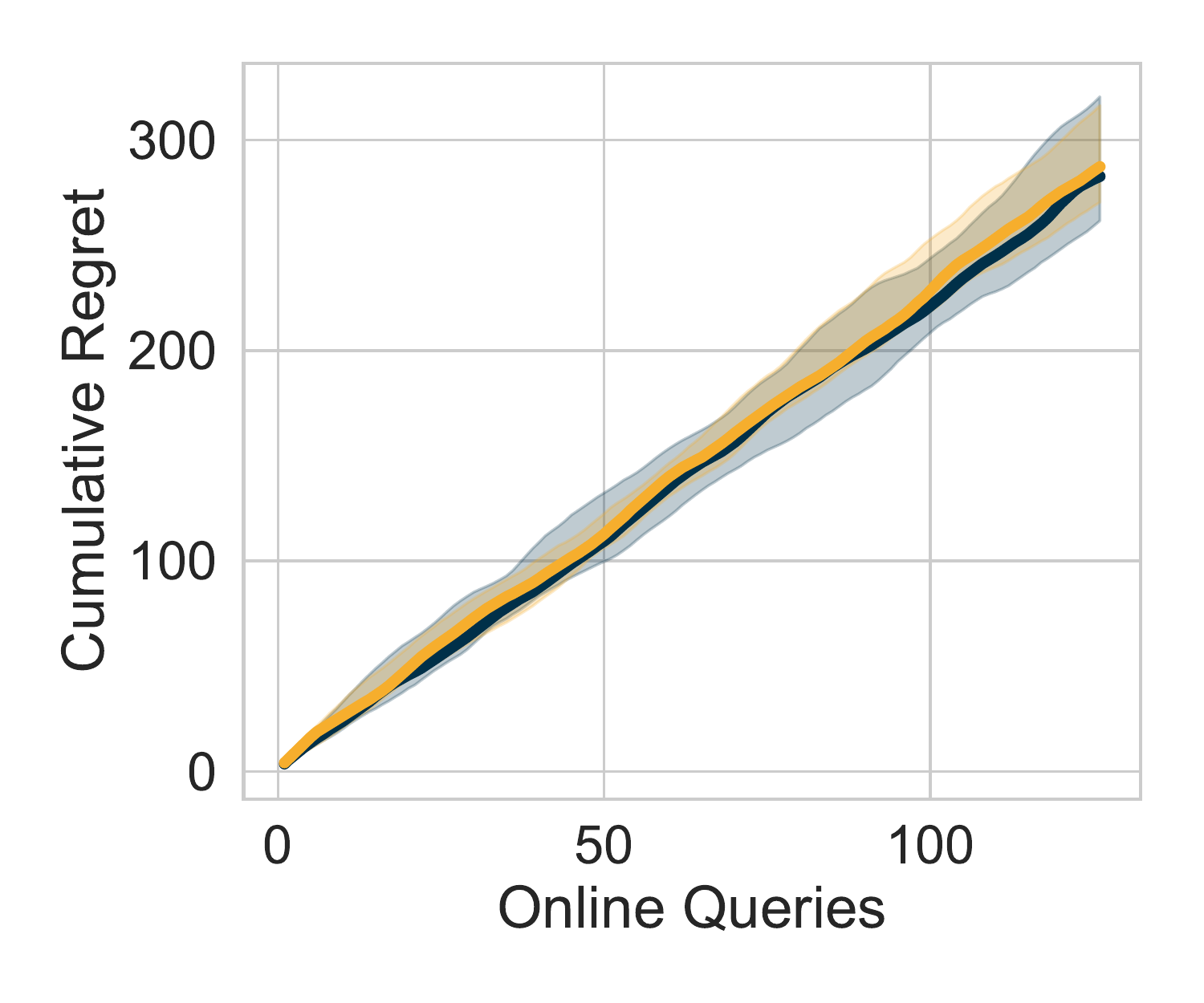}
    \label{fig:std_bayesopt_q3_levy_dop}
    \end{subfigure}
            \begin{subfigure}{0.21\textwidth}
\includegraphics[width=\textwidth]{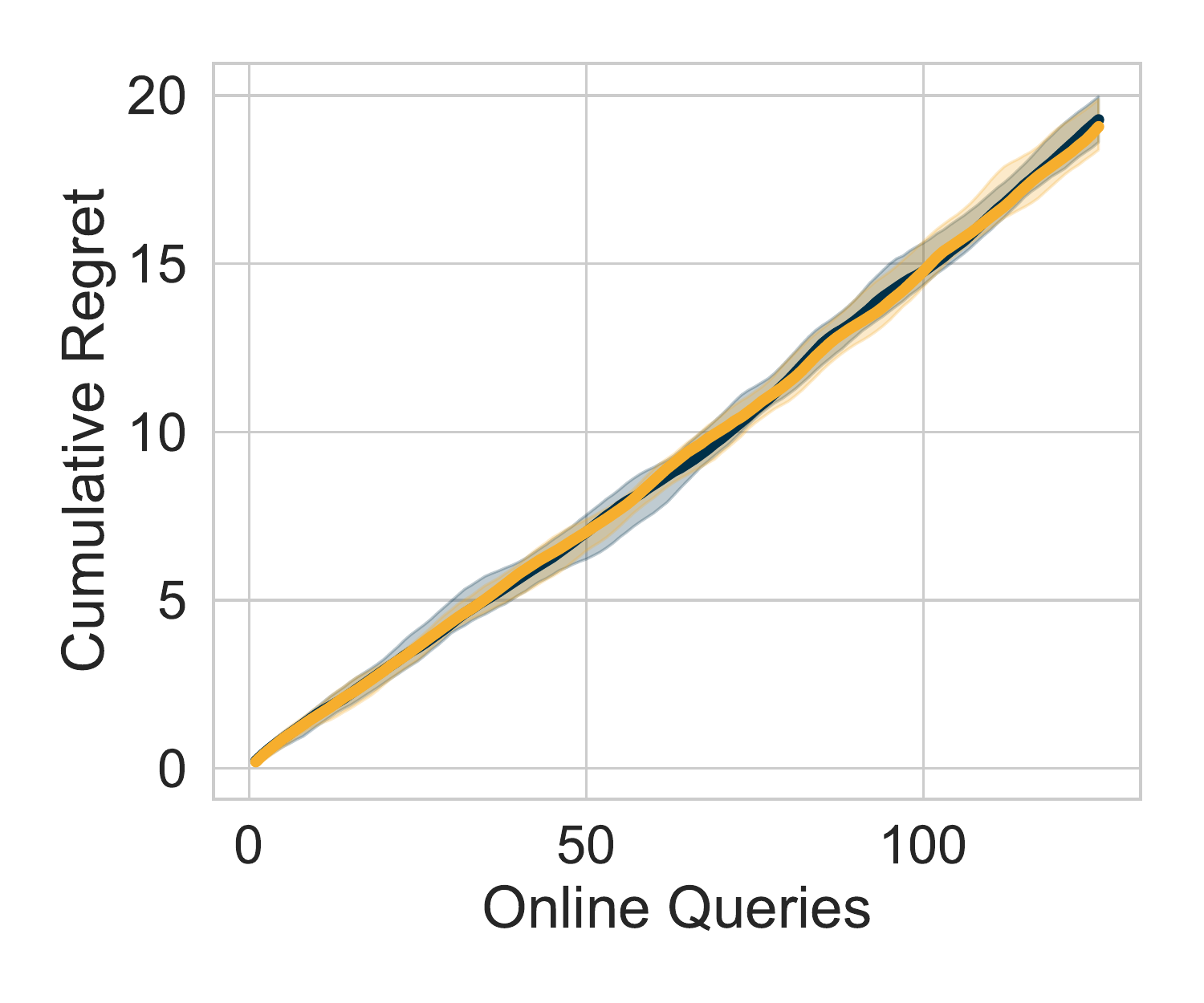}
    \label{fig:as;ldk}
    \end{subfigure}   
            \begin{subfigure}{0.21\textwidth}
\includegraphics[width=\textwidth]{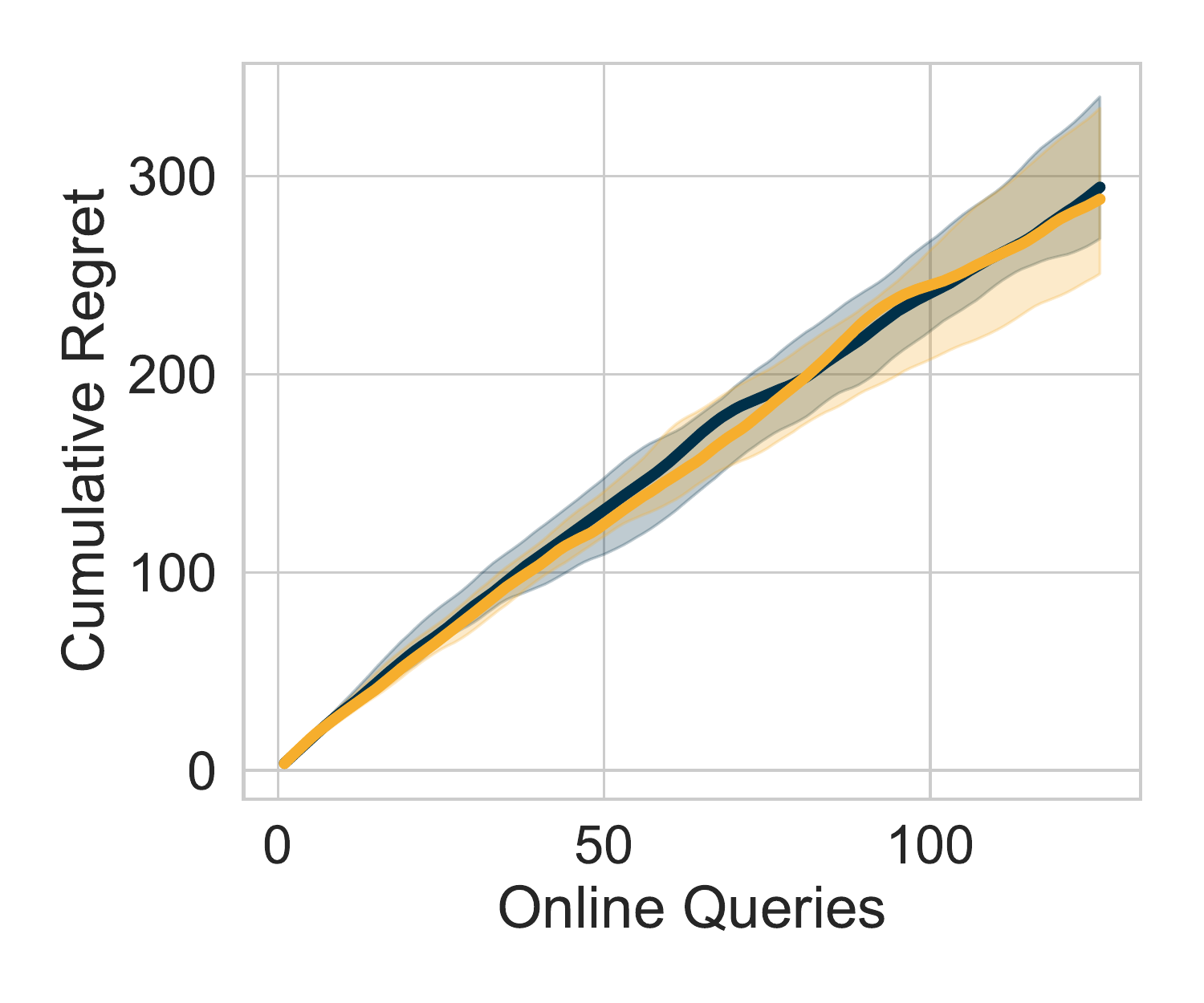}
    \label{fig:dkdkdk}
    \end{subfigure}   
\begin{subfigure}{0.21\textwidth}
    \includegraphics[width=\textwidth]{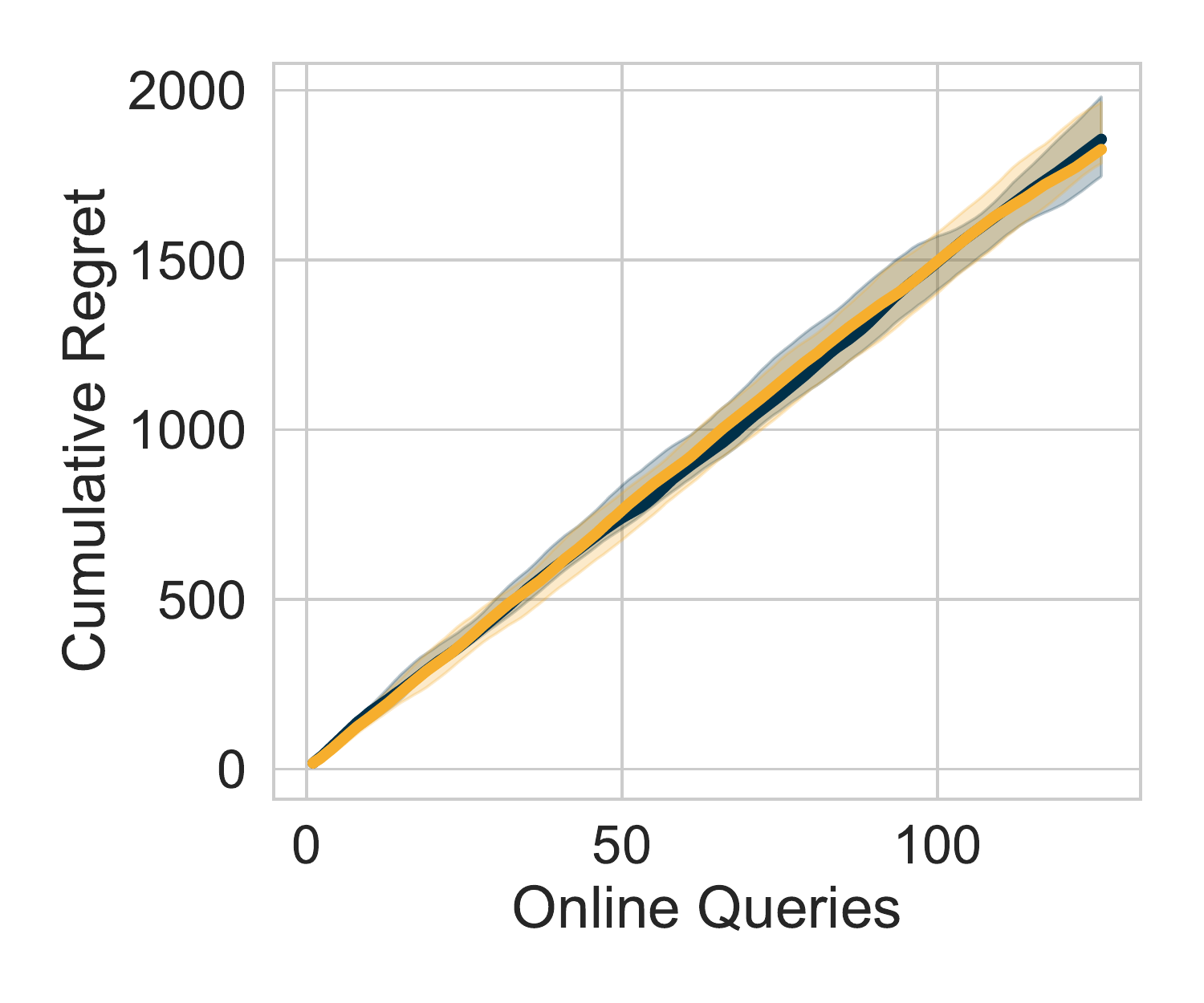}
    \label{fig:dkdk}
\end{subfigure}
        \begin{subfigure}{0.21\textwidth}
\includegraphics[width=\textwidth]{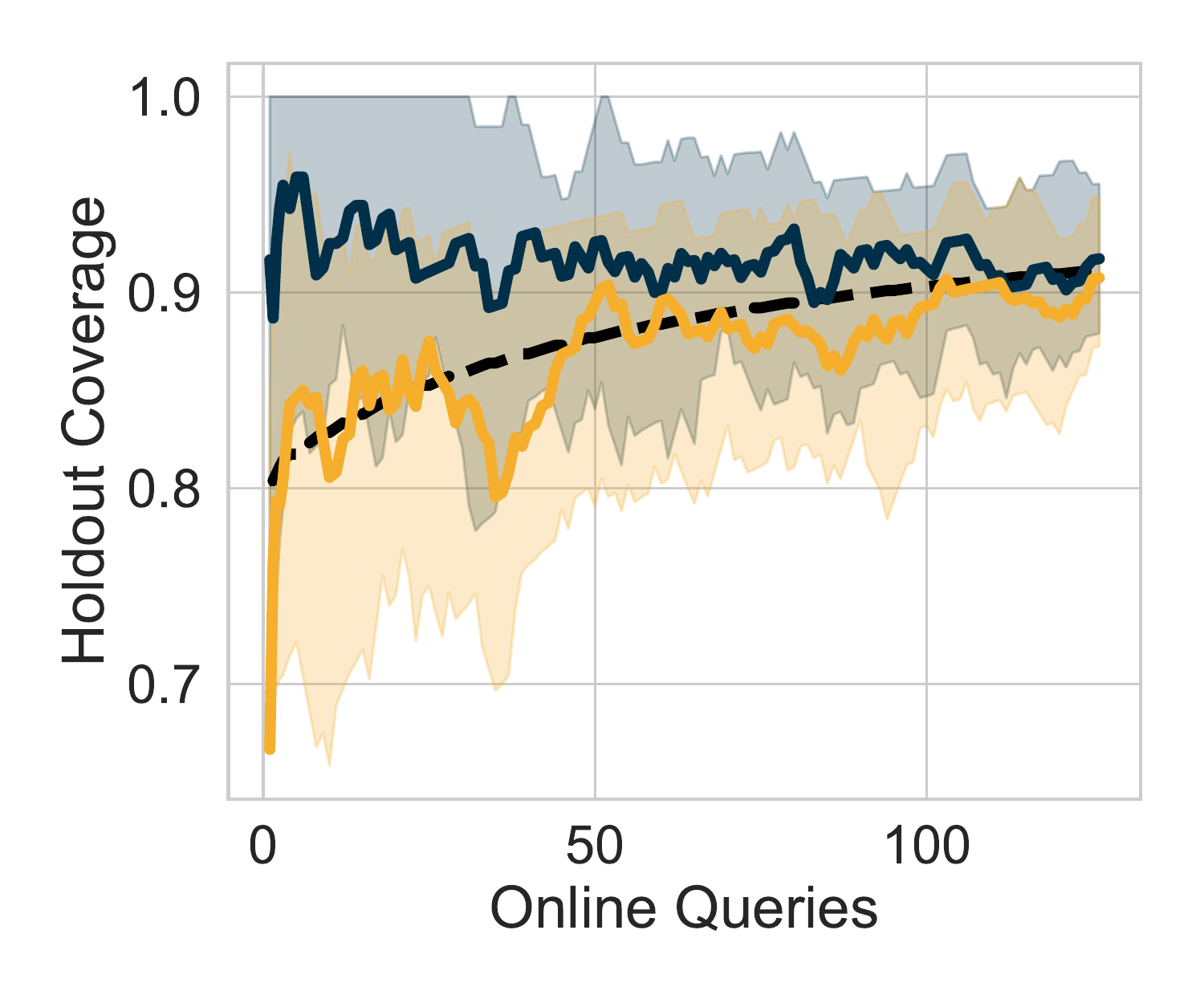}
    \label{fig:fjfrjfjfj}
    \end{subfigure}
            \begin{subfigure}{0.21\textwidth}
\includegraphics[width=\textwidth]{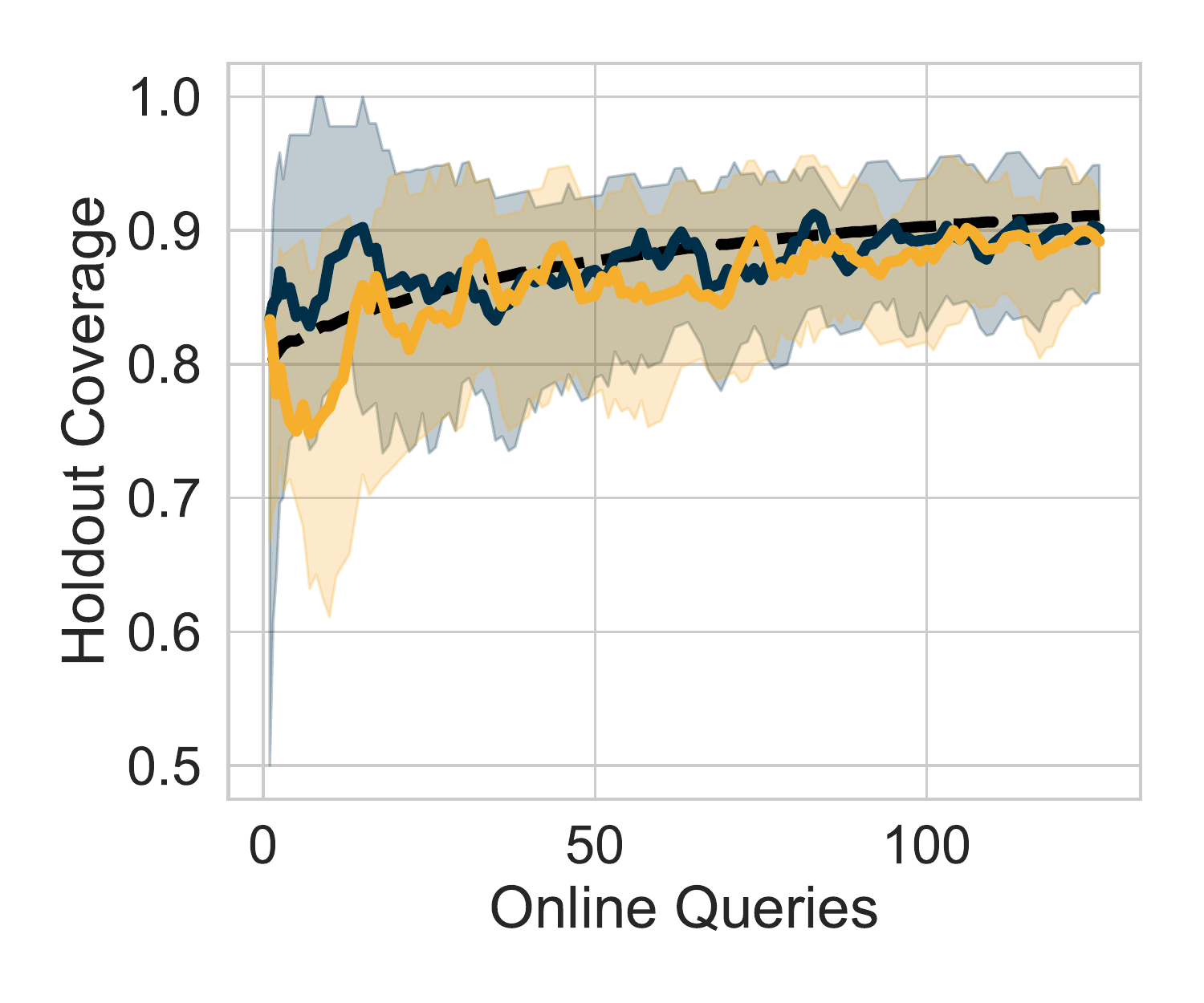}
    \label{fig:fjfjfjf}
    \end{subfigure}   
            \begin{subfigure}{0.21\textwidth}
\includegraphics[width=\textwidth]{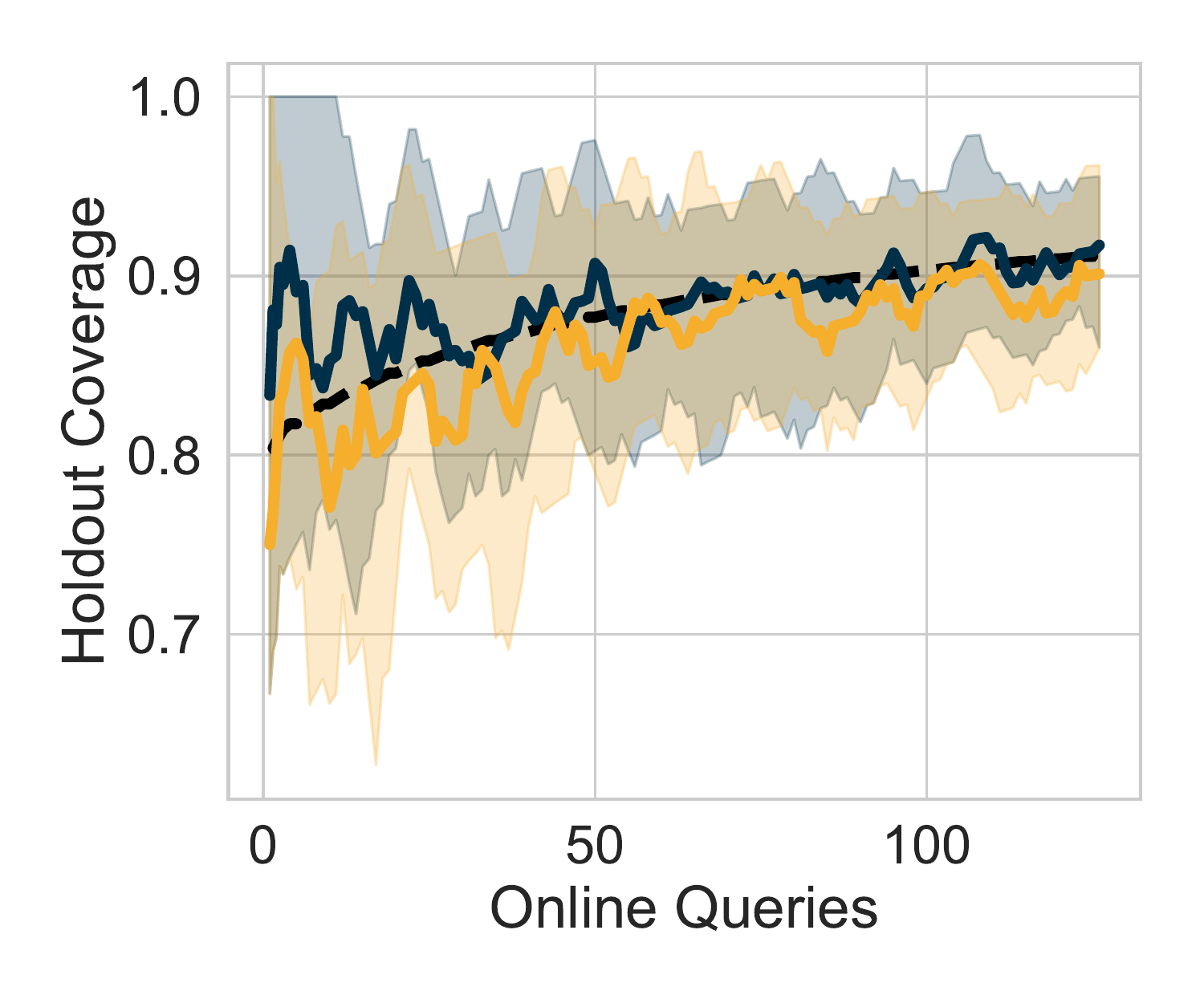}
    \label{fig:askdlf}
    \end{subfigure}   
\begin{subfigure}{0.21\textwidth}
    \includegraphics[width=\textwidth]{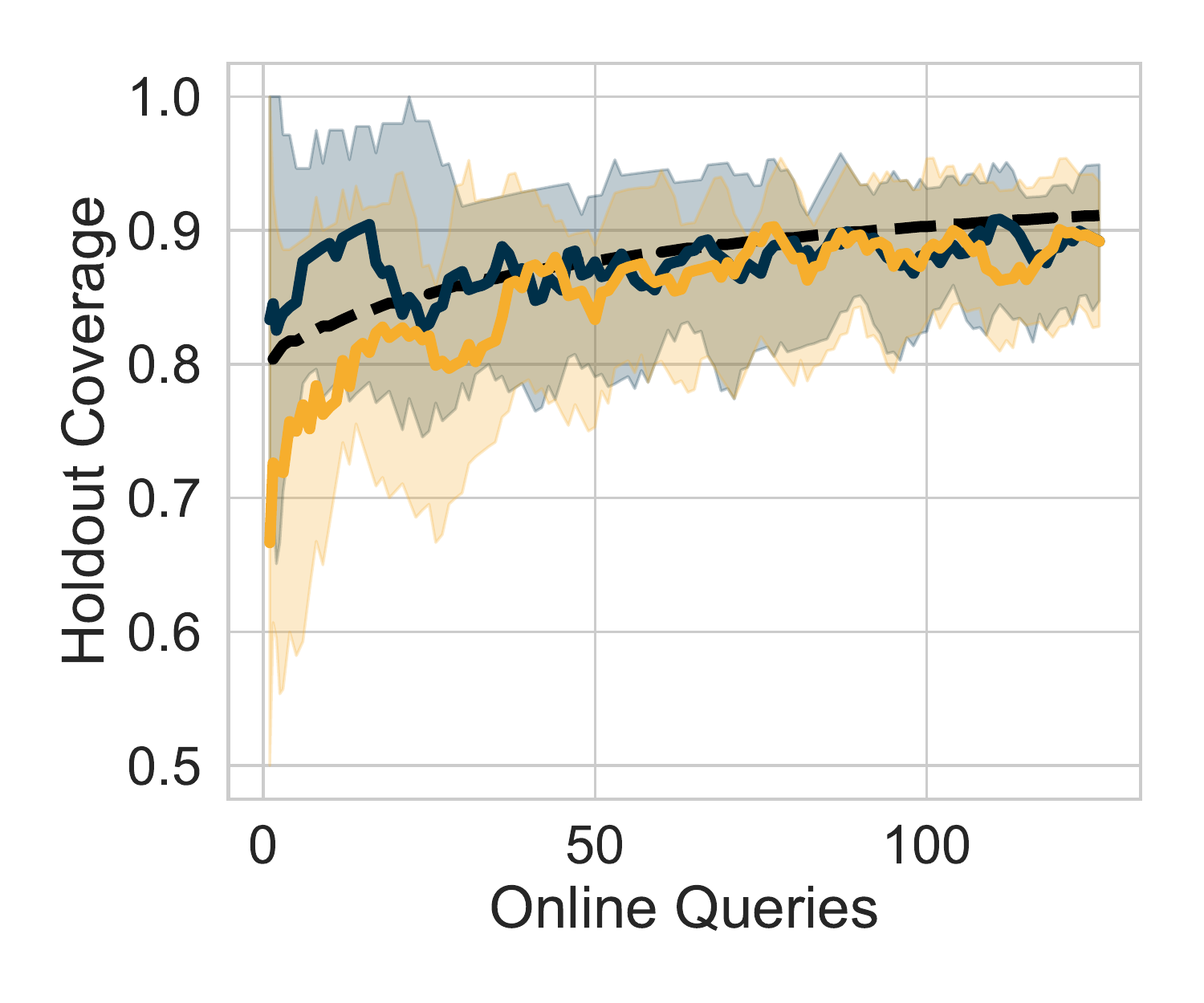}
    \label{fig:std_bayesopt_q3_levy}
\end{subfigure}
    \begin{subfigure}{0.21\textwidth}
    \includegraphics[width=\textwidth]{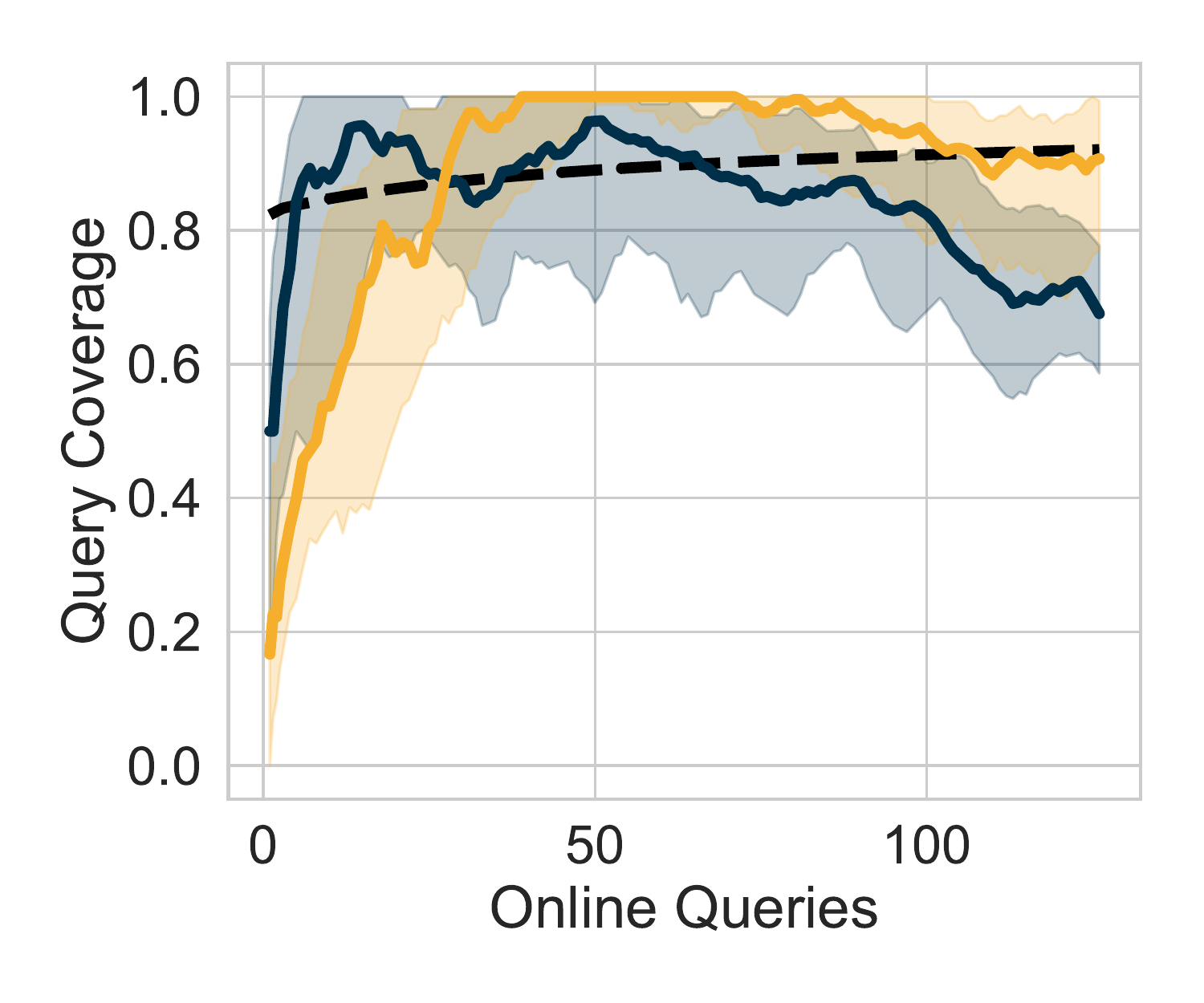}
    \caption{penalized logP}
    \label{fig:tab_bandit_plogp}
    \end{subfigure}
            \begin{subfigure}{0.21\textwidth}
\includegraphics[width=\textwidth]{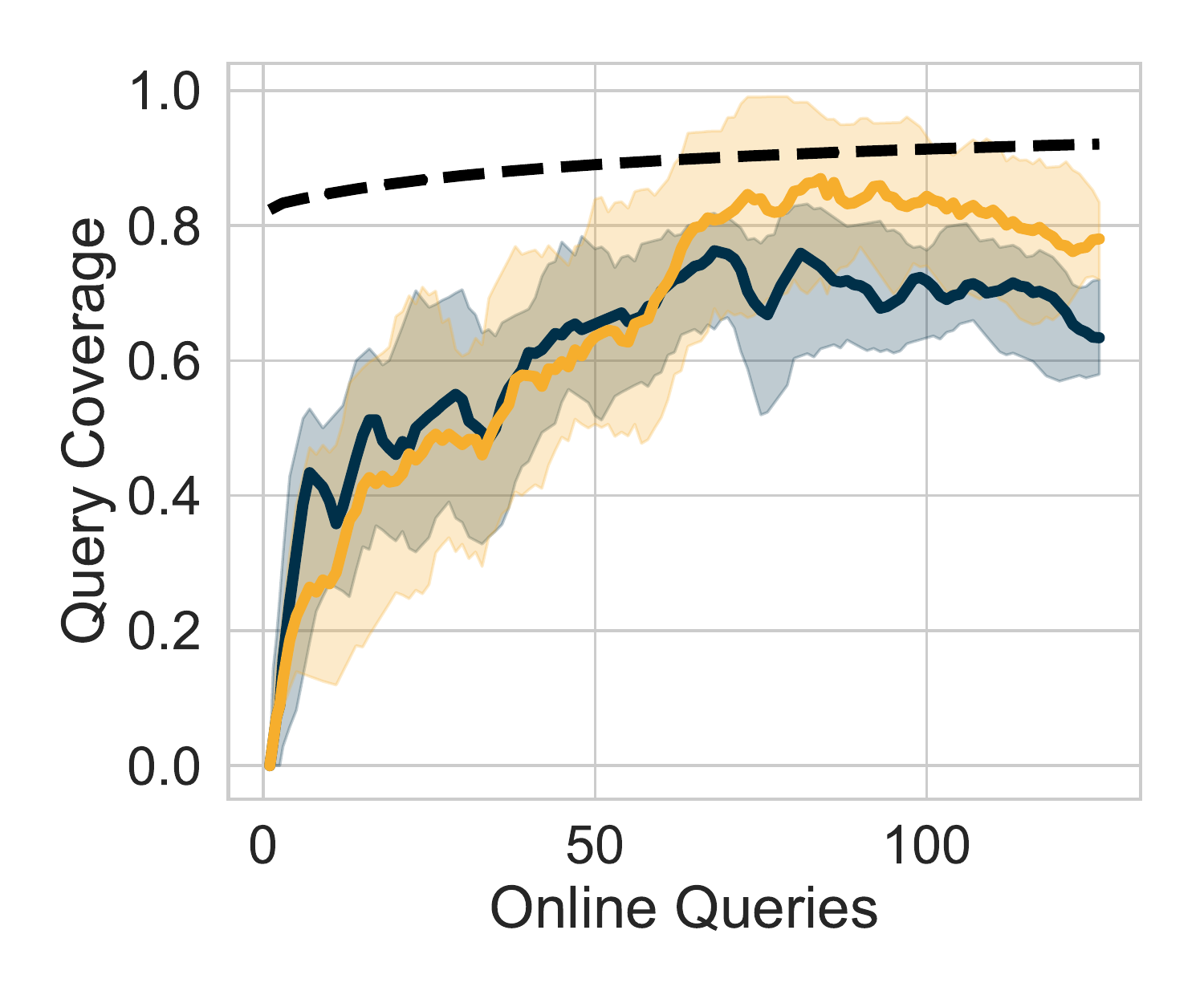}
    \caption{QED}
    \label{fig:tab_bandit_qed}
    \end{subfigure}   
\begin{subfigure}{0.21\textwidth}
\includegraphics[width=\textwidth]{figures/tab_bandits_query-conf-cvrg_zinc-3pbl-docking_v0.0.5.pdf}
    \caption{DRD3 binding affinity}
    \label{fig:tab_bandit_docking}
    \end{subfigure}   
\begin{subfigure}{0.21\textwidth}
    \includegraphics[width=\textwidth]{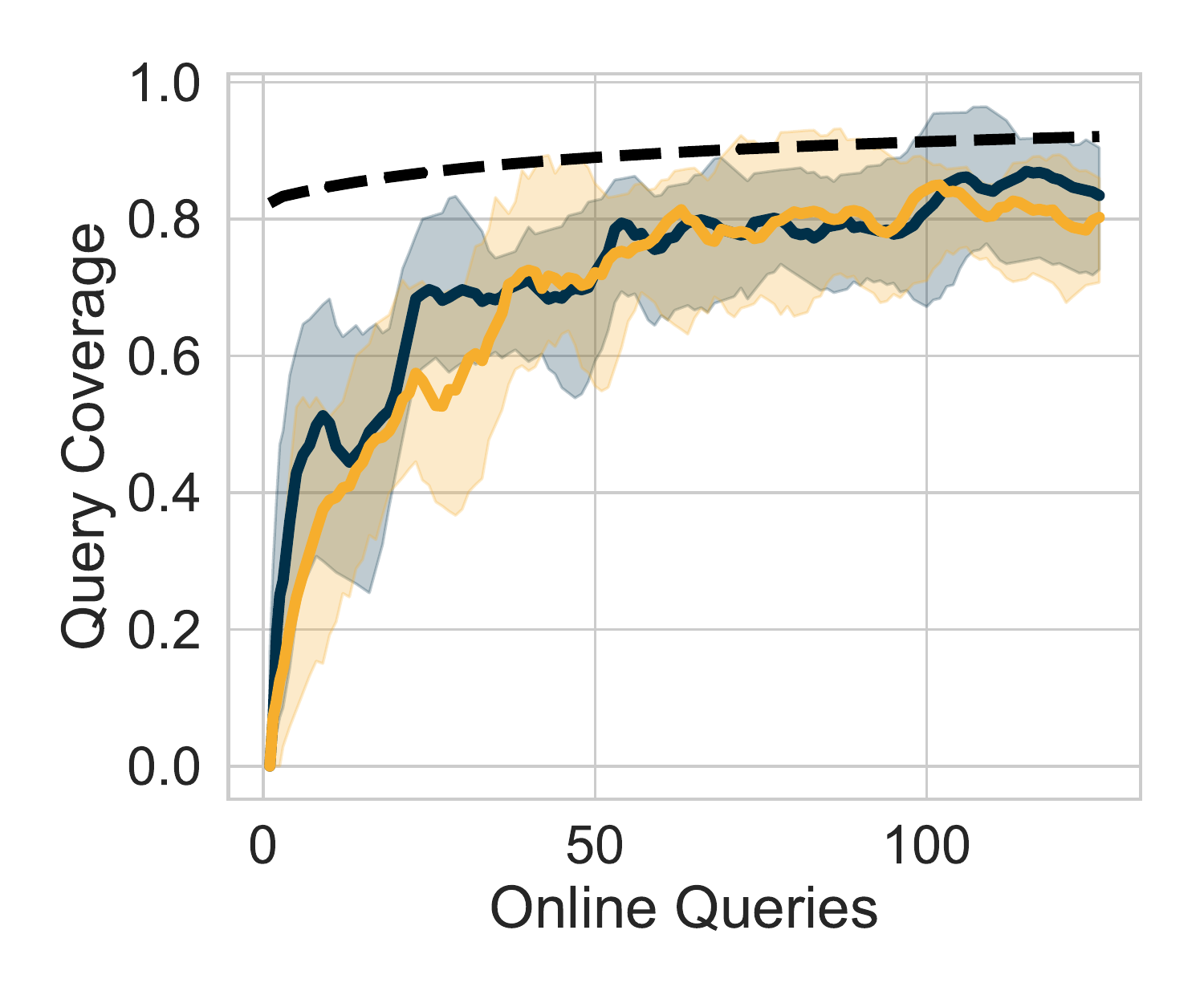}
    \caption{OAS Protein Stability}
    \label{fig:tab_bandit_stability}
\end{subfigure}
\caption{Result ranking tabular molecular datasets for drug-related properties such as solubility (logP) \textbf{(a)}, empirical drug-likeness score (QED) \textbf{(b)}, dopamine receptor (DRD3) binding affinity \textbf{(c)} and antibody stability \textbf{(d)}.
Across datasets, CUCB selects queries with more consistent coverage than UCB \textbf{(bottom two rows)}, with identical sample efficiency\textbf{ (top row)}.
The midpoint, lower, and upper bounds of each curve depict the 50\%, 20\%, and 80\% quantiles, estimated from 4 trials.
}
\label{fig:tabular_bandit_results}
\end{figure}

Sometimes instead of solving $\max_{\vec x \in \mathcal{X}} a(\vec x, \mathcal{D})$, the search space is restricted to a discrete subset of candidates $X_{\mathrm{cand}} \subset \mathcal{X}$.
This restriction is particularly common for tasks with discrete decision variables, such as biological sequence design \citep{stanton2022accelerating}.
This existence of a fixed candidate set simplifies the computation of conformal acquisition functions substantially, since we can use samples from $X_{\mathrm{cand}}$ directly when training the ratio estimator $\hat r$, rather than relying on bootstrapped SGLD as discussed in Section \ref{subsec:handling_covariate_shift}.

To emulate this kind of application, we compared standard and conformal UCB on a selection of small and large molecule ranking tasks.
In particular, we ranked a subset of small molecules drawn from the ZINC dataset \citep{krenn2020self} for three target properties, penalized logP (solubility), QED (drug-likeness), and DRD3 (dopamine receptor) binding affinity \citep{gomez2018automatic, huang2021therapeutics}.
We also ranked a subset of large antibody molecules drawn from the OAS dataset \citep{hornung2014oas} for stability.
For simplicity we did not use sequence-based representations of the molecules, instead relying on RDKit chemical descriptors \citep{Landrum2016RDKit2016_09_4} and BioPython sequence descriptors \citep{cock2009biopython} to generate continuous feature representations of the small and large molecules, respectively.

Starting with the 32 worst entries in our labeled dataset, we selected 128 candidates sequentially $(q=1)$, revealing the corresponding label and retraining the surrogate after each new selection. We share our results in Figure \ref{fig:tabular_bandit_results}.
Because selection is restricted to a prespecified candidate set, the coverage is less consistent than the black-box optimization setting, however we find that conformal UCB still selects queries with better coverage overall, without sacrificing sample-efficiency (measured by cumulative regret, i.e. the difference between the selected candidate label and the best possible label of the remaining candidates).

\subsection{Comparing Bayesian Credible Sets and Conformal Bayes Prediction Sets in the Well-Specified Regime}

\begin{figure}[h]
    \centering
    \begin{subfigure}{0.32\textwidth}
        \includegraphics[width=\textwidth]{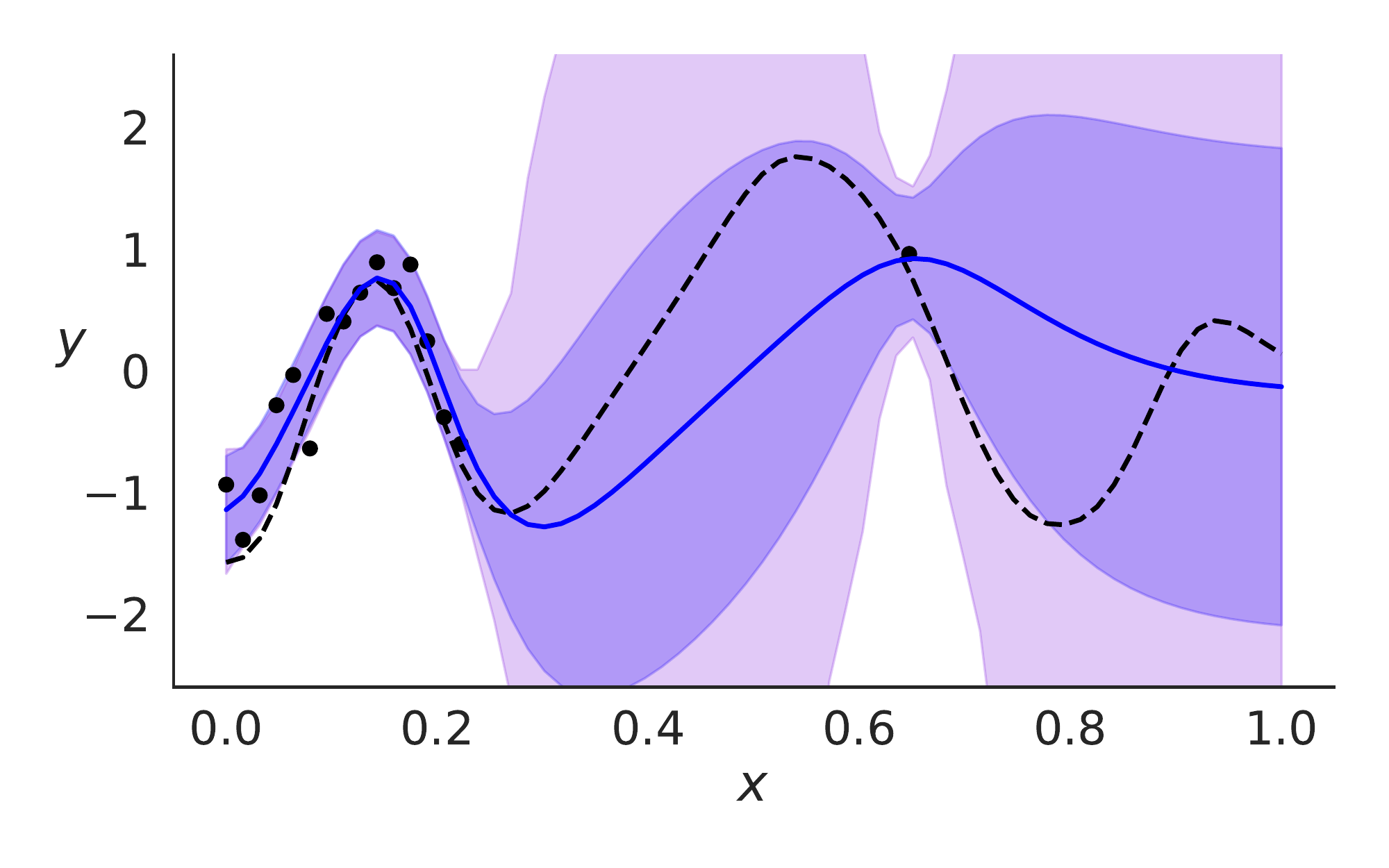}
        \caption{$t=2$}
    \end{subfigure}
    \begin{subfigure}{0.32\textwidth}
        \includegraphics[width=\textwidth]{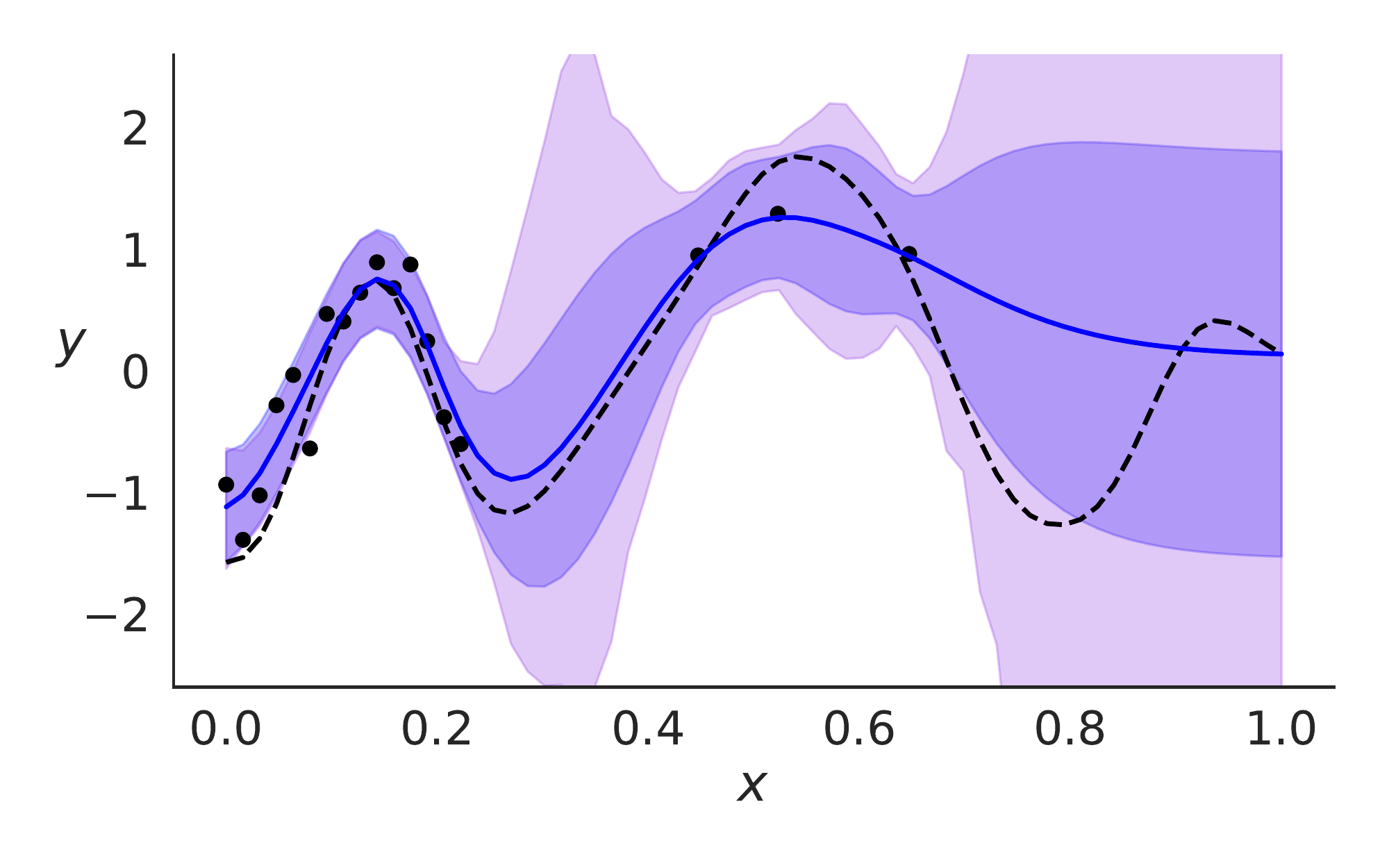}
        \caption{$t=4$}
    \end{subfigure}
    \begin{subfigure}{0.32\textwidth}
        \includegraphics[width=\textwidth]{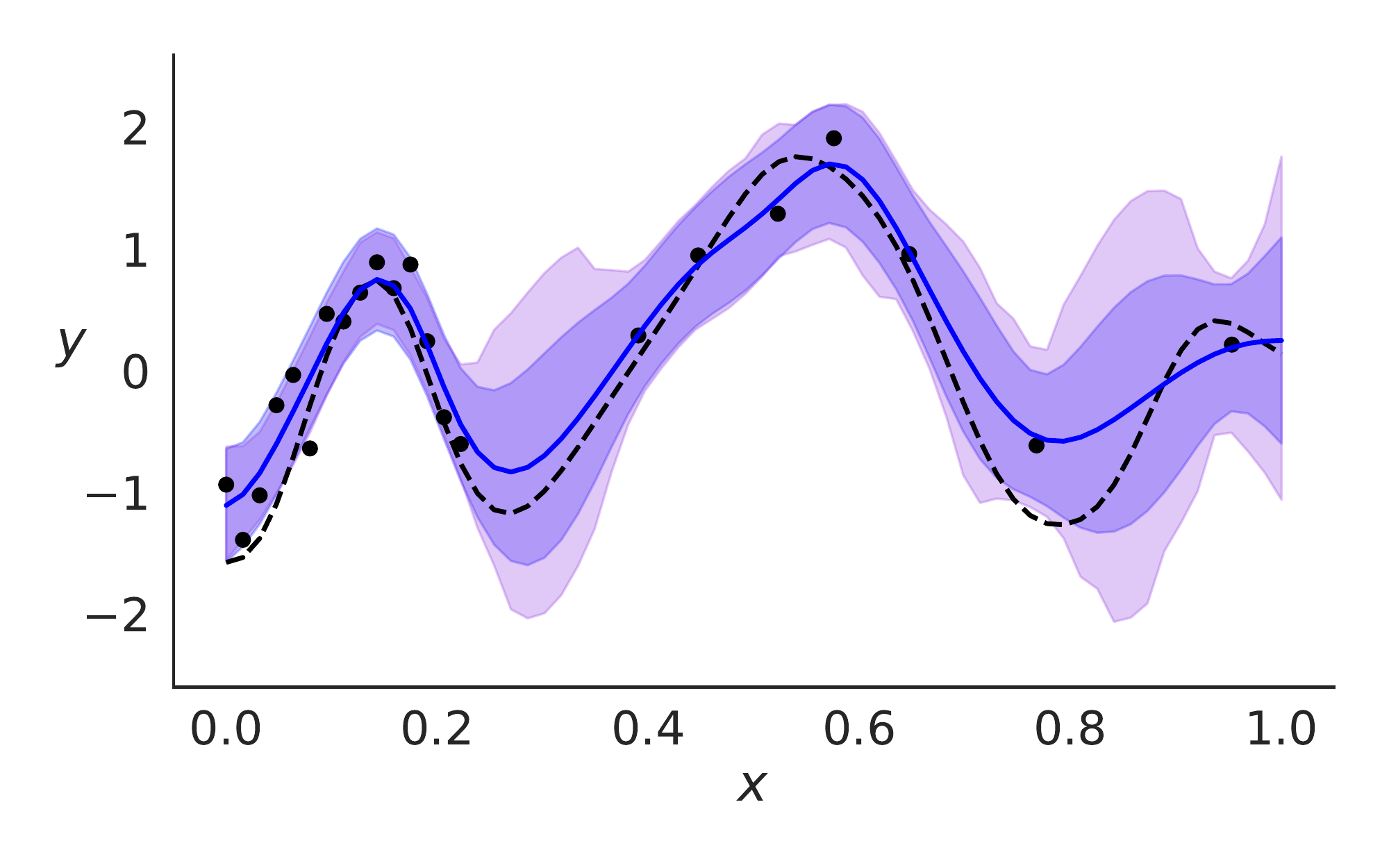}
        \caption{$t=8$}
    \end{subfigure}
    \caption{
        A qualitative example of the difference between conformal Bayes prediction sets and Bayes credible sets in the well-specified regime ($f^*$ generated with a Mat\'ern-$5/2$ kernel with lengthscale $\ell = 0.1$ and $n = 16$ basis points).
        In regions with plentiful training data conformal and credible predictions sets are essentially indistinguishable.
        In regions where training data is sparse, conformal prediction sets are underconfident and much wider than credible sets.
        The credible sets are well-calibrated across the full domain regardless of the amount of training data because the GP prior is highly concordant with the actual target function.
    }
    \label{fig:gpr-mean-example}
\end{figure}

In Appendix \ref{subsec:gen_target_fn} we discussed a method for generating target functions from a pre-specified RKHS.
Not only do target functions generated in this way have a computable RKHS norm, they also allow us to compare the behavior of conformal Bayes prediction sets and Bayes credible sets in the well-specified regime.
Since we know the kernel we used to generate the target function, we can use the same kernel during inference, eliminating one of the possible causes of poor coverage. 

As we noted in Section \ref{subsec:conformal_prediction_background}, conformal Bayes produces the most efficient (i.e. the smallest by volume in expectation under $p(f)$) prediction sets among all prediction rules which are guaranteed to be valid at the $1 - \alpha$ level.
Nevertheless the validity guarantee does come with a price.
In Figure \ref{fig:gpr-mean-example} we show that in the well-specified regime conformal Bayes tends to be underconfident where training data is sparse, whereas Bayes credible sets are well-calibrated across the whole domain.
This result is expected and typical of Bayesian methods.
If we have very good knowledge of the nature of $f^*$ we are better off fully exploiting that knowledge than relying solely on vague assumptions (e.g. pseudo-exchangeability).
The choice between conformal Bayes and conventional Bayesian methods is inherently context-dependent, a function of the available data, the user's confidence in their prior and the cost of miscalibration if that confidence is misplaced.

\begin{figure}[!t]
    \centering
    \begin{subfigure}{0.3\textwidth}
        \includegraphics[height=2.6cm]{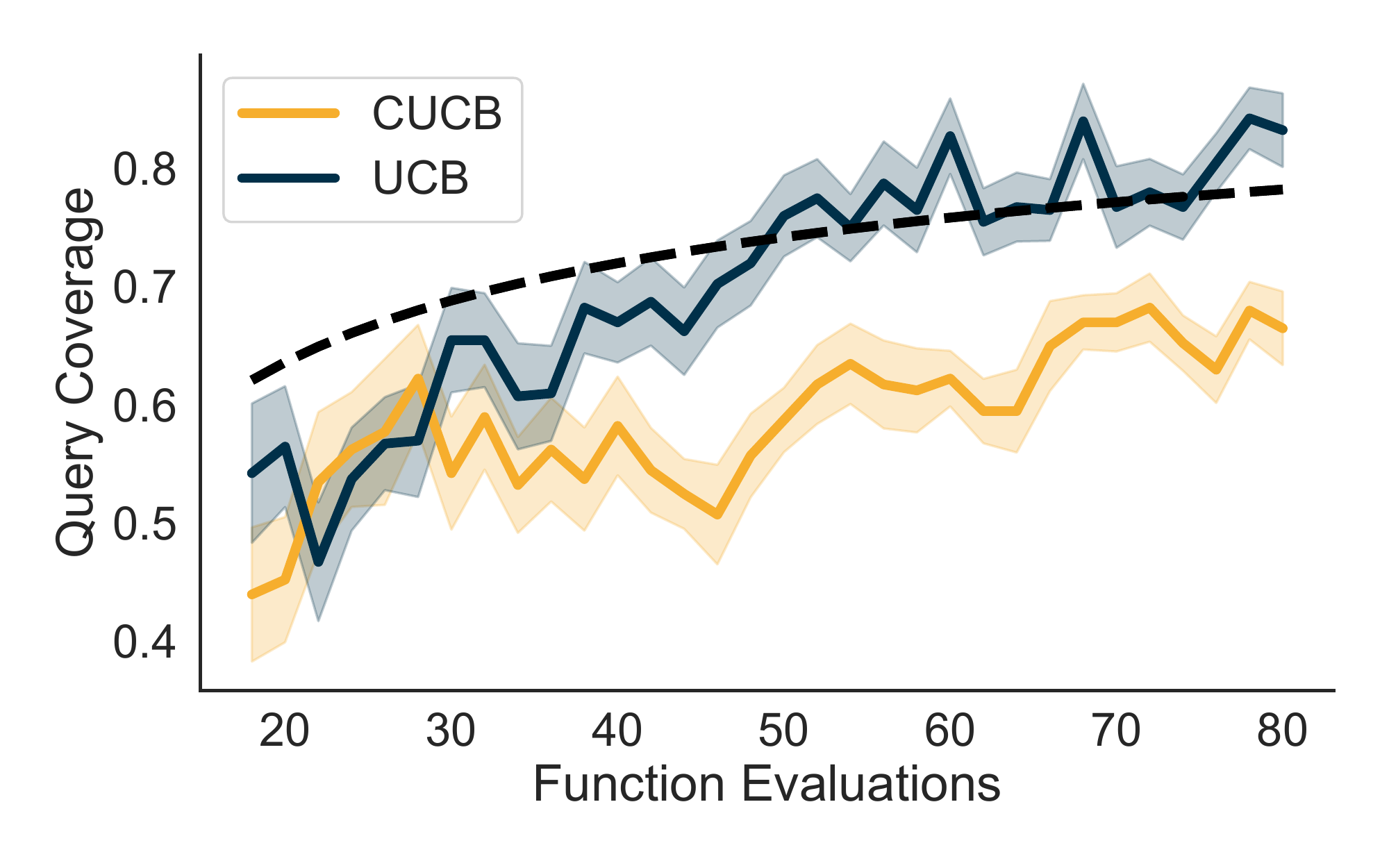}
        \caption{Query Coverage}
    \end{subfigure}
    \begin{subfigure}{0.3\textwidth}
        \includegraphics[height=2.6cm]{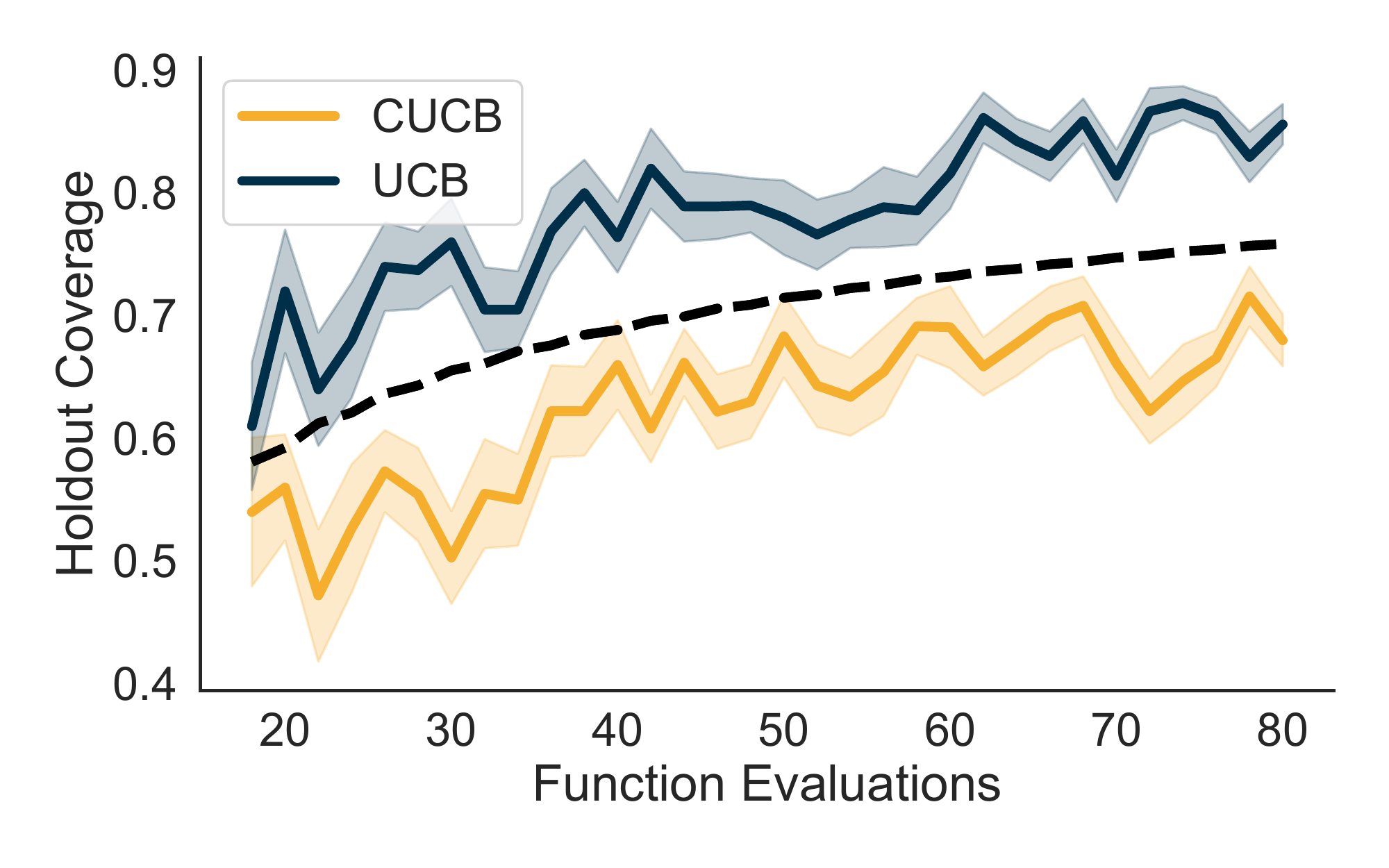}
        \caption{Holdout Coverage}
    \end{subfigure}
    \begin{subfigure}{0.3\textwidth}
        \includegraphics[height=2.6cm]{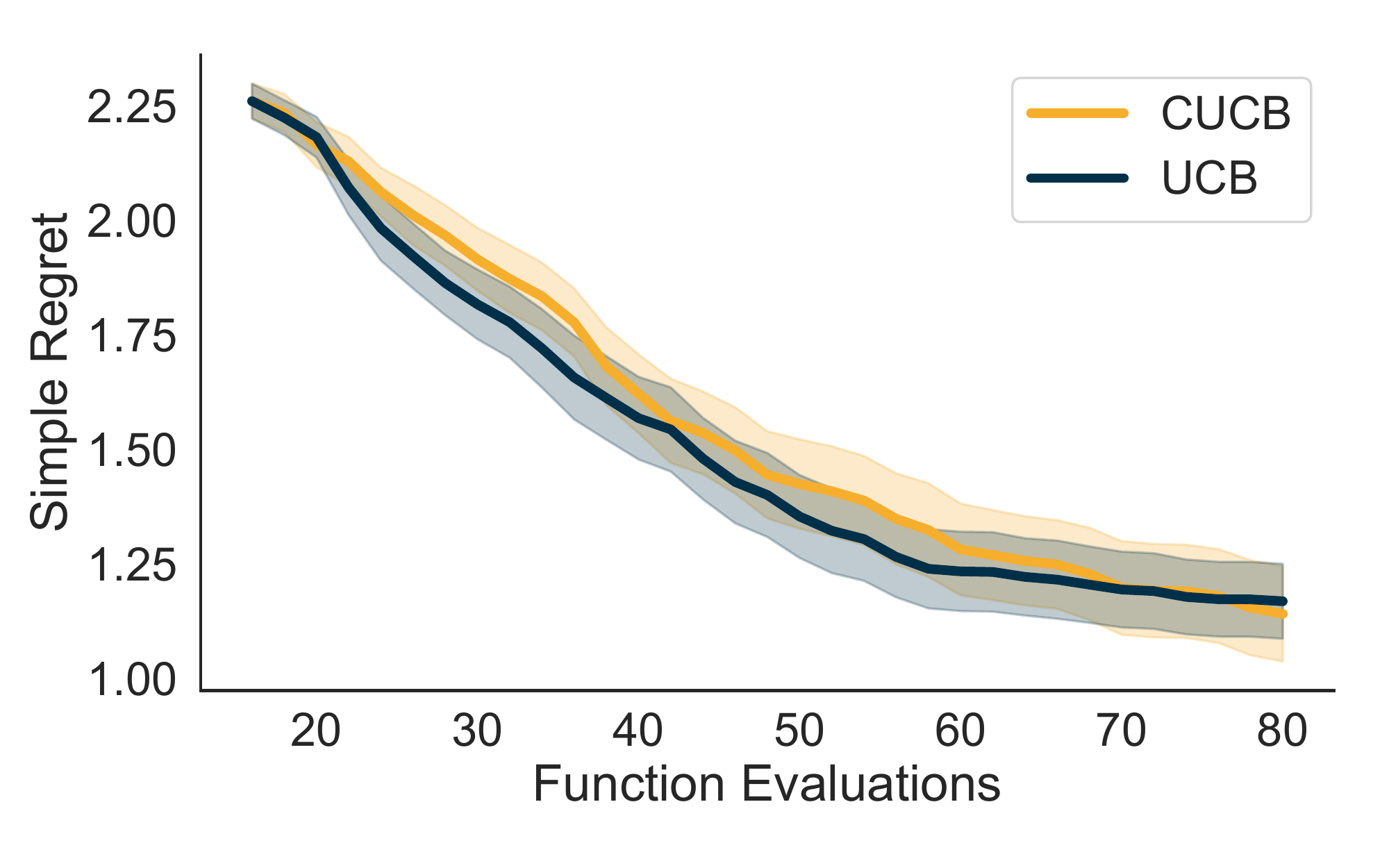}
        \caption{Simple Regret}
    \end{subfigure}
\caption{
    In this experiment we examine the coverage and simple regret of BayesOpt in the well-specified regime.
    We plot the mean and the standard error estimated from 25 trials.
    Here $f^*$ is generated with a Mat\'ern-$5/2$ kernel with lengthscale $\ell = 0.1$ and $n = 128$ basis points in $[0,1]^5$.
    As expected, Bayesian credible sets (indicated by the \textbf{UCB} curve) are fairly well-calibrated.
    The coverage of conformal prediction sets is less satisfactory (the \textbf{CUCB} curve), due in part to the fact that the training data is not IID, and in part to density ratio approximation error.
}
\label{fig:well_specified_bbo}
\end{figure}

In Figure \ref{fig:well_specified_bbo} we repeat the experiment in Section \ref{subsec:single_obj_results} using the same procedure as above to generate $f^*$, increasing the input dimension to 5 and the number of basis points to 128. 
We report the query and holdout coverage, along with the simple regret, $f^*(\vec x^*) - \max_{\vec x_i \in \mathcal{D}} f^*(\vec x_i)$.
Although there is no advantage to using conformal prediction in the well-specified regime, we find the performance to be comparable.

\clearpage
\section{IMPLEMENTATION DETAILS}
\label{app:imp_details}

\subsection{Bringing Everything Together}
\label{subsec:method_summary}

  \begin{algorithm}[h]
    \caption{Pseudocode for the conformal BayesOpt inner loop}
    \label{alg:conf_bo_inner_loop}
    \begin{algorithmic}
      \STATE \textbf{Input:} train data $\mathcal{D} = \{(\vec x_i, y_i)\}_{i=0}^{n-1}$, initial solution $\vec x_n$, score function $s$, miscoverage tolerance $\alpha$, sigmoid temperature $\tau_{\sigma}$, SGLD learning rate $\eta_{\vec x}$, \# SGLD steps $t_{\max}$, SGLD temperature $\tau_{\mathrm{SGLD}}$, classifier learning rate $\eta_\theta$, EMA parameter $\gamma$. \\
      \STATE Initialize classifier $q_\theta$, set weight average $\bar \theta = \vec 0$. \\
      \STATE Initialize classifier dataset $\mathcal{D}' = \{(\vec x_i, 0)\}_{i=0}^{n-1}$ \\
      \FOR {$t = 0, \dots, t_{\max} - 1$}
      \STATE Estimate $\hat r_t(\vec x_i), \forall i \in \{0, \dots, n\}$ with $q_{\bar \theta}$. \hfill (Eq. \ref{eq:density_ratio_bayes_rule})
      \STATE $(\vec w_t)_i = \hat r_t(\vec x_i) / \sum_k \hat r_t(\vec x_k), \forall i \in \{0, \dots, n\}$.
      \STATE $Y_{\mathrm{cand}} \leftarrow \{\hat{\vec y}_j\}_{j=0}^{m-1}$ s.t. $\hat{\vec y}_j \sim \hat p(\vec y | \vec x'_t, \mathcal{D})$. \\
      \STATE $\vec m = \mathrm{outcome\_mask}(\mathcal{D}, \vec x_n, \vec w_t, Y_{\mathrm{cand}}, s, \alpha, \tau_\sigma)$. \hfill (Algorithm \ref{alg:differentiable_conformal_masks}) \\
      \STATE Estimate acquisition value $a(\vec x_n)$. \hfill (Eq. \eqref{eq:conf_acq_mc_est})
      \STATE Update $\vec x_n \leftarrow \mathrm{sgld\_step}(\vec x_n, a(\vec x_n), \eta_{\vec x}, \tau_{\mathrm{SGLD}})$.
      \STATE Update $\mathcal{D}' \leftarrow \mathcal{D}' \cup \{(\vec x_n, 1)\}$.
      \STATE Update $\theta \leftarrow \theta - \eta_\theta \nabla_\theta \ell(\theta, \mathcal{D}')$
      \STATE Update $\bar \theta \leftarrow (1 - \gamma) \bar \theta + \gamma \theta$.
      \ENDFOR
      \STATE \textbf{Return:} $\vec x_n$
    \end{algorithmic}
  \end{algorithm}

In Algorithm \ref{alg:conf_bo_inner_loop} we summarize the entire conformal BayesOpt inner loop used to select new queries.

\subsection{Stable Predictions on the Training Set}
We found that computing the GP posterior negative log-likelihood (and its gradients) on training data to be numerically unstable and so used stochastic diagonal estimation to estimate the posterior variances.
Plugging in $K = \kappa(X, X)$ into that posterior mean and variance, we get that the posterior mean is $K(K+\sigma^2)^{-1}\vec y$ and the posterior covariance is $\Sigma = \sigma^2 I + K - K(K + \sigma^2 I)^{-1}K.$
Unfortunately, the second term ends up being unstable as it requires solving (and then subtracting) a (batched) system of size $n \times n$.
To see the reason for instability, note that as $\sigma^2 I \rightarrow 0$ then the entire covariance matrix tends to zero.

We originally tried backpropagating through an eigendecomposition; however, this produced ill-defined gradients, see the explanation in \citet{ionescu2015training}.
Instead we computed a stochastic diagonal estimate, using the identities
\begin{align}
\Sigma &= \sigma^2 I + \sigma^2 K(K+\sigma^2 I)^{-1}, \\
\text{diag}(\Sigma)_i &\approx \sigma^2\left(1 + \left(\sum_{j=1}^J \vec z^{(j)} \odot K(K+\sigma^2 I)^{-1}\vec z^{(j)}\right)_i\left(\sum_{j=1}^J \vec z^{(j)} \odot \vec z^{(j)}\right)_i^{-1}\right),
\end{align}
where the probe vector $\vec z^{(j)}$ has i.i.d Bernoulli entries and $\odot$ is the Hadamard product.
This estimator comes from \citet{bekas2007estimator} and is in spirit quite similar to Hutchinson's trace estimator for the log determinant.
We used $J = 10$ probe vectors.

\subsection{Hyperparameters}

For all GP models in this paper, we used the default single task GP (\texttt{SingleTaskGP}) model from BoTorch, which uses a scaled Matern-$5/2$ kernel with automatic relevance determination and a $\text{Gamma}(3, 6)$ prior over the lengthscales and a $\text{Gamma}(2, 0.15)$ prior over the outputscales. 
We used constant prior mean functions.
For the likelihood, we used a softplus transformation to optimize the raw noise, constraining the noise to be between $5 \times 10^{-4}$ and $0.5.$
To fit the GP kernel hyperparameters $\phi$, we used BoTorch's default fitting utility, \texttt{fit\_gpytorch\_model}, which uses L-BFGS-B to maximize the log-marginal likelihood $\log p(\mathcal{D} | \phi)$.

For the miscoverage tolerance we used a simple schedule $\alpha = \max(0.05, 1 / \sqrt{n})$. Note if $\alpha < 1 / n$ then $\mathcal{C}_\alpha(\vec x) = \mathcal{Y}, \; \forall \vec x \in \mathcal{X}$.

We initialized $\mathcal{D}_0$ with $10$ Sobol points drawn from a random orthant of the normalized input space.
The input normalization was computed from known bounds of the black-box functions.

\begin{table}[h!]
\centering
\parbox{.9\linewidth}{
\centering
\begin{tabular}{|l|c|}
\hline
\multicolumn{2}{|c|}{\textbf{Black-Box Optimization Hyperparameters}}\\
\hline
\textbf{Name} & \textbf{Value} \\
\hline
\# Optimization rounds                 & 50                        \\
\hline
$q$ (query batch size) & $\{1, 3\}$ \\
\hline
$|\mathcal{D}_0|$ & 10 \\
\hline
$\sigma$ (normalized measurement noise scale) & 0.1 \\
\hline
$\tau_\sigma$ (sigmoid temp.) & 1e-2   \\
\hline
$k \; (\text{i.e. } |Y_{\mathrm{cand}}|)$ & 256 \\
\hline
\# SGLD chains                     & 5                    \\
\hline
$t_{\mathrm{max}}$  (\# SGLD total steps)                       & 100                       \\
\hline
$t_{\mathrm{burn}}$  (\# SGLD burn-in steps)                       & 25                       \\
\hline
$\eta_{\vec x}$ (SGLD learning rate)      & 1e-3 \\
\hline
$\tau_{\mathrm{SGLD}}$ (SGLD temp.) & 1e-3 \\
\hline
$\eta_\theta$ (classifier learning rate) & 1e-3 \\
\hline
$\gamma$ (classifier EMA weight) & 2e-2 \\
\hline
$\lambda$ (classifier weight decay) & 1e-4 \\
\hline
Random seeds & $\{0,\dots, 24\}$ \\
\hline
\end{tabular}
}
\end{table}

\begin{table}[h!]
\centering
\parbox{.9\linewidth}{
\centering
\begin{tabular}{|l|c|}
\hline
\multicolumn{2}{|c|}{\textbf{Tabular Ranking Hyperparameters}}\\
\hline
\textbf{Name} & \textbf{Value} \\
\hline
\# Optimization rounds                 & 128                        \\
\hline
$q$ (query batch size) & 1 \\
\hline
$|\mathcal{D}_0|$ & 32 \\
\hline
$\sigma$ (normalized measurement noise scale) & n/a \\
\hline
$\tau_\sigma$ (sigmoid temp.) & 1e-6   \\
\hline
$k \; (\text{i.e. } |Y_{\mathrm{cand}}|)$ & 64 \\
\hline
\# number classifier gradient updates & 256 \\
\hline
$\eta_\theta$ (classifier learning rate) & 1e-3 \\
\hline
$\gamma$ (classifier EMA weight) & 1 \\
\hline
$\lambda$ (classifier weight decay) & 1e-4 \\
\hline
Random seeds & $\{0,\dots, 3\}$ \\
\hline
\end{tabular}
}
\end{table}

\subsection{Computational Complexity}
\label{subsec:complexity}

The cost of training the surrogate GP regression model is the same in our case as conventional BayesOpt, namely $\mathcal{O}(n^3)$ if exact GP inference is used without any approximations.

The cost of \textit{retraining} the surrogate GP on a single new example $(\vec x_n, \hat{\vec y}_j)$ is $\mathcal{O}(n)$, since one can make use of efficient low-rank updates to the root decomposition of $(K_{XX} + \sigma^2 I)^{-1}$ \citep{gardner2018gpytorch}.
The surrogate GP can be retrained on all $k$ candidate labels in parallel on a GPU, which keeps the wall-clock cost of retraining to $\mathcal{O}(n)$ but increases the memory footprint by a factor of $k$. The increased memory footprint persists for the duration of the selection phase, during which the acquisition function is optimized to select the next query.

The cost of drawing a sample function $f^{(j)} \sim p_\alpha(f | \hat{\mathcal{D}}_j)$ is the same as drawing a sample from $p(f | \mathcal{D})$, and whereas conventional BayesOpt methods typically draw multiple samples from $p(f | \mathcal{D})$, we find that drawing a single sample from each $p_\alpha(f | \hat{\mathcal{D}}_j)$ is sufficient since $p_\alpha(f(\vec x_n) | \hat{\mathcal{D}}_j)$ tends to concentrate near $\hat{\vec y}_j$.

Therefore the complexity of each acquisition function gradient evaluation is dominated by the same $\mathcal{O}(qn^2)$ cost of exact GP test-time inference with $q$ test examples as conventional BayesOpt, at the cost of a $k$-factor increase in memory usage.

Although this analysis would seem to indicate that conformal BayesOpt should run in similar wall-clock time to conventional BayesOpt, in practice it is considerably slower because the SGLD chains in conformal BayesOpt must be allowed to burn in, and then the bootstrapped ratio estimator must be given time to converge, which requires many more gradient evaluations than optimizing a conventional BayesOpt acquisition function with L-BFGS.
In our experiments we found conformal BayesOpt to be around an order of magnitude slower in terms of wall-clock time than conventional BayesOpt.
This increase in runtime is a considerable drawback and merits further investigation in future work.

\subsection{Compute Resources}
Our experiments were conducted on a range of NVIDIA GPUs, including RTX 2080 Tis, Titan RTXs, V100s, and A100s in high-performance computing clusters.
All experiments used a single GPU at a time.
It would require approximately 250 GPU hours to reproduce the experiments in this paper by our estimate,
\begin{align*}
    1 \text{ GPU hr/seed} \times 25 \text{ seeds per variant} \times 1 \text{ variant per experiment} \times 10 \text{ experiments} = 250 \text{ hrs}.
\end{align*}
Other experimental runs, e.g. development and debugging, probably consumed an order of magnitude more GPU hours.

\subsection{Software Packages}

\begin{itemize}
    \item Python 3, PSF License Agreement \citep{vanrossum2009python3}.
    \item Matplotlib, Matplotlib License Agreement.
    \item Seaborn, BSD License.
    \item NumPy, BSD License \citep{harris2020array}.
    \item PyTorch, BSD License \citep{paske2019pytorch}.
    \item GPyTorch, MIT License \citep{gardner2018gpytorch}.
    \item BoTorch, MIT License \citep{balandat_botorch_2020}.
\end{itemize}

\end{document}

%% file: macros.tex
\renewcommand{\vec}[1]{\mathbf{#1}}

\newcommand{\argmax}{\mathop{\mathrm{argmax}}}

\newcommand{\xxcomment}[4]{\textcolor{#1}{[$^{\textsc{#2}}_{\textsc{#3}}$ #4]}}

\newcommand{\todo}[1]{\xxcomment{orange}{TO}{DO}{#1}}

\definecolor{palette0}{HTML}{780000}
\definecolor{palette5}{HTML}{669BBC}

\renewcommand{\todo}[1]{}